\documentclass[acmsmall]{acmart}

\usepackage{booktabs} % For formal tables

\usepackage[ruled]{algorithm2e} % For algorithms

\SetAlFnt{\small}
\SetAlCapFnt{\small}
\SetAlCapNameFnt{\small}
\SetAlCapHSkip{0pt}
\IncMargin{-\parindent}
\usepackage{multirow}
\usepackage{color}
\definecolor{blue}{RGB}{0,0,0}
\definecolor{red}{RGB}{255,0,0}
\definecolor{blueG}{RGB}{0,0,255}

\usepackage{rotfloat}
\usepackage{booktabs}
\usepackage{threeparttable}
\usepackage{adjustbox}
\usepackage{rotating}
\usepackage{caption}

\newcommand{\tabincell}[2]{\begin{tabular}{@{}#1@{}}#2\end{tabular}}

\usepackage{graphicx}%Put this in the preamble.

\AtBeginDocument{%
	\providecommand\BibTeX{{%
			\normalfont B\kern-0.5em{\scshape i\kern-0.25em b}\kern-0.8em\TeX}}}

\begin{document}
	
	\title{Challenges in Building Intelligent Open-domain Dialog Systems}
	
	\author{Minlie Huang}
	\author{Xiaoyan Zhu}
	\affiliation{%
		\institution{Department of Computer Science and Technology, Institute for Artificial Intelligence, Beijing National Research Center for Information Science and Technology, Tsinghua University, Beijing 100084}
		\city{Beijing}
		\country{China}
	}
	\email{aihuang@tsinghua.edu.cn}
	
	\author{Jianfeng Gao}
	\affiliation{%
		\institution{
			Microsoft Research
		}
		\city{Redmond}
		\country{WA, USA}
	}
	\email{jfgao@microsoft.com}
	
	\renewcommand{\shortauthors}{Huang et al.}
	
	\begin{abstract}
		There is a resurgent interest in developing intelligent open-domain dialog systems due to the availability of large amounts of conversational data and the recent progress on neural approaches to conversational AI \citep{gao2019neural}. Unlike traditional task-oriented bots, an open-domain dialog system
		aims to establish long-term connections with users by satisfying the human need for communication, affection, and social belonging.  
		This paper reviews the recent work 
		on neural approaches that are devoted to addressing three challenges in developing such systems: \emph{semantics}, \emph{consistency}, and \emph{interactiveness}. 
		\emph{Semantics} requires a dialog system to not only understand the content of the dialog but also identify user's emotional and social needs during the conversation. 
		\emph{Consistency} requires the system to demonstrate a consistent personality to win users trust and gain their long-term confidence. \emph{Interactiveness} refers to the system's ability to generate interpersonal responses to achieve particular social goals such as entertainment and conforming. 
		The studies we select to present in this survey is based on our unique views and are by no means complete. Nevertheless, we hope that the discussion will inspire new research in developing more intelligent open-domain dialog systems.
	\end{abstract}

%%%%%copyright
\setcopyright{acmcopyright}
\acmJournal{TOIS}
\acmYear{2020} 
\acmVolume{1} 
\acmNumber{1} 
\acmArticle{1} 
\acmMonth{1} 
\acmPrice{15.00}
\acmDOI{10.1145/3383123}

	

	\maketitle
	
	\section{Introduction}
	
	Building intelligent open-domain dialog systems that can converse with humans coherently and engagingly has been a long-standing goal of artificial intelligence (AI). 
	Early dialog systems such as Eliza \cite{weizenbaum1966eliza}, Parry \cite{colby1971artificial}, and Alice \cite{wallace2009anatomy}, despite being instrumental to significantly advancing machine intelligence, worked well only in constrained environments.  An open-domain social bot remains an elusive goal until recently.  
	%Since the very early dialog system, \textit{ELIZA}~\cite{weizenbaum1966eliza} in 1960s,
	%dialog systems have been significantly advanced in their intelligence capacity and applicability. 
	%%%%In 2011, for the first time, IBM Watson \cite{ferrucci2010building} demonstrated that a factoid question answering system can beat human champions on the quiz show Jeopardy! and pass the Turing Test \cite{alan1950} in understanding a natural language question and finding answers from unstructured data. 
	The Microsoft XiaoIce (`Little Ice' literally in Chinese) system, since its release in May, 2014, has attracted millions of users and can converse with users on a wide variety of topics for hours \cite{zhou2018design, shum2018eliza}.
	%Nowadays, the most notable and successful open-domain dialog system, \textit{XiaoIce} \cite{zhou2018design, shum2018eliza}, attracts millions of users and can converse with users continuously for hours. 
	In 2016, the Alexa Prize challenge was proposed to advance the research and development of social bots
	%conversational agents, known as "social bots"\footnote{In this paper, we think \textit{a social bot} is a mixture of task-oriented and open-domain dialog systems. It may have many sub-systems for dealing with task completion, question answering, chit-chatting, topic-grounded conversation, and so on.}, 
	that are able to converse coherently and engagingly with humans on popular topics such as sports, politics, and entertainment, for at least 20 minutes \cite{chen2018gunrock,ram2018conversational} \footnote{Even though the dialog systems in this challenge are very complicated, they are more informational systems where user emotion need is less considered.}
	. 
	The evaluation metric, inspired by the Turing Test \cite{alan1950}, is designed to test the social bots' capacity of delivering coherent, relevant, interesting, free-form conversations and keeping users engaged as long as possible. 
	%%% 这里要补充CoAI和Alex 
	%%%两种重要的social chatbot的行为
	% Although dialog systems have been significantly advanced by these efforts, 
	However, the general intelligence demonstrated by these systems is still far behind humans. Building open-domain dialog systems that can converse on various topics like humans remains extremely challenging \cite{gao2019neural}. 
	
	In this paper we focus our discussion on three challenges in developing neural-based open-domain dialog systems, namely \emph{semantics}, \emph{consistency} and \emph{interactiveness}. 
	The rest of the paper is structured as follows. In the rest of Section 1, we compare open-domain dialog bots with traditional task-oriented bots and elaborate the three challenges. In Section \ref{sec:frameworks}, we survey three typical approaches to building neural-based open-domain dialog systems, namely, retrieval-based, generation-based, and hybrid methods. In Sections \ref{sec:semantics}, \ref{sec:consistency}, and \ref{sec:interactiveness}, we review the approaches that have been proposed to address the three challenges, respectively. In Section \ref{sec:evaluation}, we discuss recent work on open-domain dialog evaluation. In Section \ref{sec:datasets}, we present an incomplete survey of frequently-used or recently-proposed benchmarks for open-domain conversation modeling. 
	We conclude the paper by presenting several future research trends in Section \ref{sec:future}.  
	
	\subsection{Open-Domain Dialog vs. Task-Oriented Dialog}
	%%%%任务导向和一般闲聊的区别
	Generally speaking, there are two types of dialog systems: 
	task-oriented and open-domain dialog.
	%%%%%%%%this for should be deleted?
	Task-oriented dialog systems are designed for specific domains or tasks, such as flight booking, hotel reservation, customer service, and  technical support, and have been successfully applied in some real-world applications. Open-domain dialog systems, however,  are much more challenging to develop due to its open-ended goal.
	
	As outlined by \citet{gao2019neural}, although both task-oriented dialog and open-domain dialog can be formulated as an optimal decision making process with the goal of maximizing expected reward, the reward in the former is better-defined and much easier to optimize than the latter. Consider a ticket-booking bot. It is straightforward to optimize the bot to get all necessary information to have the ticket booked in minimal dialog turns. The goal of an open-domain dialog agent is to maximize the long-term user engagement. This is difficult to optimize mathematically because there are many different ways (known as dialog \emph{skills}) to improve the engagement (e.g., providing entertainment, giving recommendations, chatting on an interesting topic, providing emotional comforting) and it requires the systems to have a deep understanding of dialog context and user's emotional needs to select the right skill at the right time, and generate interpersonal responses with a consistent personality.
	
	Open-domain dialog systems also differ from task-oriented bots in system architecture. A task-oriented bot is typically developed based on a pre-defined task-specific schema\footnote{A task schema typically defines a set of user intents, and for each intent defines a set of dialog acts, slot-value pairs.} and is designed as a modular system which consists of domain-specific components
	like language understanding, dialog management\footnote{Dialog management performs both dialog state tracking \cite{henderson2013deep,Mrk2017ACL} and response selection via policy \cite{zhao2016towards,Peng2017EMNLP,Su2016ACL,lipton2018bbq}. }, and language generation\footnote{Recently, there are end-to-end methods \cite{wen2017network,bordes2016learning,zhang2019tois} that output a response given the previous dialog history. But in general, domain knowledge about the task needs to be explicitly considered, which differs significantly from open-domain dialog systems.}. These components can be either hand-crafted based on domain knowledge or trained on task-specific labeled data. 
	On the other hand, due to the open-ended nature, open-domain dialog systems need to deal with open-domain knowledge without any pre-defined task-specific schemas or labels. In recent years, there has been a trend towards developing fully data-driven, end-to-end systems that map user's input to system's response using neural networks. Since the primary goal of open-domain dialog bots is to be AI companions to humans with an emotional connection rather than completing specific tasks, they are often developed to mimic human conversations by training neural response generation models on large amounts of 
	\cite{sordoni2015neural,vinyals2015neural,shang-2015-NRM}.
	
	Unlike task-oriented bots, most neural response generation models developed for open-domain dialog systems are not grounded in real world, which prevents these systems from effectively conversing about anything that relates to the user's environment. Only recently have researchers begun to explore how to ground open-domain dialog systems in real-world entities and knowledge \cite{ghazvininejad2017knowledge,mostafazadeh2017image,qin2019conversing}. Knowledge grounding is also crucial for the system to provide interpersonal responses. For instance, the conversations between friends are quite different from those between strangers. So the system needs to be grounded in the personas of the speaker and addressee, respectively \cite{Li2016_ACL-persona}. The tone of system responses needs to be adjusted according to user's emotional states and affect by grounding in affect or emotion of the user \cite{huber2018emotional,winata2017nora,xu2018emo2vec}.

	\begin{figure}[!htp]
		\centering
		\includegraphics[width=1.0\linewidth]{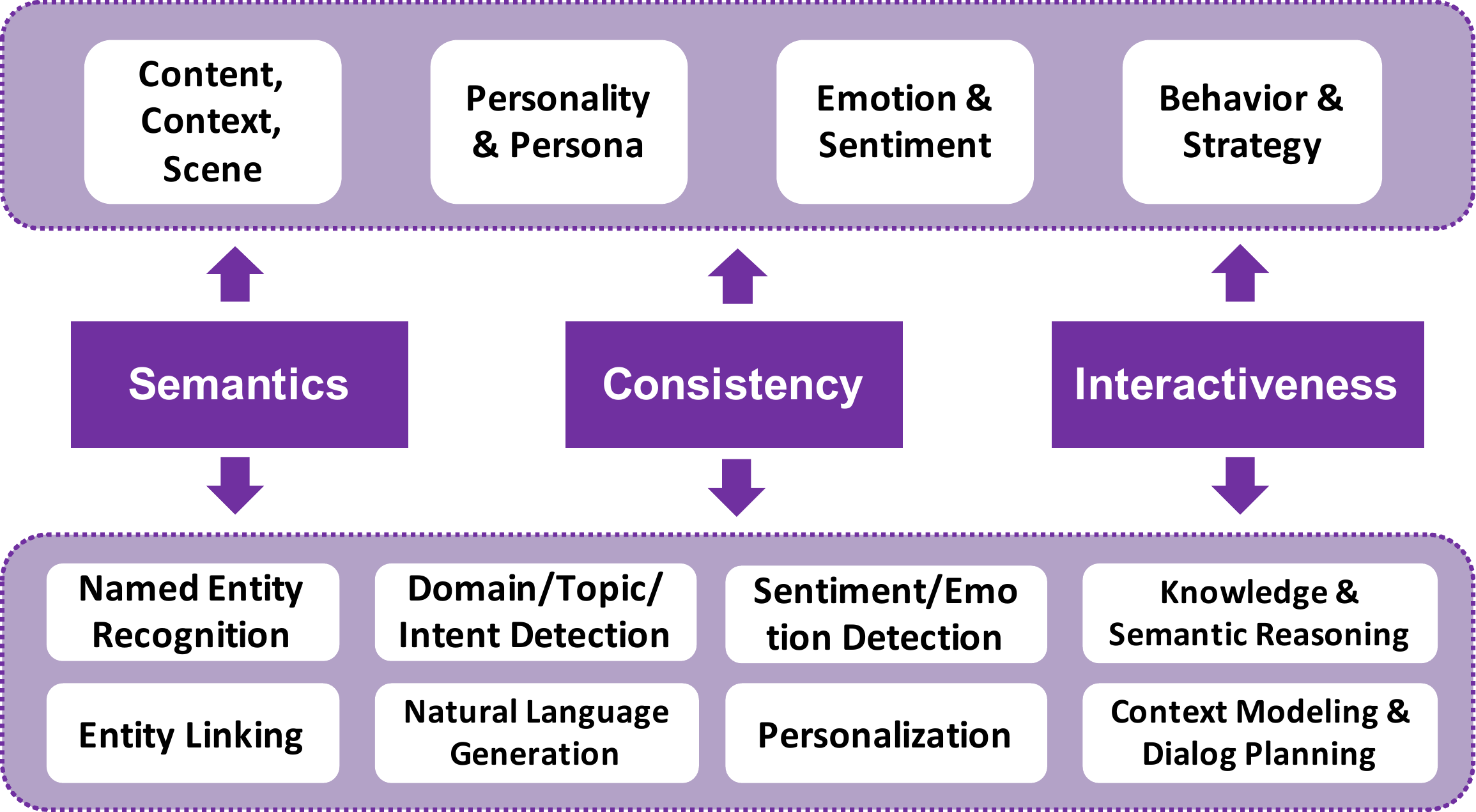}
		\caption{Key issues (the middle block), key factors (the upper block), and key technologies (the bottom block) in open-domain dialog systems.}
		\label{fig:overview}
	\end{figure} 
	
	\subsection{Three Challenges}
	Thanks to the recent success of neural approaches to natural language processing and conversational AI, there have been substantial advancements on open-domain dialog systems. 
	
	Most neural approaches formulate conversation as a task of generating output response given user input and dialog context. Almost all neural response generation models follow the neural text generation frameworks, such as sequence-to-sequence (Seq2Seq) \cite{2014sequence,vinyals2015neural,shang-2015-NRM}, conditional variational autoencoder (CVAE) \cite{sohn2015learning}, and generative adversarial network (GAN) \cite{li2017adversarial}. 
	These models are trained on large-scale conversational data, and can deliver plausible responses to any topic that users are interested in. However, these neural-based systems often fail to establish long-term emotional connections with users due to many issues as illustrated by the examples in Table \ref{tab:dialog-exm} and summarized in Figure \ref{fig:overview}. In what follows, we elaborate three key issues which we will discuss in the rest of the paper \footnote{Note that the challenges discussed in this section are also fundamental to traditional, non-neural dialog systems.}.
	%at many times but also fail to produce appropriate output in many cases. However, we argue that there still exist three key issues to be solved with tremendous research efforts. We present some examples to demonstrate these issues in Table \ref{tab:dialog-exm}, and discuss the issues more formally as follows (see Figure \ref{fig:overview}):

	%%%%这几个例子要修改为地道一点英语
	%%%%%jianfeng --- can you make some better or natural examples？
	%%%%reviewer's comment: The examples in Table 1 could be made into more natural native English.  They feel stilted and awkward; not very natural -- even for the human side of the dialogs. 
	
	\begin{table*} [!htp]
		\centering
		\small
		\begin{tabular}{l|l}
			\hline
			\multicolumn{2}{c}{Semantics Issue} \\
			\hline
			User & I come from Shanghai. \\
			Machine & Cool! I am from Beijing, we come from the same city! \\
			\hline
			User & Back to Beijing now. It is so cold but I am very happy! \\
			Machine & When will you come to Beijing? \\
			\hline
			\multicolumn{2}{c}{Consistency Issue} \\
			\hline
			User & Are a boy or a girl? \\
			Machine & I am a boy. \\
			User & Are you a girl? \\
			Machine & Yes, I am a girl. \\
			\hline
			\multicolumn{2}{c}{Interactiveness Issue} \\
			\hline
			User & Terrible breakfast today. \\
			Machine & How about together tomorrow morning? \\
			User & Sure. Would be great with your recommendation. \\
			Machine & Really? \\
			\hline
		\end{tabular}
		\caption{
			Examples for demonstrating the semantics, consistency, and interactiveness issues. The first two examples show semantic conflicts between user post and machine response, the third session shows bad consistency within the dialog context due to the lack of a coherent personality, and the last session has bad interactiveness due to the lack of grounding. The results in the first two blocks are from a standard Seq2Seq model with an attention mechanism, and the last session is from a commercial system. }
		\label{tab:dialog-exm}
	\end{table*}

	%\noindent $\bullet$ \textbf{Semantics} 
	\paragraph{Semantics} 
	Semantics is the heart of any dialog system because conversation is a semantic activity \cite{Zhou2018Commonsense}. 
	It requires not only to \emph{understand the content of the conversation}, which is often in multi-modality including text, image and video,  
	%understand the content, context, and scene of the conversation, but also to 
	% Semantics understanding even goes beyond the utterances, requiring to 
	but also to 
	% process multi-modal information 
	\emph{understand users} by identifying information beyond the dialog content such as a user's personality  and persona\footnote{Personality is someone's character or nature while a persona is a superficial identity of the character or nature.}, emotion, sentiment, and the user's profile and background. 
	From the technical perspective, semantics mainly involves the key techniques of \emph{natural language understanding and user understanding}, including named entity recognition, entity linking, domain detection, topic and intent detection, user sentiment/emotion/opinion detection, and knowledge/ commonsense reasoning.
	
	%\noindent $\bullet$ \textbf{Consistency}: 
	\paragraph{Consistency} 
	In order to gain user's long-term confidence and trust, it is crucial for a dialog system to present consistent behaviors and respond consistently given user's input and dialog history \cite{Li2016_ACL-persona,Qian2017AssigningProfile,zheng2019personalized,zhou2018design}.
	%%%personaldialog
	%and background. 
	For instance, a social bot should not deliver a response that conflicts with her pre-set persona, or her previous responses in temporal dependency, causality, or logic. 
	Specifically, the system's response needs to be consistent in three dimensions. First is persona consistency where the response needs to fit the pre-defined personality of the dialog system. 
	Second is stylistic consistency where a consistent speaking style is presented. Third is contextual consistency in which the response needs to be coherent and consistent with respect to the dialog context. 
	%Undoubtedly, consistency is also related to semantic understanding. However, this issue is more about the relationship between a system's behavior and other factors that the system should conform to. 
	From the technical perspective, consistency mainly involves personalization, stylistic generation, and multi-turn context modeling. %, and dialog planning. 
	
	%\noindent $\bullet$ \textbf{Interactiveness}: 
	\paragraph{Interactiveness} 
	As mentioned above, meeting user's social needs, such as emotional affection and social belonging, is the primary design goal of an open-domain dialog system. 
	% such as making entertainment, or seeking for suggestions or emotional comfort.
	Interactiveness refers to \emph{the system's ability to achieve complex social goals such as entertainment and conforming by optimizing its behaviors and dialog strategies in multi-turn conversation}.
	% Therefoti-turn co}nversationsre, it is essential to deliver more interactive conversations and make the user more engaged. 
	To improve interactiveness, it is important to understand the user's emotion state or affect \cite{Zhou2018EmotionalCM, zhou2018design}, to respond not only reactively but also proactively \cite{Yu2016strategy,askquestion18,rao2018learning}, to control the topic maintenance or transition \cite{wang2018chat}, and to optimize the interaction strategy (i.e., dialog policy) in multi-turn conversations to maximize long-term user engagement.
	From the technical perspective, interactiveness mainly involves sentiment and emotion detection, dialog state tracking, topic detection and recommendation, dialog policy learning, and controllable response generation.
	% sentiment and emotion detection, context modeling, topic detection and recommendation, dialog planning, and dialog policy learning.
	% by jointly optimizing the behaviors, topics, and other strategies in multi-turn conversations.

	%%\textcolor{blue}{
	%%%%增加讨论单个模型和系统之间的联系和差别
	We summarize the techniques required to address the three issues in Figure \ref{fig:overview}, including
	%but we will not discuss all of them in this paper. Instead, we will make this paper more focused and self-consistent on the models and methods for addressing the three issues. Nevertheless, the techniques 
	named entity recognition, entity linking, domain/topic/intent detection, and sentiment/emotion detection.
	%are vital for building an intelligent dialog system. In particular, 
	As demonstrated in the Alexa Prize challenge which targets at developing dialog systems for conversing coherently and engagingly with humans on various popular topics, the winning dialog systems \cite{fang2017sounding, chen2018gunrock} are composed of different modules that are developed based on these techniques, including language understanding, dialog management, and natural language generation. 
	In such modular designs, 
	the semantic issue is mainly related to the understanding module which is intended to understand the dialog (e.g., content, entity, topic, etc.) and user (e.g., opinion, personality, emotional needs). The other two issues are mainly related to the dialog management and generation modules, aiming to generate responses that are not only consistent in content and personality, but also interactive so as to increase the long-term user engagement. 
	%For consistency, we mainly focus on \emph{consistency in persona, style, and context}, because such issues are commonly observed in existing dialog models, particularly with multi-turn interactions. 
	%This issue also elicits cross-discipline research from psychology and social science, e.g., studying how to embody a consistent personality via language expression. We believe that such research is crucial for building human-like intelligent dialog systems. 
	%The interactiveness issue is mainly about the \emph{behaviors and strategies of a dialog system} to accomplish long-term, complex goals such as providing emotional comfort, or even psychological counseling \cite{althoff2016large,perez2017understanding,zhang2019finding,cao2019observing}.
	%In these scenarios, a dialog system should optimize its strategy in multi-turn interactions by comprehensively considering emotional interaction, topic maintenance or transition, proactive (question-asking, invitation, or clarification) or reactive behaviors, and so on.
	% Notice that
	These issues are highly interleaved. For example, understanding dialog and user (semantics) is fundamental to generating consistent and interactive responses.
	\section{Frameworks for Building Open-domain Dialog Systems}
	\label{sec:frameworks}
	
	As discussed in Section 1.1, open-domain dialog systems are typically implemented using an end-to-end architecture, rather than a modular architecture used by task-oriented bots for which task-specific schemas and labels are available for the development of these dialog modules.
	At the heart of an open-domain dialog system is a response generation engine, which takes user input at $t$-th dialog turn $X_t=x_1^tx_2^t\cdots x_n^t$ and dialog context $C_t$, which will be explained in a minute, and generates response $Y_t=y_1^t y_2^t\cdots y_m^t$ as
	
	\begin{equation}
	\hat{Y}_t = \mathop{\arg\max}_{Y \in \Omega} \mathcal{P}_\theta (Y|X_t,C_t)
	\label{eqn:response-generation}
	\end{equation}
	%%%%%
	where $\Omega$ denotes the set of all candidate responses, $\mathcal{P}_\theta$ is a learned model of scoring candidate responses, parameterized by $\theta$, and argmax the search algorithm to find among all candidates the best one with the highest score.
	%So, the central task is to estimate the conditional probability $\mathcal{P}(Y|X_t,C_t)$. 
	
	%In this section, we will give a high-level formulation of open-domain dialog systems. As aforementioned, it is very challenging to design specific, pipelined  components in open-domain dialog systems as in task-oriented dialog systems. Firstly, it is difficult to define the intent types for natural language understanding and not affordable to employ heavy annotation for supervised intent classification because open-domain systems have to deal with unconstrained domains and topics. Secondly, dialog state tracking and policy learning are also much more complicated because the number of possible states and dialog acts is dramatically larger than that in task-oriented dialog systems. Lastly, traditional template-based language generation is not applicable as the templates can not be exhausted in open domains. 
	
	%Thus, an end-to-end solution is a better choice than an pipelined solution. Formally, given the current input $X_t=x_1^tx_2^t\cdots x_n^t$ from the user and a context variable $C_t$, the output response $Y_t=y_1^t y_2^t\cdots y_m^t$ can be obtained by the following equation: 
	
	%\begin{equation}
	%    \hat{Y}_t = \mathop{\arg\max}_{Y \in \Omega} \mathcal{P}(Y|X_t,C_t)
	%\end{equation}
	%%%%%
	%where $\Omega$ denotes the possible search space and $t$ is the current dialog turn. So, the central task is to estimate the conditional probability $\mathcal{P}(Y|X_t,C_t)$. 
	
	This formulation unifies three typical methods of building open-domain dialog systems: retrieval-based, generation-based, and hybrid. 
	In retrieval-based methods, the search space $\Omega$ is obtained by retrieving candidate responses from a pre-collected human conversational dataset consisting of input-context-response pairs.
	% al system and filtered with coarse filtering heuristics. 
	$\mathcal{P}_\theta (Y|X_t,C_t)$ is implemented as a matching or ranking function which scores the relevance of each candidate given $X_t$ and $C_t$. 
	In generation-based methods, the search space $\Omega$ is very large, namely $Y \in V^m$ where $V$ is the vocabulary size and $m$ is the response length, and $\mathcal{P}_\theta (Y|X_t,C_t)$ is typically implemented as an auto-regressive model that generates a sentence word by word. 
	In the hybrid methods, it is typical to first retrieve \emph{prototype} responses from a dataset and then generates a response by utilizing prototype responses.
	% encode the outputs to facilitate generating a new response. 
	
	Note that the introduction of context $C_t$ offers a lot of flexibility to model various aspects of dialog. For instance, when $C_t=\varnothing$, it models single-turn dialog; Setting $C_t=X_1Y_1X_2Y_2\cdots X_{t-1}$ models multi-turn dialogs. $C_t$ can also encode other (non-content) contexts such as persona \cite{Qian2017AssigningProfile,Zhang2018Personalizing-dogpet,zheng2019personalized} for personalized dialog generation, emotion labels \cite{Zhou2018EmotionalCM,asghar2018affective} for emotional response generation, and knowledge graphs \cite{Zhou2018Commonsense,ghazvininejad2017knowledge} for knowledge-aware response generation.

	\subsection{Retrieval-based Methods}
	%%%画一个图：第一个系统流程图--柯沛
	%%%第二个是 深交互、浅交互流程图--柯沛
	% 原图在survey_image.pptx, 已上传
	
	\begin{figure}[!htp]
		\centering
		\includegraphics[width=0.7\linewidth]{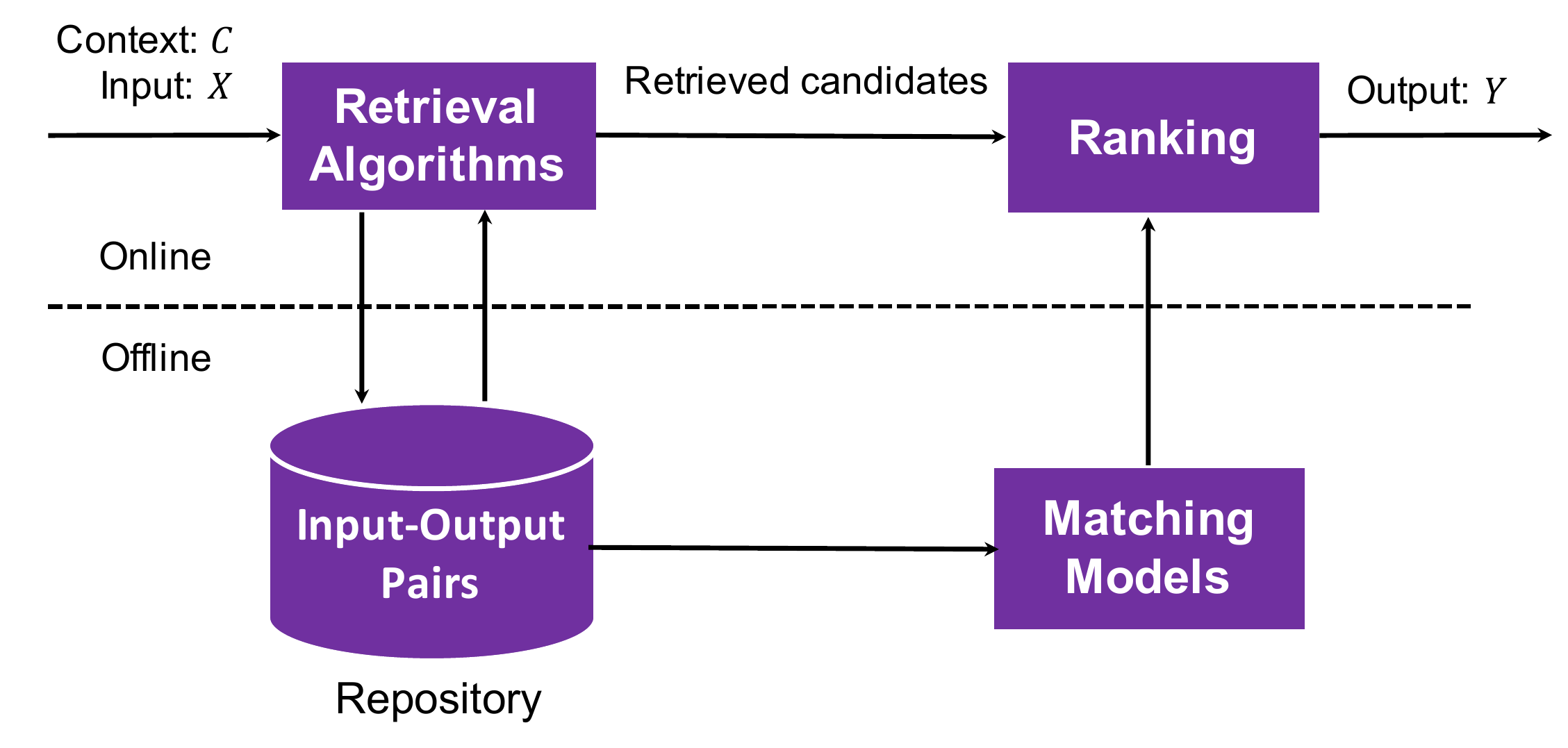}
		\caption{Framework of retrieval-based methods. The online process finds the most relevant output from the retrieved candidate with a matching model while the offline process trains the matching model with the auto-constructed data.}
		\label{fig:retrievalframework}
	\end{figure} 
	%%%Index of 去掉
	%%%matching models 字体大小不一样？

	%%%图非常好：vector的两个圆点 用蓝色跟Fea Ext的背景色 一致
	
	Given a dialog corpus and the user's post, IR-based systems can use any retrieval algorithm to choose an appropriate response from the corpus~\cite{jafarpour2010filter,carpenter2011cleverbot,leuski2011npceditor}. In such a setting, the system retrieves the most similar post to the given user post, and the response to the retrieved post is returned as the response to the user's post.
	%%%%
	Traditional learning-to-rank methods were introduced by \citet{ji2014information} for response selection from a large-scale post-response repository. Afterwards, many neural models have been proposed. 
	Figure \ref{fig:retrievalframework} illustrates the process of retrieval-based response generation methods. 
	Using input $X \oplus C$ \footnote{Hereafter, we will use $X \oplus C$ to denote the input query that combines the current user input $X$ and the dialog context $C$.}
	as a query, such methods first retrieve a list of candidates from a large repository which consists of input-context-output pairs, and choose the top-scored candidate as output response $Y$ using the matching function $\mathcal{P}_\theta (Y|X,C)$, which can be implemented using either traditional learning-to-rank algorithms \cite{liutieyan2009l2r}, or modern neural matching models \cite{lu2013deep,Huang:2013fz,fan2017matchzoo}. The model parameters $\theta$ is commonly learned by minimizing the margin-based pair-wise ranking loss as follows\footnote{
		Note that the method of pair-wise ranking is widely used in the literature, but other ways such as point-wise and list-wise ranking methods \cite{liutieyan2009l2r} are also feasible.}:
	\begin{equation}
	\mathcal{L} = max(0, \gamma + match_\theta (Y_-,X \oplus C) - match_\theta (Y_+,X \oplus C) )
	\end{equation}
	%%%%****
	where $\gamma$ is a margin (a hyper-parameter), $Y_+$ is a ground-truth (positive) response, $Y_-$ is a negative response which can be randomly sampled from the dataset or generated by corrupting $Y_+$, and $match_{\theta}(Y,X \oplus C)$ is the matching function to be learned.
	%a machine learning model that is scores high for $Y_+$ and low for $Y_-$ given input $X \oplus C$.
	
	Alternatively, we can also use a likelihood loss defined as:
	
	\begin{equation}
	\begin{aligned}
	\mathcal{L} &= - \log \mathcal{P}_\theta(Y_+|X \oplus C) \\
	\mathcal{P}(Y_+|X \oplus C)  &= \frac{\exp\{match_\theta(Y_+,X \oplus C)\} } {\exp\{match_\theta(Y_+,X \oplus C)\}+\sum_{i=1}^k {\exp \{match_\theta(Y_-^i,X \oplus C)\} } } \\
	\end{aligned}
	\end{equation}
	
	%%%%Although both loss functions are widely used, in our experiments we find that the likelihood loss works better than the margin-based loss for response ranking. There are two possible interpretations. First, the hyper-parameter $\gamma$ is difficult to tweak. Second, in the cases where there are highly competitive negative examples, the margin-based loss is close to zero, thereby leading to very little (slow) model update.
	%no gradient will be derived. 
	%%%The likelihood loss does not suffer from these issues. % there are no such issues.  
	
	% Existing models for learning $match(Y,X \oplus C)$ can be roughly classified into two types: one is deep interaction networks, and tclassificationhe other is shallow interaction networks, as illustrated in Figure \ref{}. In deep interaction networks, candidate $Y$ and input $X \oplus C$ have interactions at low-level layers before computing the representation for the  layer, while in shallow interaction networks, $Y$ and $X \oplus C$ are independently represented before being concatenated in the classification layer. 
	
\begin{figure}[!htp]
		\centering
		\includegraphics[width=1.0\linewidth]{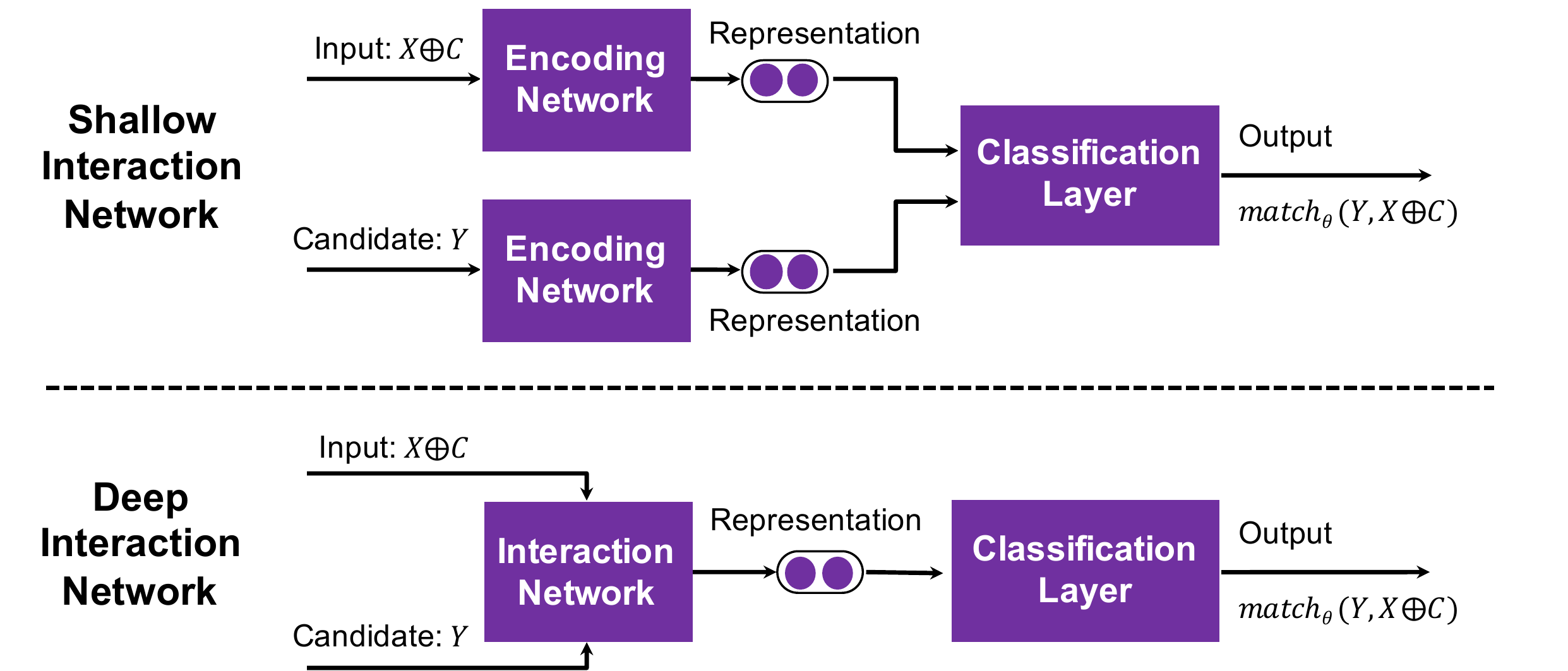}
		\caption{Frameworks of shallow and deep interaction networks. In shallow interaction network, the feature vectors of input $X \oplus C$ and candidate $Y$ are obtained independently, and there may be shallow interactions such as subtraction or element-wise multiplication between the two vectors before the classification layer. In deep interaction network, the input and candidate make interactions in the early stage to obtain a feature vector for the classification layer. }
		\label{fig:deepshallow}
	\end{figure}

	% Beyond many traditional machine learning models, recent 
	Modern neural models of $match(Y,X \oplus C)$ can be roughly grouped into two categories, shallow and deep interaction networks\footnote{Shallow or deep is regarding \emph{interaction}, namely whether the learned representations are obtained by early-stage interactions (deep), or late-stage (sometimes no) interactions (shallow). The two words are not referring to whether the model structure is deep or not.}, 
	as illustrated in Figure \ref{fig:deepshallow}. In shallow interaction networks, candidate $Y$ and input $X \oplus C$ are first encoded independently into the two vectors which then have some \emph{shallow} interactions such as subtraction or element-wise multiplication before being fed to the classification layer. In deep interaction networks, $Y$ and $X \oplus C$ interact via an interaction network to form a fused representation, which is then fed to the classification layer.
	%at low- and high-level layers, and a fused representation is generated and then fed to the classification layer.
	
	%%%%%下面的几个工作，按照深交互和浅交互重写，请朱祺【？？？】
	%%%%缩短，按照我们的思路重新，如果不重要文章引用 去掉；
	% 朱祺：我觉得应该先介绍浅交互，再介绍深交互，这样与出现的时间顺序一致

	%%%%hml-todo---修改这个图 ---done

	For shallow interaction networks, many efforts have been devoted to learning good representations for query and candidate independently. 
	\citet{Huang:2013fz} proposed to use deep structured similarity models (DSSMs) to extract semantic features from query and document independently before computing their relevance. DSSM is further augmented by introducing Convolutional layers \cite{Shen:2014id,gao2014modeling,arc,Severyn:2015gm} and recurrent layers with Long Short-Term Memory (LSTM) units \cite{palangi2015deep}. 
	To effectively incorporate dialog history, \citet{DL2R} reformulated input query $X$, and combined matching scores computed based on the reformulated and original queries, and retrieved queries and responses, respectively. 
	\citet{zhou2016multi} used a hierarchical Recurrent Neural Network (RNN) to encode a candidate and the utterance sequence in context, respectively, before computing their matching score. 
	%between a candidate and the utterance sequence in context. 
	These shallow models are simple to implement and efficient to execute.
	%in architecture, but efficient in computation.
	%%%%hml：以上这些是典型的工作吗？值得说的吗？
	
	For deep interaction networks, query $X \oplus C$ and response $Y$ interact via a neural network to generate a single feature vector that preserves all query-response interaction information at different levels of abstraction. The matching score is then derived from the vector using another neural network. 
	%which preserves matching information from low-level to high-level interactions. 
	% DeepMatch \cite{lu2013deep} used topic models of different resolutions %%%%hml topic models???
	% to extract matching features at different levels and used a DNN to calculate matching score based on these features. 
	%%%%
	\citet{arc} extracted matching features from all $n$-gram combinations of input $X$ and response $Y$ to obtain low-level feature maps with a Convolutional Neural Network (CNN). Afterwards, the feature maps are transformed with multiple CNN layers to form the final representation for classification.
	\citet{wu2017sequential} proposed a sequential matching network (SMN) for multi-turn dialog where each contextual utterance in $X \oplus C$ is encoded conditioned on $Y$, and these utterances are connected sequentially by GRUs. The matching score is computed on top of the weighted sum of the GRUs' states.  %%%%\textcolor{blue}{
	\citet{zhou-etal-2018-multi} proposed a deep attention matching network. The query and its candidate response are firstly represented with self-attention inspired by the transformer network \cite{vaswani2017attention}, and then the interactions between them were made with cross-attention to obtain word-by-word matching matrices, and finally the matching score is computed by aggregating all the matching information with a 3D matching tensor.
	\citet{yang2018response} extended SMN with external
	knowledge in information-seeking conversation systems. The method first expands response candidates using pseudo-relevance feedback, and then makes the candidates interact with the query to obtain word-by-word matching matrices. The subsequent operations are very similar to SMN. 
	\citet{zhang2018modeling} proposed a deep utterance aggregation model which shares a similar structure with SMN. The difference lies in that gated self-attention was used to obtain the representations of the query and a response candidate, and the subsequent operations are almost the same to SMN.
	\citet{wu2018response} proposed to consider topic clues for query-response matching. The authors first extracted topical words for the query and response respectively using LDA. Then, a query representation is conditioned not only on the response representation but also on the attentive read of the topical words of the response. A response representation is computed similarly conditioned on the message's topical words and the query representation.  
	%%%%}
	%Note that there are also noticeable 
	Other matching models that were proposed originally for non-dialog tasks such as paraphrase detection, language inference, and reading comprehension \cite{Wang2017bimpm,pang2016pyramid}, have also been adapted and applied to dialog response ranking.
	%which are applicable here with slight modification.
	
	%%%%jianfeng: this paragraph is newly added
	One of the most notable deep interaction networks for learning the matching function (as defined by Eq. 2) is BERT \cite{devlin2018bert}, which achieves state-of-the-art performance on many NLP tasks, including response selection. 
	$X_t \oplus C_t$ and a candidate response $y$, normally separated by a special token [SEP], form the input of a multi-layer Transformer \cite{vaswani2017attention} blocks (12-48 blocks). Each block consists of multi-head a self-attention module, layer normalization, a feed forward layer, and residual connections. The vectors at the output layer are fed to a fine-tuned classifier to determine whether the response $y$ is appropriate for the input. This structure has been widely adopted in retrieval-based methods \cite{henderson2019convert}. 
	
	%%%%\textcolor{blue}{
	There is a short review on deep retrieval-based dialogue systems \cite{boussaha2019deep} where the authors discussed existing work with respect to single-turn matching models, multi-turn matching models, and ensemble models. In comparison, we summarize existing work from the interaction perspective: whether a candidate response makes deep matching with the input (post, or along with the context) at early or late stage. In general, deep interaction networks usually work better than shallow interaction networks \cite{tao2019multi}.
	%%%%}
	
	%%%%%%%%hml：以上这些是典型的工作吗？值得说的吗？
	%%%%%%%%hml: SMN、DSSM、这些简写要注意！
	%%%%%%%%nyl: 添加了Pyramid、BiMPM两个深交互模型

	%%%%%%%生成方面的工作
	\subsection{Generation-based Methods}
	
	Neural generative models have been widely applied to open-domain dialog generation. Inspired by the early template-based generation method~\cite{a32} and statistical machine translation (SMT) ~\cite{a30}, sequence-to-sequence (Seq2seq) models~\cite{2014sequence,vinyals2015neural,shang-2015-NRM,sordoni2015neural} have become the most popular choice for dialog generation.
	Other frameworks, including conditional variational autoencoder (CVAE) \cite{serban2017hierarchical-1,zhao2017cvae,ke2018senfunc,shen18cvae,zhao18discrete,du18vae} and generative adversarial network (GAN) \cite{li2017adversarial,xu2018dpgan}, are also applied to dialog generation. {
		Very recently, Transformer-based language models pretrained with large-scale corpora are another popular choice \cite{radford2018improving,wolf2018transfer,golovanov-etal-2019-large,zhang2019dialogpt}, which obtains strong performance in dialog generation \cite{wolf2018transfer}. 
	}
	%%%这里要不要把cvae 统一到encoder-decoder的框架上来？
	
	Generation-based models usually formulate $\mathcal{P}(Y|X_t \oplus C_t)$ as:
	\begin{equation}
	\mathcal{P}(Y|X_t \oplus C_t) = \prod_{i=1}^m P(y_i|y_{<i};X_t \oplus C_t).
	\end{equation}
	where $y_{<i}=y_1y_2 \cdots y_{i-1}$.
	%%%%
	Typically, the output response is generated word by word, e.g., at each time step a word is sampled according to $P(y|y_{<i};X_t \oplus C_t)$. % which is usually expressed by a recurrent neural network. 
	Using RNNs, during the course of generation, the generated prefix is autoregressively encoded into the input to generate the next word.
	
	Most neural generation models adopt an encoder-decoder framework. The encoder transforms the input $X_t \oplus C_t$ into semantic vectors as 
	\begin{equation}
	\mathbf{X}_t \oplus \mathbf{C}_t = \textbf{Encoder}(X_t \oplus C_t).
	\end{equation}
	
	Then, at each $i$-th step of generation, the decoder updates its state vector $\textbf{s}_i$ and samples a word from distribution $\mathbf{o}_i$ as follows:
	\begin{equation}
	\begin{aligned}
	y_i \sim \textbf{o}_i &= P(y|y_{<i}; X_t \oplus C_t) \\
	&=\text{softmax} (\textbf{W}_o \textbf{s}_i) \\
	\end{aligned}
	\end{equation}
	%%%%
	where $\mathbf{W}_o$ is the weight matrix of the decoder.
	The decoder's state is updated by
	\begin{equation}
	\textbf{s}_i = \textbf{Decoder}(\mathbf{s}_{i-1},[\textbf{Att}( \textbf{X}_t \oplus \textbf{C}_t; \textbf{s}_{i-1});\textbf{y}_{i-1}])
	\end{equation}
	where $\mathbf{Att}( \mathbf{X}_t \oplus \mathbf{C}_t;\mathbf{s}_{i-1})$ is an attentive read of the encoded input conditioned on state $\mathbf{s}_{i-1}$, typically using attention mechanism \cite{Bahdanau2015Neural};
	%%%%对话中注意力机制最应该cite那个论文？
	and $\mathbf{y}_{i-1}$ is the vector representation of the previously generated word $y_{i-1}$.  
	
	%The encoder is responsible for obtaining the vector representation of the input:
	%\begin{equation}
	%\textbf{X}_t,\textbf{C}_t = \textbf{Encoder}(X_t, C_t).
	%\end{equation}
	%Note that this is a highly abstractive formulation, and $\textbf{X}_t$/$\textbf{C}_t$ should be viewed as a matrix respectively. For instance, an input sequence $X_t=x_1^t x_2^t \cdots x_n^t$ can be transformed into a matrix $\textbf{X}_t=\textbf{h}_1^t \textbf{h}_2^t \cdots \textbf{h}_n^t$ using a recurrent neural network where each $\textbf{h}_j^t$ is the hidden state at position $j$. 
	%%%%下面这句话请柯沛帮忙确认一下
	% bert的encoder也是用的transformer
	%The encoder function can also adopt more fancy encoding schemes such as Transformer \cite{vaswani2017attention}. 
	%pretraining BERT \cite{devlin2018bert}.
	
	%%%%%这个地方开始可以讲几句最新non-autoregressive的趋势？柯沛
	The formulation of generation-based models mentioned above is auto-regressive in that these models generate a target sequence word by word, each word conditioned on the words that are previously generated. To make the decoding parallelizable, non-autoregressive models based on Transformer have been proposed to generate all the tokens simultaneously \cite{kaiser2018lt,lee2018refine}. 
	Non-autoregressive modeling factorizes the distribution over a target sequence given a query into a product of conditionally independent per-step distributions, as follows:
	% In this setting, $\mathcal{P}(Y|X_t,C_t)$ can be formulated as follows:
	\begin{equation}
	\mathcal{P}(Y|X_t \oplus C_t) = \prod_{i=1}^m P(y_i|X_t \oplus C_t).
	\end{equation}
	Though the performance of such non-autoregressive models is still not as good as their autoregressive counterparts, it opens new opportunities for fast training using very large scale datasets \cite{gu2017NAD,lee2018refine}. 
	%might be a new direction for language generation.

	\begin{figure}[!htp]
		\centering
		\includegraphics[width=0.8\linewidth]{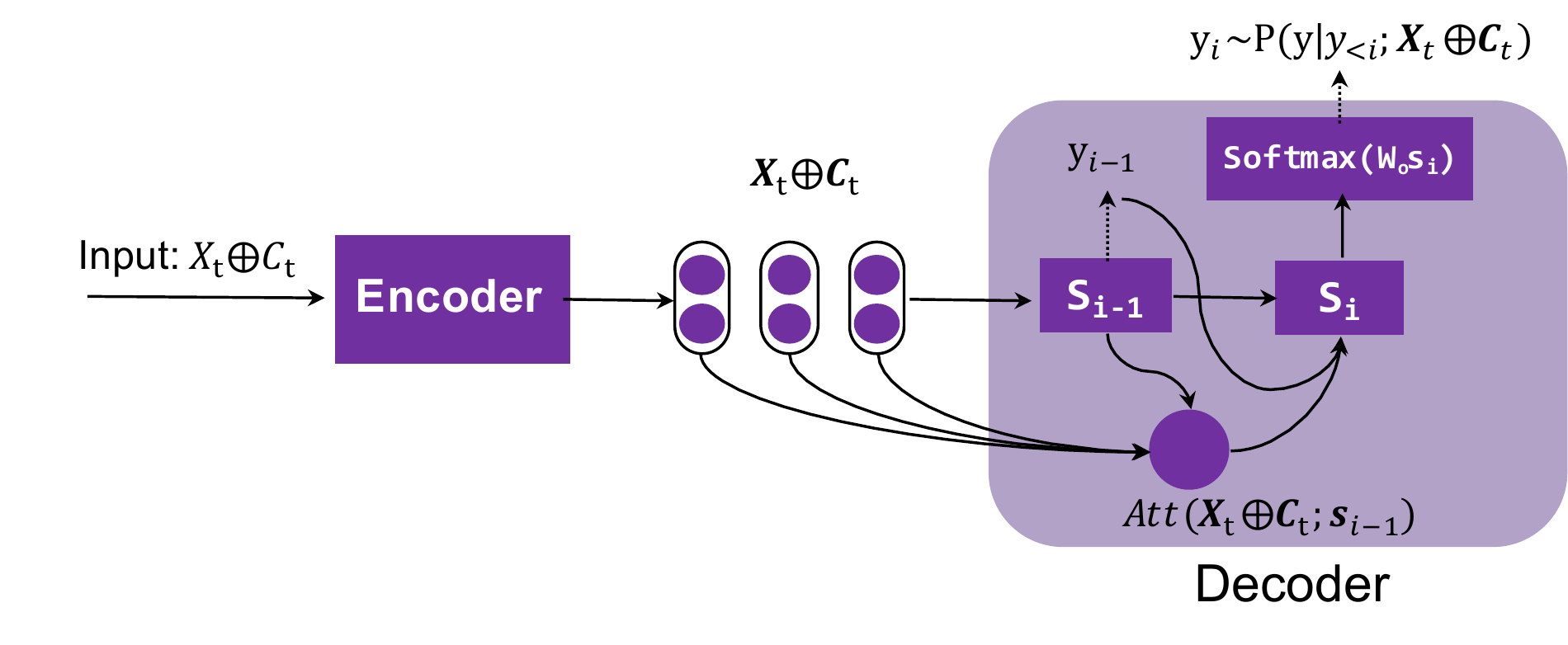}
		\caption{Typical encoder-decoder framework for generation-based models. The input $X_t \oplus C_t$ is encoded into vectors $\textbf{X}_t \oplus \textbf{C}_t$. In the decoder, a word $y_i$ is sampled from $P(y|y_{<i},X_t \oplus C_t)=softmax(\textbf{W}_o\textbf{s}_i)$ and the decoder's state is updated with $y_{i-1}$ and $\textbf{Att}( \textbf{X}_t \oplus \textbf{C}_t;\textbf{s}_{i-1})$ as input.}
		\label{fig:encoder-dec}
	\end{figure} 

%%%%jianfeng: this paragraph is added 	
	Noticeably, the large-scale pre-trained models, such as BERT and GPT-2 \cite{devlin2018bert, radford2018improving}, can be easily applied in the above encoder-decoder framework. The encoder can be a pre-trained BERT model or a GPT-2 model, the decoder a GPT-2 model.  Both the parameters of the encoder and the decoder are initialized using the pre-trained models and then fine-tuned on a dialog corpus \cite{radford2018improving,wolf2018transfer,golovanov-etal-2019-large,zhang2019dialogpt,adiwardana2020towards}. The fine-tuning process is often tailored to the dialog scenario via encoding with dialog state embeddings\cite{wolf2018transfer}, classifying golden and negatively sampled responses given the same dialog context\cite{golovanov-etal-2019-large}, designing dialog-specific pre-training tasks \cite{mehri2019pretraining,budzianowski2019hello}, and so on. These models have shown strong performance in the NeurIPS Conversational Intelligence Challenge 2 (ConvAI 2)\footnote{http://convai.io/} and were used in the TREC Conversational Assistance Track (Conversational Information Seeking)\footnote{http://www.treccast.ai/}.  Notably, \citet{zhang2019dialogpt} released the DialoGPT model that was trained on 147M conversation-like exchanges extracted from on Reddit comment threads, providing a good starting point for future research.
	
	%%%%%CVAE 是否可以统一在上面的框架上？
	
	% In the following sections, we will survey many generation models from the perspectives of improving semantics, consistency, and interactiveness. 

	%%%%%%混合方法的典型和值得说的工作；朱祺
	\subsection{Hybrid Methods}
	Retrieval-based methods retrieve an output response from a repository of human-human conversations. Such human-produced conversations are fluent, grammatical, and of high quality. However, the scale of the repository is critical to the success of the methods, which unfortunately is never large enough for open-domain dialog systems. 
	Moreover, retrieval-based methods cannot generate unseen responses. 
	On the other hand, generation-based methods can produce novel responses. 
	But they often generate undesirable responses that are either ungrammatical or irrelevant.   
	Hybrid methods combine the strengths of both and usually adopt a two-stage procedure \cite{yang2019hybrid,weston2018retrieve,zhou2018design}. 
	In the first stage, some relevant conversations, known as prototype responses in \cite{wu2019prototype}, are retrieved from a dataset using input $X \oplus C$ as a query. 
	Then, prototype responses are used to help generate new responses in the second stage.
	
	%encode the relevant conversations to facilitate generating a new response with generation-based models. The basic assumption is that if more relevant information is encoded, the generation model would generate better conversations than simply generating from a limited input $X \oplus C$.  
	
	% \cite{Song:2018ve,Contractor:2018vl,wu2019prototype,Weston18refine}
	% An Ensemble of Retrieval-Based and Generation-Based Human-Computer Conversation Systems
	% Exemplar Encoder-Decoder for Neural Conversation Generation
	% Response Generation by Context-aware Prototype Editing
	
	%%%下面这些工作，可以如果比较典型的可以保留，否则可以去掉
	%  \cite{a40}.   是retrieval的方法
	% \cite{a41}. 无关
	% ~\cite{DL2R}. 是retrieval的方法
	% ~\cite{qiu2017alime,song2016two} 第一个是用生成模型给检索结果打分，比较糙，第二个放2018年的改进版
	Based on the Seq2Seq architecture, \citet{Song:2018ve} 
	used additional encoders to represent the set of retrieved responses, and applied the attention \cite{Bahdanau2015Neural} and copy \cite{gu2016copy} mechanism in decoding to generate new responses. \citet{Contractor:2018vl} first retrieved similar conversations from training data using a TF-IDF model. The retrieved responses were used to create exemplar vectors that were used by the decoder to generate a new response. 
	%the top $k$ similar context-response pairs. The new response is then generated from each retrieved response and the input context, where the loss of generating from each retrieved response is weighted the similarity between the corresponding retrieved context and the input context. 
	\citet{wu2019prototype} first retrieved a prototype response from training data and then edited the
	prototype response according to the differences between the prototype context and current context. The motivation is that the retrieved prototype provides a good start-point for generation because it is grammatical and informative, and the post-editing process further improves the relevance and coherence of the prototype.
	\citet{zhang2019ensemblegan} proposed
	an adversarial learning framework to enhance a retrieval-generation ensemble model. Their model consists of a language-model-like generator, a ranker generator, and a ranker discriminator. This model encourages the two generators to generate responses that are scored higher by the discriminative ranker, while the discriminator down-weighs adversarial samples 
	and selects those responses that are favored by the two generators.
	
	%a most similar context-response pair is first retrieved, and a insertion word set and a deletion word set are obtained by comparing the retrieved context and the input. The response is then generated by attending to the retrieved response and the two word sets (so-called edit vectors).

	%\subsection{Generation-based Methods}
	%%%讲一下起源于SMT， ---》RL 
	%% 建议和2.2 合并,概述部分已添加至2.2的开始

	%\subsection{Hybrid Methods}

	%\section{Key Issues} 

	%%%%柯沛，是否可以补充cvae是如何提高多样性的？
	
	% 提升多样性这段也可以分为以下三个方面来叙述
	% (1) Encoder improvement: 通过编码更多信息来提升回复质量，可直接用现在的encoding more information这一节
	
	% (2) Representation improvement: 基于对话一对多的本质来修改对话的表示，从而生成多样的回复。主要工作有以下两篇：
	% Mechanism-Aware Neural Machine for Dialogue Response Generation. 这篇引入diverter和mechanism embedding来学含有类别信息的对话表示
	% Learning Discourse-level Diversity for Neural Dialog Models using Conditional Variational Autoencoders. 这篇引入CVAE将对话的表示建模为高斯分布，通过从分布中采样来生成带有不同对话意图的回复
	
	% (3) Decoder improvement: 该节可分为两部分，即probability bias和post-processing

	%\subsection{Improving Semantics by Modeling Diversity and Informativeness}
	\section{Semantics}
	\label{sec:semantics}
	
	A typical symptom of a dialog system that suffers from the semantics issue is that it often generates bland and generic responses, such as ``I don't know'', ``thank you'', ``OK'' , or simply repeats whatever a user says ~\cite{sordoni2015neural,vinyals2015neural,hiContext,gao2019neural}. 
	We observe similar phenomena in human conversations. When we don't understand what the other party is talking about but have to respond, we often pick those safe but bland responses.
	%like ``OK'' and ``I don't know''. 
	
	To make an engaging conversation, the dialog system needs to produce contentful, interesting, and interpersonal responses based on its understanding of the dialog content, user's sentiment and emotion, and real-world knowledge that is related to the dialog.
	In this section, we review some of the most prominent neural approaches that have been proposed recently to address the semantics issue. We first describe the ways of improving the encoder-decoder framework to generate diverse and informative responses by improving the understanding (embedding) of dialog context and users. Then, we describe the methods of grounding dialog in real-world knowledge to make system responses more contentful. 
	
	%The semantics issues is well exposed in the quality of generated responses: existing models are still far from real understanding, and thus constantly produce improper responses. The most observable phenomena is that these models tend to output generic, meaningless responses, such as ``I don't know'', ``Me too'' or ``I'm OK''~\cite{contextConv,vinyals2015neural,hiContext}. Issues including response blandness and word repetition are discussed in \cite{gao2019neural} with more details.
	
	\subsection{Improving Diversity and Informativeness in Neural Response Generation}
	
	% Essentially, these quality issues are originated from the principle of maximum likelihood estimation, naturally causing that common words have larger generation probabilities. Such responses may lead the conversation to break down. Since most existing models that devoted to improving diversity and informativeness are within the encoder-decoder framework, we summarize these approaches of improving generation quality from four aspects: encoder, intermediate representation, decoder, and post-processing.
	
	Most state of the art neural response generation models are based on the encoder-decoder framework which consists of four components: (1) an encoder that encodes user input and dialog context, (2) an intermediate representation, (3) an decoder that generates candidate responses, and (4) a ranker that picks the best candidate as the response. In what follows, we review the proposed methods in four categories, each  focusing on improving one of the four components. 
	
	% $\bullet$ \noindent \textbf{Encoder}. 
	\paragraph{Encoder} 
	Encoding richer information from query $X \oplus C$, such as longer dialog history \cite{sordoni2015neural}, persona \cite{Li2016_ACL-persona}, hidden topics \cite{serban2017hierarchical-1}, has proved to be helpful for generating more informative responses. \citet{xing2017topic} extracted topic words, rather than hidden topic vectors, using LDA, and encoded such words in a topic-aware model. The model generates a response by jointly attending to query $X \oplus C$ and the topic words. 
	Topic words are also used to model topic transition 
	%or maintenance 
	in multi-turn conversations \cite{wang2018chat}. 
	The hybrid methods described in Section 2.3 \cite{Contractor:2018vl,Song:2018ve,wu2019prototype} encode the retrieved prototype responses to help generate more informative responses. 
	% As aforementioned, encoding some relevant responses in hybrid methods \cite{Contractor:2018vl,Song:2018ve,wu2019prototype} can facilitate to generate more informative responses.
	
	% $\bullet$ \noindent \textbf{Intermediate Representation}. 
	\paragraph{Intermediate Representation} 
	Instead of encoding $X \oplus C$ using a fixed-size vector as in \cite{2014sequence}, methods have been proposed to use more flexible intermediate representations (e.g., additional latent variables) to enhance the representation capability to address the one-to-many issue in dialog, and to improve the interpretability of the representation in order to better control the response generation.
	%learning during decoding, thereby to address the one-to-many issue in dialog. 
	\citet{zhao2017cvae} introduced CVAE for dialogue generation and adopted a Gaussian distribution, rather than a fixed-size vector, as the intermediate representation, thus obtaining more diverse responses via sampling the latent variable. 
	\citet{du18vae} introduced a sequence of continuous latent variables to model response diversity, and demonstrated empirically that it is more effective than using a single latent variable. 
	\citet{zhao18discrete} proposed an unsupervised representation learning method to use discrete latent variables, instead of dense continuous ones, which improves the interpretability of representation.
	\citet{zhou2017mechanism,zhou2018elastic} assumed that there exist some latent responding mechanisms, each of which can generates different responses for a single input post. These responding mechanisms are modeled as latent embeddings, and can be used to encode the input into mechanism-aware context to generate responses with the controlled generation styles and topics. 
	\citet{gao2019jointly} proposed a SpaceFusion model which induces a latent space that fuses the two latent spaces generated by Seq2Seq and auto-encoder, respectively, in such a way that after encoding $X \oplus C$ into a vector in the space, the distance and direction from the predicted response vector given the context roughly match the relevance and diversity, respectively.
	%%%%这里还要讲一下2-3个主要工作？
	
	% $\bullet$ \noindent \textbf{Decoder}. 
	\paragraph{Decoder} 
	Assigning additional probability mass to \emph{desirable} words in decoder is a commonly used method to gain some control of what to generate. 
	%the encourage generating informative responses. 
	Mathematically, this can be implemented by adjusting the output word distribution as follows:
	\begin{equation}
	P_{new}(y_i|y_{<i};X,C) = \textbf{Normalize}(P(y_i|y_{<i};X,C) + P_{bias}(y_i|y_{<i};X,C))
	\end{equation}
	where $y_{<i}=y_1y_2\cdots y_{i-1}$ is the generated prefix; $P_{bias}$ assigns additional probability mass to the desirable words to be generated; and $\textbf{Normalize}(\cdot)$ is a normalization function to ensure a probability distribution. 
	Many existing controllable decoding methods essentially fall into this formulation. The most notable example is CopyNet \cite{gu2016copy}, which copies desirable but infrequent words from the input to the output, thus assigning higher probabilities to those words. 
	In \cite{zhang2018learning}, $P_{bias}$ is formulated as a Gaussian distribution, which assigns higher probabilities to rare words to control the specificity of a response, where the specificity score of a word is proportional to its IDF (inverse document frequency) score.
	
	% $\bullet$ \noindent \textbf{Post-processing}. 
	\paragraph{Candidate Ranker} To obtain more diverse responses, beam search is commonly used to generate multiple candidates, which (together with retrieved candidates in hybrid dialog systems) are then ranked by another model, which uses information that is not available in decoding (e.g., mutual information between input and response) or is too expensive to use in decoding (e.g., a large pre-trained language model such as BERT \cite{devlin2018bert}) to select the final response.
	\citet{lidiversity} proposed to use Maximum Mutual Information (MMI) as the objective to rank candidates to promote the diversity of generated responses.
	As the standard beam search often produces near-identical results, recent work addresses it by encouraging the diversity among (partial) hypotheses in the beam. For example, \citet{li2016simple} penalized lower-ranked siblings extended from the same parents, so that the N-best hypotheses in the beam at each time step are more likely to expand from different parents, and thus more diverse. 
	\citet{Ashwin2018DBS} divided the hypotheses into several groups and applied beam search group-by-group. The model favours the hypotheses that are dissimilar to the ones in the previous groups. 
	Constrained beam search \cite{balakrishnan2019constrained} was also proposed to generate desirable responses by constraining a generated response to obey the input structure. 
	
	% \subsection{Improving Semantics with Knowledge}
	\subsection{Knowledge Grounded Dialog Models}
	Knowledge is crucial for language understanding and generation. 
	To build effective human-machine interactions, it is indispensable to ground the concepts, entities, and relations in text in commonsense knowledge or real-world facts such as those stored in Freebase and Wikipedia.
	%or even with commonsense knowledge \cite{speer2013conceptnet} to interpret common, daily life concepts. 
	% This is particularly important for building intelligent open-domain dialog systems which aim to make the user more engaged and interactive in conversation. 
	An knowledge-grounded open-domain dialog system
	%, equipped with rich knowledge and knowledge grounding capability, 
	should be able to identify the entities and topics mentioned in user input, link them into real-world facts, retrieve related background information, and thereby respond users in a proactive way e.g., by recommending new, related topics to discuss. 
	% to enable the ability of planning topics in conversation and behaving more proactively to make users more engaged. 
	
	Knowledge has been shown useful in both retrieval-based and generation-based dialog systems. 
	A well-known example of the former is Microsoft XiaoIce \cite{zhou2018design}. XiaoIce relies on a large knowledge graph (KG) to identify the topics and knowledge related to user input for both response generation and topic management.
	In~\cite{young2018augmenting}, a Tri-LSTM model is proposed to use commonsense knowledge as external memories to facilitate the model to encode commonsense assertions for response selection. 
	An early example of using knowledge for generating responses is \cite{han2015exploiting}, where manually crafted templates are used to generate responses which are filled with relevant knowledge triples. 
	In \cite{ghazvininejad2017knowledge}, a knowledge-grounded model is proposed to generate a response by incorporating some retrieved posts that are relevant to the input. 
	%The knowledge used by \citet{ghazvininejad2017knowledge} is in the form of unstructured posts retrieved by an information retrieval model, and 
	However, the quality of these unstructured posts is mixed. 
	Pre-compiled structured knowledge, which is in the form of fact triples, is believed to be of higher quality and has been shown to more helpful in conversation generation \cite{zhuwenya2017,liu2018knowledge}. 
	\citet{zhuwenya2017} dealt with a scenario where two speakers are conversing based on each other's private knowledge base in the music domain. The generation model can generate a word in response from the context or the knowledge base. 
	In \cite{liu2018knowledge}, a knowledge diffusion model is proposed to not only answer factoid questions based on a knowledge base, but also generate an appropriate response containing knowledge base entities that are relevant to the input.
	\citet{Zhou2018Commonsense} exploited the use of large-scale commonsense knowledge for conversation generation. First, a one-hop subgraph is retrieved from ConceptNet \cite{speer2013conceptnet} for each word in an input post. Then, the word vectors, along with the graph vectors which extend the meaning of the word via its neighboring entities and relations, are used to encode the input post. During decoding, a graph attention mechanism is applied in which the model first attends to a knowledge graph and then to a triple within each graph, and the decoder chooses a word to generate from either the graph or the common vocabulary.
	\citet{qin2019conversing} presented a new end-to-end approach that jointly models response generation and on-demand machine reading for generating contentful conversations. The key idea is to provide the model with relevant long-form text on the fly as a source of external knowledge. The model performs QA-style reading comprehension on this text in response to each conversational turn, thereby allowing for more focused integration of external knowledge than prior approaches. 
	
	%%%%---jianfeng, please check this table and the section 
	%%%%%\Jianfeng{I suggest you construct a table to compare all knowledge grounded dialog models in terms of (1) the knowledge used for grounding, e.g., \cite{ghazvininejad2017knowledge,Zhang2018Personalizing-dogpet} uses grounding in the form of independent snippets of text: Foursquare tips and background information about a given speaker. \cite{qin2019conversing} use the full text of  a web page and its underlying structure. \cite{dinan2019second, moghe2018towards} exploiting crowdsourced conversations with detailed grounding labels. (2) how the grounding is implemented, e.g., via explicit grounding label such as \cite{moghe2018towards} or methods of retrieval, attention or QA such as \cite{ghazvininejad2017knowledge,qin2019conversing}. (3) modeling techniques, such as MemNN of \cite{ghazvininejad2017knowledge} and machine reading comprehension model of \cite{qin2019conversing}. }%%%%%%----great suggestions, will do within three days

	\setlength{\rotFPtop}{0pt plus 1fil}
	\setlength{\rotFPbot}{0pt plus 1fil}
	%\begin{table}[t]
	\begin{sidewaystable}
		
		\centering
		%\scriptsize
		%\setlength\tabcolsep{3pt}
		
		\setlength{\tabcolsep}{1.0mm}
		%\begin{turn}{90}
		\begin{tabular}{lllll}
			\toprule
			
			Authors & Knowledge modality & Grounding method & Issues focused & Models \\
			%%%分别从知识形态、知识与对话的映射方法、研究问题、采用的方法本身        
			\midrule
			
			\citet{Zhou2018Commonsense} & Knowledge graph & Retrieval & Knowledge-aware generation & Seq2Seq+Graph Attention \\ \hline
			
			\citet{ghazvininejad2017knowledge} & Unstructured text & Retrieval & Knowledge-aware generation & Memory Networks \\ \hline
			
			\citet{zhuwenya2017} & Knowledge base & Retrieval & Knowledge-aware generation & \tabincell{l}{Seq2Seq+\\\ \ Knowledge Retriever} \\ \hline
			
			\citet{liu2018knowledge} & Knowledge base & Retrieval & Knowledge-aware generation & \tabincell{l}{HRED+\\\ \ Knowledge Retriever} \\ \hline

			\citet{qin2019conversing} & Unstructured text & QA & Knowledge-aware generation & SAN + Generator  \\ \hline
			
			\citet{chendykgchat} & Knowledge graph & Multi-hop reasoning & \tabincell{l}{Knoweldge-aware generation\\\ \ +Zero-shot adaptation} & \tabincell{l}{Seq2Seq+\\\ \ Multi-hop Reasoning}  \\ \hline
			
			\citet{dinan2018WOW} & Unstructured text & Retrieval & Knowledge-aware generation & Transformer  \\ \hline
			
			\citet{gopalakrishnan2019topical} & Unstructured text & Retrieval & Knowledge-aware generation & Transformer  \\ \hline
			
			\citet{moghe2018towards} & \tabincell{l}{Unstructured text+\\\ \ Fact table} & Grounding label & Knowledge-aware generation & HRED/GTTP/BiDAF  \\ \hline 
			
			\citet{moon2019opendialkg} & Knowledge graph & Grounding label & Knowledge selection & KG path decoder \\ \hline
			
			\citet{wu2019proactive} & \tabincell{l}{Unstructured text+\\\ \ Knowledge graph} & Grounding label & Proactive conversation & BERT/PostKS \\ \hline
			
			\citet{zhou2018DoG} & Unstructured text & Grounding label & Knowledge-aware generation & Seq2Seq \\ \hline
			
			\citet{lian2019learning} & Unstructured text & Grounding label & Knowledge selection & PostKS \\ \hline
			
			\citet{liu2019knowledge} & \tabincell{l}{Unstructured text+\\\ \ Knowledge graph} & QA+Grounding label & Knowledge selection & RL+BiDAF \\ \hline
			
			\citet{ren2019thinking} & Unstructured text & Grounding label & Knowledge selection & BiDAF+GTTP \\ \hline
			
			\citet{zhang2019improving} & Unstructured text & Grounding label & Knowledge selection & BiDAF+Seq2Seq \\ \hline
			
			\citet{li2019incremental} & Unstructured text & Grounding label & Knowledge-aware generation & \tabincell{l}{Incremental Transformer+\\\ \ Two-pass Decoder} \\

			%%%%%%【注意】：有些工作同时考虑selection+generation；总结的维度可以进一步讨论，如果有更好的归纳方式
			\bottomrule
		\end{tabular}
		%\end{turn}
		
		\captionsetup{width=.88\textwidth}
		\caption{Survey on existing knowledge-grounded studies. Grounding method refers to the means of a grounded knowledge linking to an utterance. Retrieval means that the grounded knowledge is retrieved based on key words in utterances. QA means the knowledge is extracted using machine reading comprehension methods. Grounding label means the knowledge used in the conversation is explicitly annotated by hand. In the last column, PostKS means selecting knowlege by mininizing the KL loss between a prior and a posterior distribution over knowledge \cite{wu2019proactive,lian2019learning};
			SAN refers to the Stochastic Answer Network for machine reading comprehension model proposed in \cite{liu2018stochastic}; GTTP (Get To The Point) refers to the hybrid pointer generator network for abstractive summarization proposed in \cite{see2017get}; HRED refers to the hierarchical neural response generation model \cite{serban2015hierarchical}; and BiDAF refers to the Bi-Directional Attention Flow network for reading comprehension \cite{seo2016bidirectional}.} 
		\label{tab:knowledge-grounded}

		%\end{table}
		
	\end{sidewaystable}

	We summarize the aforementioned knowledge-grounded dialog systems in Table \ref{tab:knowledge-grounded}. 
	Most these studies focus on two problems: (1) \emph{knowledge selection} -- selecting appropriate knowledge to be incorporated in the next response given the dialog context and previously-selected knowledge \cite{lian2019learning,ren2019thinking, zhang2019improving,liu2019knowledge}, and (2) \emph{knowledge-aware generation} -- injecting the required knowledge into a generated response \cite{ghazvininejad2017knowledge,Zhou2018Commonsense,li2019incremental,qin2019conversing}. 
	In addition, zero-shot adaptation to updated, unseen knowledge graphs without conversational data \cite{chendykgchat} is worth more comprehensive exploration in the future. %%First is mimicking the topic transition within a knowledge graph in human conversations. Second is zero-shot adaptation to updated, unseen knowledge graphs without any conversational data \cite{moon2019opendialkg,chendykgchat}. %%%%
	solving the problem would allow dialog systems to generate proper responses with selected knowledge even though the knowledge has never been used.
	%%%%%
	
	Recently, there is a significant burst in constructing document or knowledge grounded dialog corpora~\cite{zhou2018DoG,moghe2018towards,dinan2018WOW,moon2019opendialkg,chendykgchat,wu2019proactive,gopalakrishnan2019topical,qin2019conversing}, which will be described in Section \ref{sec:datasets} in details.
	
	%In human-human conversations, utterances are grounded on knowledge, which could be derived from commonsense. It is weird for a conversational system to say ``a clear sky is red'', which is contradictive to human knowledge. As a result, it is assumed that a conversational model with a ``memory look-up'' module into the knowledge base can mimic human conversations more closely.
	%Knowledge-grounded utterance generation is first investigated for Knowledge-Based Question-and-Answering (KB-QA). Yin \textit{et al.}~\shortcite{yin2016neural} built on the encoder-decoder framework equipped with the ability to enquire the knowledge-base. The model can generate right and natural answers by referring to the facts in the knowledge-base. A more comprehensive strategy is to incorporate copying and retrieving mechanisms and predicts word generation by copying from the query utterance and/or retrieving from the knowledge base jointly~\cite{he2017generating}. A Tri-LSTM model was proposed to use commonsense knowledge as external memories to facilitate LSTMs to encode commonsense assertions in order to enhance response selection~\cite{young2018augmenting}. With proper knowledge added, utterances are more grounded.% by background information in conversations.
	
	\section{Consistency}
	\label{sec:consistency}
	%%%%%\Jianfeng{You might consider re-structuring this section as follows. (1) define the scope of consistency, e.g., aim to generating responses that are consistent in persona, style, and topic/semantics. Then for each, we review recent papers. (2) methods of consistent persona can be grounded into two categories depending on whether the persona information is implicitly or explicitly model. (3) Stylistic response generation. (4) Topic/semantics consistency has not be comprehensively studied yet. Existing methods include... (5) a survey of corpora people have developed to facilitate future research.}
	
	A human-like dialog system needs to embody consistent behaviors, so that it can gain the user's confidence and trust \cite{shum2018eliza,zhou2018design}. The consistency issue refers to generating responses that are consistent in persona, style, and context (with respect to topic, logic, causality, etc.). We group existing studies into three lines: (1) persona consistency modeling including implicit and explicit methods, (2) stylistic response generation, and (3) contextual consistency. 

	\subsection{Persona Consistency}
	Existing dialog models that address persona consistency can be roughly grouped into two catetories: {\it implicit personalization} and {\it explicit personalization}.
	In the former, the persona is implicitly represented by a persona vector. For instance, 
	\citet{Kim2014Acquisition} proposed a ranking-based approach to integrate a personal knowledge base and user interests in dialogue system.
	\citet{Bang2015Example} extended the user input by exploiting examples retrieved from her personal knowledge base to help identify the candidate responses that fit her persona.
	\citet{Li2016_ACL-persona,zhang2017neural} used an embedding vector to represent a user (speaker) persona and fed the user embedding into each decoding position of the decoder. Such models need to be trained using conversational data labeled by user identifiers, which is expensive to collect for large quantities. 
	Thus, \citet{wang2017groupbias} proposed to train personalized models with only group attributes (e.g., male or female). The group attributes are embedded to vectors and then fed into the decoder for response generation. 
	\citet{zhang2019consistent} proposed a neural conversation model that generates consistent responses by maintaining certain features related to topics and personas throughout the conversation. Unlike other work that requires external supervision such as user identities, which are often unavailable, this approach trains topic and persona feature extractors in a self-supervised way by utilizing the natural structure of dialogue data.
	Although \citet{zhang2018addressee,ouchi2016addressee} showed that user embedding is an effective technique to distinguish roles of speakers and addressees in multi-party conversation, personalization in these models are handled in an implicit way and thus not easy to interpret and control in generating desired responses.
	
	In \cite{Qian2017AssigningProfile}, an explicit persona model is proposed to generate personality-coherent responses given a pre-specified user profile. The chatbot's persona is defined by a key-value table (i.e., profile) which consists of name, gender, age, hobbies, and so on. During generation, the model first chooses a key-value from the profile and then decodes a response from the chosen key-value pair forward and backward. This model can be trained on generic dialogue data without user identifier. XiaoIce also uses an explicit persona model \cite{zhou2018design}.

	%%%\textcolor{blue}{
	We have discussed two categories of methods for modeling persona consistency: \emph{implicit modeling} \cite{Li2016_ACL-persona,zhang2019consistent} which utilizes learned user persona features to capture user-level consistency implicitly, and \emph{explicit modeling} \cite{Qian2017AssigningProfile,zhou2018design} which controls the conversation generation using explicitly-defined user profile. However, most existing methods are insufficient in modeling the user's psychological personality. For instance, we do not yet have a dialog system that can exhibit extrovert or introvert personality. Building such an intelligent dialog system requires breakthroughs in multi-disciplined research on psychology, cognitive, and social science.

	%%%\subsection{Personalized Dialog Corpora}
	%%%%%因为我们后面有专门的数据介绍，所以去掉了。。。
	%%%There have been increasing efforts of building personalized dialogue corpora. In \cite{Zhang2018Personalizing-dogpet}, a multi-turn dialogue corpus is constructed, where each dialogue session involves two speakers and the persona of each speaker is defined by several sentences describing the speaker's hobbies or preferences. 
	%%%\citet{mazare2018millionsagent} presented a simple method of constructing a large-scale personalized dataset from social media where user's persona is defined by a set of sentences of particular patterns describing their preferences. 
	%%%\citet{joshi2017personalization-goaldialog} developed a personalized version of the bAbI dialoge dataset \cite{bordes2016learning} by associating each goal-oriented dialog with user traits such as gender, age, and favorite foods. 
	%%%In \cite{zheng2019personalized}, a large-scale personalized dialog corpus has been developed. The corpus consists of multi-turn conversations collected from Weibo with speaker IDs. Each speaker is associated with her personal information including gender, age, location, and interest.
	%%%\textcolor{blue}{
	%%%%\citet{wang2019persuasion} proposed a personalized corpus that consists of persuasive conversations for donation, where the corpus considers an individual's demographic
	%%and psychological backgrounds including personality, morality, value systems, and so on. This corpus is also useful for modeling dialog strategies for accomplishing long-term, complex goals such as donation persuasion. 
	%%%}

	\subsection{Stylistic Response Generation}
	Stylistic response generation \cite{wang2017steering,oraby2018controlling} can be viewed as a form of personalization in conversation. 
	There are two main challenges: how to disentangle content and style in representation, and how to construct training data containing pairs of responses that are of the same content but in different styles. 
	\citet{wang2017steering} utilized a small-scale stylistic data and proposed a topic embedding model to generate responses in specific styles and topics simultaneously.
	\citet{oraby2018controlling} demonstrated that it is possible to automate the construction of a parallel corpus where each meaning representation can be realized in different styles with controllable stylistic parameters.

	%%%%hml
	Stylistic conversation generation is closely related to domain adaptation and transfer learning \cite{zhang2017neural,mo2016personalizing,Casanueva2015Knowledge,wang2017steering}.
	The idea is to first train a general conversation model on a large corpus in source domain and then to transfer the model to a new speaker or target domain using small amounts of personalized (or stylistic) data in target domain.
	\citet{Casanueva2015Knowledge} proposed to automatically gather conversations from similar speakers to improve the performance of policy learning of personalized dialogue systems.
	\citet{zhang2017neural} proposed a two-phase transfer learning approach, namely \emph{initialization then adaptation}, to generate personalized responses. 
	They also proposed a quasi-Turing test method to evaluate the performance of the generated responses.
	\citet{npm} presented a transfer learning framework similar to~\citet{zhang2017neural}, but proposed to use a new adaptation mechanism based on reinforcement learning.
	\citet{ar1} proposed a multi-task learning approach where the response generation and utterance representation are treated as two sub-tasks for speaker role adaptation.

	\subsection{Contextual Consistency}
	
	Unlike the studies on persona consistency, the work on modeling contextual consistency is yet to be explored. Early work has focused on better representing dialog contexts \cite{serban2015hierarchical,serban2017hierarchical-1} using hierarchical models, which can be viewed as \emph{implicit} modeling of contextual consistency. Recently, \citet{welleck2018NLI} and \citet{dziri2019NLI} characterized the contextual consistency as a natural language inference
	(NLI) problem \cite{dagan2005pascal}. 
	In this setting, a response is considered consistent if it can be inferred from the dialog context or the given persona. 
	\citet{welleck2018NLI} constructed a dialog NLI dataset based on Persona-Chat\cite{Zhang2018Personalizing-dogpet}. \citet{zhang2019consistent} proposed to learn topic features from dialog context on-the-fly and utilize controllable response generation techniques to generate topic-consistent responses. 
	%This direction is worthy of further exploration.

	%%%%%%%%%%%%%%%%%%%%%%%%%%%%%%%%%%%%%%%%
	\section{Interactiveness}
	
	%%%%\textcolor{blue}{
	\label{sec:interactiveness}
	%%%%Interactiveness refers to the system's ability to generate interpersonal responses to maximize long-term user engagement. 
	This issue is mainly about how to optimize the \emph{behaviors and strategies of a dialog system} to maximize long-term user engagement and accomplish long-term, complex goals such as providing emotional comfort, or even psychological counseling \cite{althoff2016large,perez2017understanding,zhang2019finding,cao2019observing}.
	To improve interactiveness, it is important to understand user's emotion and affect, in addition to dialog content, and to optimize the system's behavior and interaction strategy in multi-turn conversations.
	%%%%}
	%to respond not only reactively but also proactively,to control the topic maintenance or transition [94], and  
	
	\subsection{Modeling User Emotion}
	%Emotional intelligence is a key intelligent behavior of human beings. Undoubtedly, 
	Emotion perception and expression is vital for building a human-like dialog system. 
	Earlier attempts to building emotional dialog systems are mostly inspired by psychology findings. Those systems are either rule-based or trained on small-scale data, and work well only in a controlled environment.
	Thanks to the availability of large-scale data and the recent progress on neural conversational AI, many neural response generation models have been proposed to perceive and express emotions in an open-domain dialog setting.
	%~\cite{Zhou2018EmotionalCM,asghar2018affective,sun2018emotional}. 
	~\citet{Zhou2018EmotionalCM} proposed Emotional Chatting Machine (ECM) to generate emotional responses given a pre-specified emotion. ECM consists of three components: (1) an emotion category embedding which is fed into each decoding position, (2) an internal emotion state which assumes that the emotion state decays gradually and finally to zero during decoding, and (3) an external memory which allows the model to choose emotional (e.g., \emph{lovely}) or generic (e.g., \emph{person}) words explicitly at each decoding step. 
	The authors also presented some typical emotion interaction patterns in human-human conversations such as {\it empathy} and {\it comfort}, which would inspire more fine-grained design of emotion interaction between human and machine. \citet{asghar2018affective} developed a method of affective response generation that consists of three components: (1) the affective vectors based on Valence/Arousal/Dominance dimensions \cite{warriner2013norms,colombo2019affect}, which serve as a supplement to word vectors; (2) the affective loss functions which maximize or minimize the affective consistency between a post and a response; and (3) the affective beam search algorithm for seeking affective responses. 
	In \cite{zhou2017mojitalk}, a conditional variational autoencoder is proposed to generate more emotional responses conditioned on an input post and some pre-specified emojis. 
	\citet{huber2018emotional} studied how emotion can be grounded in an image to generate more affective conversations. In addition to text, the decoder  takes as input the scene, sentiment, and facial coding features extracted from a given image. Recently, an empathetic dialog corpus is developed to facilitate the research on modeling empathetic interactions in conversation \cite{rashkin2019empa}. We will present dialog datasets in Section \ref{sec:datasets}.

	Controlling the emotion or sentiment has become a popular topic in language generation \cite{hu2017controllable,radford2017learning,ghosh2017affect}.  
	In \cite{radford2017learning}, an RNN-based language model is trained on large-scale review data where some neurons are reported to be highly correlated with sentiment expression. 
	%%%todo: 相关的language generation的工作
	\citet{ghosh2017affect} proposed an affective language model which generates an affective sequence from a leading context. At each decoding position, the model estimates an affective vector of the already generated prefix by keyword spotting using the Linguistic Inquiry and Word Count (LIWC) dictionary \cite{pennebaker2001linguistic}. The vector is then used %with an adjustable hyper-parameter 
	to generate the next word. 
	In \cite{wang2018sentigan}, to generate the reviews of a particular polarity, the authors proposed a multi-class generative adversarial network which consists of multiple generators for multi-class polarities and a multi-class discriminator.

	Despite the research effort reviewed, it is still challenging for a dialog system to express complex emotions in natural language. One difficulty is emotion representation. 
	A simple approach is to project an emotion label to a vector \cite{Zhou2018EmotionalCM}, which is implicit, unexplainable, and subtle. 
	A more sophisticated method is to use Valence/Arousal/Dominance representations: the emotion of each word, sentence, and user state can be represented as V-A-D vectors \cite{warriner2013norms,colombo2019affect}, which is intended to capture psychological and linguistic clues beyond the emotion vector. 
	Another issue of most existing work is that the user's emotion transition during a conversation \cite{majumder2019dialoguernn} is not explicitly modeled. This is crucial for a dialog system to establish a long-term connection with a user because the user is more willing to engage with the system if the system can always detect negative change of her emotion during the conversation and cheer her up through e.g., shifting to new topics that are more comfortable for both parties.   %because the user is constantly influenced during conversational interactions. 
	%Therefore, explicit emotion state modeling is an important direction for future research. 

	%%%%%%%%%%%%%%%%%%%%
	\subsection{Modeling Conversation Behavior and Strategy}
	%%%proactive; 
	%%%提问
	%%%多轮对话中的策略优化
	
	%Dialog behavior and strategy is crucial for building more interactive dialog systems. In human-human conversations, the speakers are taking the turns alternately. And, to achieve a particular social goal, humans have the planning paths, skills, and strategies to achieve that goal, through reactive or proactive behaviors, topic maintenance or switch, or even more sophisticated skills.However, simulating such behaviors and strategies in a conversational system is quite challenging,because they are implicit, subtle, and hard to be formulated.
	
	As pointed out in \cite{zhou2018design}, an open-domain dialog system needs to have enough social skills to have engaging conversations with users and eventually establish long-term emotional connections with users. 
	These social skills include topic planning and dialog policy which 
	can determine whether to drive the conversation to a new topic when e.g., the conversation has stalled, or whether or not to be actively listening when the user herself is engaged in the conversation. 
	%However, encoding or learning these social skills for a dialog system is a non-trivial task. 
	\citet{openquest15} elucidated the challenges of proactiveness in dialogue systems and how they influence the effectiveness of turn-taking behaviour in multimodal and unimodal dialogue systems. 
	\citet{Yu2016strategy} proposed several generic conversational strategies (including grounding on entities and OOV words, topic switch, activity initiation, and joke telling) to handle possible system breakdowns in non-task-oriented dialog systems, and designed policies to select these strategies according to dialog context.
	%several dialog strategies including context tracking, lexical semantics, and general diversion strategies are studied to exhibit proactive behaviors. Although the realization of these strategies is based on templates, the dialog policy can be optimized with reinforcement learning during multi-turn interactions. 
	\citet{earlysign18} addressed the problem of predicting from the very beginning
	of a conversation whether it will get out of hand. The authors developed a framework for capturing pragmatic devices, such as politeness strategies and rhetorical prompts, used to start a conversation, and
	analyzed their relation to its future trajectory. Applying this framework in a controlled setting, it is possible to detect early warning signs of antisocial behavior in online discussions.
	
	%a corpus for detecting early warning signs of conversational failure in online discussions. 
	%The authors investigated how politeness strategies and other rhetorical devices are tied to the future trajectory of a conversation.  
	
	The above studies inspire researchers to devise new methods of incorporating social skills into an open-domain dialog system.
	%via rules or machine learned models. 
	%develop a proactive, fluent dialog systems. 
	In \cite{li2016stalematebreakerIRbased}, a retrieval-based method is proposed to first detect the sign of stalemate using rules, and then retrieve responses that contain the entities that are relevant to the input, assuming that a proactive reply should contain the entities that can be triggered from the ones in the input. 
	\citet{prosuggestion18} proposed a proactive suggestion method where a look-ahead post for a user is decoded in addition to the system response, conditioned on the context and the previously generated response. The user can use the generated post directly, or type a new one during conversation. 
	\citet{askquestion18} argued that asking good questions in conversation is shown to be an important proactive behavior. A typed decoder is proposed to generate meaningful questions by predicting a type distribution over topic words, interrogatives, and ordinary words at each decoding position. The final output distribution is modeled by the type distribution, leading to a strong control over the question to be generated. 
	\citet{rao2018learning} also argued that question asking is fundamental to communication, and that a good question is the one whose expected answer will be useful. They built a neural network model for ranking clarification questions, evaluated on a dataset of clarification questions (post-question pairs) extracted from StackExchange. 
	%%%}
	\citet{ke2018senfunc} conducted a systematic study of generating responses with different sentence functions, such as interrogative, imperative, and declarative sentences. These sentence functions play different roles in conversations.
	% can be used to achieve differ speaking purposes in conversation. 
	For instance, imperative responses are used to make requests, give directions and instructions, or elicit further interactions while declarative responses make statements or explanations.
	%%%%\textcolor{blue}{
	\citet{tang2019target} proposed a new dialog planning task in which the conversation should eventually reach a target (defined by a topical keyword) from any initial topics. In such a task, it is required to plan proactively the topic path to the final target.%%%}

	%%%\textcolor{blue}{
	There are two important directions for future research. First is the comprehensive investigation of conversation behaviors in human-human dialog. This is still largely ignored, possibly due to the lack of real-world conversations. The dialog data in online forums \cite{earlysign18} and psychological counseling \cite{althoff2016large,perez2017understanding,zhang2019finding,cao2019observing} are of high value  for this research. But the data in a wide variety of scenarios are still in significant shortage. 
	Second is to create a more sophisticated real-world dialog setting for system development and evaluation. Existing work largely targets at modeling atomic strategy in dialog systems, namely, single strategy for emotion interaction \cite{Zhou2018EmotionalCM}, topic control \cite{wang2018chat}, question asking \cite{askquestion18,ke2018senfunc}, and so on.
	Most of the studies are merely evaluated with the single-turn setting. However, to accomplish more complex social goals such as emotional comfort or counseling, it is necessary to design composite strategies that consider emotion, topic, and proactivity comprehensively in multi-turn conversation. Therefore, there is increasing demand for collecting or constructing more complex dialog data with well-designed task goals, and for developing more sophisticated dialog policy models. 
	%%%}
	
	%%%%\Jianfeng{I suggest you organize Sec. 3, 4, 5 in a top-down manner. (1) Start with problem definition and an overview of solutions; (2) Group these solutions into different categories and review them one by one; (3) Finally review the public datasets -- this part can be very brief since we will provide details in Sec. 7.}

	%%%%In multi-turn interactions, dialog strategy seems to be even more crucial to build an intelligent system. ****** 
	%%%% multi-turn 写 topic 和 RL控制策略
	
	%\subsection{Improving Consistency with Personalization}

	%%%%%%%%%section 7
	%%%%%%%%%%%%%%%%%%%%%%%%%%%
	\section{Open-domain Dialog Evaluation}
	\label{sec:evaluation}
	%Dialog evaluation is a fundamental problem in dialog systems.
	Evaluating the quality of an open-domain dialog system is challenging because open-domain conversations are inherently open-ended~\cite{ram2018conversational}. 
	For example, if a user asks the question "what do you think of Michael Jackson?", there are hundreds of distinct but plausible responses. 
	%This makes evaluating social, non-task oriented dialog systems extremely challenging.
	% In some sense, dialog evaluation is almost as difficult as generating good conversations in open-domain dialog systems. 
	% If there is a good metric to automatically measure the quality of generated conversations, it can be definitely used to guide dialog generation. 
	% To make full assessment of the generation quality, researchers commonly resort to 
	Evaluation of a dialog system can be performed manually or in an automatic way. 
	In manual evaluation, human judges are hired to assess the generated results in terms of predefined metrics, with well-documented guidelines and exemplars. Evaluation is conducted by either scoring each individual result (point-wise) or comparing two competing results (pair-wise). In some dialog evaluation challenges, manual evaluation is commonly adopted in the final-stage competition ~\cite{dinan2019second,ram2018conversational}. For instance, the second conversational intelligence challenge \cite{dinan2019second} adopted manual evaluation by paid workers from Amazon Mechanical Turk and unpaid volunteers, and the organizers reported the rating difference between the two user groups: the volunteers' evaluation had relatively fewer good (i.e. long and consistent) dialogues, while paid workers tended to rate the models higher than the volunteers. 
	
	Since manual evaluation is expensive, time-consuming, and not always reproducible, %of high variance (sometimes not reproducible), 
	automatic evaluation is more frequently used, especially at the early stage of development.
	For retrieval-based methods, traditional information retrieval evaluation metrics such as precision{@}k, mean average precision (MAP), and normalized Discounted Cumulative Gain (nDCG)~\cite{manning2008introduction} are applicable. 
	For generation-based models, metrics such as perplexity, 
	BLEU~\cite{papineni2002bleu}, and distinct-$n$ \cite{lidiversity}, are widely used. 
	Perplexity measures how well a probabilistic model fits the data, and is a strong indicator whether the generated text is grammatical. 
	BLEU, adopted from machine translation, measures the lexical overlap between the generated responses and the reference ones. 
	Distinct-$n$ measures the diversity by computing the proportion of unique $n$-grams in a generated set. 
	However, \cite{liu2016not} argued that automatic metrics such as BLEU, ROUGE \cite{lin2004rouge}, and METEOR \cite{banerjee2005meteor} all have low correlation with manual evaluation. But as pointed out in \cite{gao2019neural}, the correlation analysis in \cite{liu2016not} is performed at the sentence level while BLEU is designed from the outset to be used as a corpus-level metric. \cite{galley2015deltableu} showed that the correlation of string-based metrics (BLEU and deltaBLEU) significantly increases with the units of measurement bigger than a sentence.  
	Nevertheless, in open-domain dialog systems, the same input may have many plausible responses that differ in topics or contents significantly. Therefore, low BLEU (or other metrics) scores do not necessarily indicate low quality as the number of reference responses is always limited in test set. Therefore, there has been significant debate as to whether such automatic metrics are appropriate for evaluating open-domain dialog systems \cite{gao2019neural}.
	%%
	%%%
	
	%%%%写两个评测框架
	Recently, trainable metrics for open-domain dialog evaluation have attracted some research efforts. 
	%there are research attempts that propose learnable metrics with machine learning models.
	\citet{lowe2017towards}
	proposed a machine-learned metric, called ADEM, for open-domain dialog evaluation. They presented a variant of the VHRED model \cite{serban2017hierarchical-1} that takes context, user input, gold and system responses as input, and produces a qualitative score between 1 and 5.
	%which was trained on the data with human judged scores. This metric estimates the score of a generated response given the input utterance as well as the ground-truth response. 
	The authors claimed that the learned metric correlates better with human evaluation than BLEU and ROUGE. 
	\cite{tao2018ruber} proposed an evaluation model, called RUBER, which does not rely on human judged scores. RUBER consists of a referenced component to measure the overlap between a system response and a reference response, and an unreferenced component to measure the correlation between the system response and the input utterance.
	However, as pointed out in \cite{sai2019re}, ADEM can be easily fooled with a variation as simple as reversing the word order in the text. Their experiments on several such adversarial scenarios draw out counter-intuitive scores on the dialogue responses. In fact, any trainable metrics lead to potential problems such as overfitting and ``gaming of the metric''
	\footnote{In discussing the potential pitfalls of machine-learned evaluation metrics, \citet{albrecht07reex} argued for example that it would be ``prudent to defend against the potential of a system gaming a subset of the features.'' In the case of deep learning, this gaming would be reminiscent of making non-random perturbations to an input to drastically change the network's predictions, as it was done, e.g., with images in \citep{szegedy13intriging} to show how easily deep learning models can be fooled. Readers refer to Chapter 5 in \citet{gao2019neural} for a detailed discussion.} 
	\cite{albrecht07reex}, 
	%%%%jianfeng: please add a footnote here
	which might explain why none of the previously proposed machine-learned evaluation metrics \citep[etc.]{corston01ml,kulesza04learning,lita05blanc,albrecht07reex,gimenez08smor,pado09measuring,stanojevic14fitting} is used in official machine translation benchmarks. 
	Readers refer to \cite{gao2019neural} for a detailed discussion.
	
	These research attempts indicate that what makes a good conversation is a challenging question to answer. \citet{see2019makes} discussed four attributes that are associated with the control of open-domain dialog generation: repetition, specificity, response-relatedness, and question-asking. They argued that existing work has ignored the importance of the conversational flow, because existing models repeat or contradict previous statements, fail to balance specificity with genericness, and are unable to balance asking questions with other dialogue acts. Experiments on Persona-Chat \cite{Zhang2018Personalizing-dogpet} show that higher engagingness scores in human judgement can be obtained by optimizing the control of the four attributes in multi-turn conversations. Therefore, considering these attributes in automatic evaluation, implicitly or explicitly, is expected to lead to new evaluation metrics that correlate well with human evaluation.

	Recently, there are research attempts to combine human evaluation and automatic evaluations for natural language generation systems. \citet{hashimoto2019unifying} argued that human evaluation captures quality but not diversity while statistical evaluation (i.e., perplexity) captures diversity but not quality. They proposed a unified framework which evaluates both in terms of the optimal error rate of predicting whether a sentence is human- or machine-generated. As mentioned above, automatic metrics such as sentence-level BLEU correlates poorly with human judgement, thereby easily leading to systematic bias against model improvements. On the other hand, the average of human judgements is unbiased but is very expensive to collect. Therefore, \citet{chaganty2018price} % used control variates to 
	combined automatic metrics with human evaluation to obtain an unbiased estimator with lower cost than using solely human evaluation.
	
	%%%%%%这里补充manual + automatic结合的评估工作
	%%%% 修改这一段，补充内容
	%%%%for better human judgements for open-domain conversational system. It also lists some relevant papers on evaluations the authors don't discuss.
	%%%\textcolor{blue}{
	All of the above research suggests that automatic evaluation of dialog systems is by no means a solved problem. We argue that, for open-domain dialog evaluation, the major difficulty derives from in the one-to-many essence: in any given dataset, the number of observable responses for the same input post is limited, yet there are many appropriate responses not presented in the dataset. Therefore, automatic metrics that are trained on a dataset will be inherently questionable because the topic coverage and the number of observable outputs are largely limited by the dataset. 
	Thus, uncovering those underlying outputs for an input post is an interesting area for future research.
	%%%}
	%%%We believe that developing a successful automatic evaluation metric has two prerequisites. First, there should be a fairly large, representative conversational dataset. This dataset should have a good coverage of daily life topics and domains. Second, for each input, there should be multiple appropriate responses to address the one-to-many essence in open-domain dialog.  
	
	%%%[Jianfeng: I don't think any machine-learned model should be used for evaluation due to the ``gaming of the metric'' described above. However, automatic evaluation is achievable given the availability of a large corpus and multiple references.]
	%%% Without such a corpus, it would not be sound enough to train a model for automatic dialog evaluation.
	
	%%%%##########

	\section{Open-domain Dialog Corpora}
	\label{sec:datasets}
	%%%%%%下面这个section 我拿不准---一个方式是每个数据集介绍一小段，一个是现在的这个形态
	%%%%%%jianfeng---which is better???
	
	%%%\textcolor{blue}{
	Recently, the availability of dialog corpora has largely advanced the development of neural models for open-domain conversation generation. 
	An incomplete survey on these dialog datasets is presented in Table \ref{tab:corpora}\footnote{Readers may refer to an old survey published in 2015, which covers datasets for both open-domain and task-oriented dialog models \cite{serban2015survey}. We only list the corpora that are frequently used or recently proposed in the literature, most of which are not covered by \cite{serban2015survey}. }. 
	These corpora differ in topic, source (where or how the data is collected), language, data scale, and the design features. 
	%We will briefly introduce these datasets as follows:
	%%%}

	\noindent \textbf{Short Text Conversation (STC)} ~\cite{shang-2015-NRM}: This corpus is collected from a Chinese social media, Weibo. There are 219,905 posts and 4,308,211 responses in the training data. It can be used for studying the one-to-many problem in dialog modeling since each post has multiple responses. On top of this corpus, \citet{Zhou2018EmotionalCM} proposed an emotional STC dataset (ESTC) in which each utterance is tagged in terms of six emotion classes by an emotion classifier with an accuracy of 62.3\%. ESTC is frequently used in building empathetic dialog systems \cite{Zhou2018EmotionalCM}. 
	
	\noindent \textbf{Twitter Triple Corpus}~\cite{sordoni2015neural}: This corpus contains 29M context-message-response triples from the Twitter FireHose, covering the 3-month period from June 2012 through August 2012. Additionally, the validation and test sets have 4,232 triples which are scored no less than 4 in 5-point scale by human annotators. However, this corpus is not publicly available. 
	%%%%---jianfeng: please check this 这个数据集可以公开获取吗？

	%%%%最重要的是清晰
	\noindent \textbf{PersonalDialog}~\cite{zheng2019personalized}: This corpus is constructed toward building personalized conversation models.
	%%%%先说有什么用
	%%%
	The data is collected from a Chinese social media, Weibo. Each dialogue is composed of a post and its following replies from different users. The personal profile of each user is collected, which includes  five personality traits: Gender, Age, Location, Interest Tags, and Self Description.
	%%%%收集方式(为了避免以后的麻烦，尽量少提是从微博上收集的)
	%%%
	This dataset contains 20.83M conversations and 8.47M user profiles. The total number of utterances are 56.25M and each utterance contains 9.35 tokens. A considerable amount of dialogues (3.43M sessions) in this dataset have multiple turns (more than 4 utterances).
	%%%%数据统计
	%%%
	This corpus is the first dialogue corpus that contains real social conversations and diversified personality traits for each user. %%%It will facilitate not only the study of personalized dialogue generation, but also other researches on sociolinguistics, for instance.
	%%%%最后能讲1-2句独特的地方最好，如果

	\noindent \textbf{DailyDialog}~\cite{li2017dailydialog}: This corpus contains multi-turn dialogs on daily life topics. The raw data were crawled from several websites which serve for English learner to practice English. The dataset contains 13,118 dialogs, with an average of 7.9 turns per dialog and 14.6 words per turn. The appealing feature of this corpus is that it provides manual annotation on intent (\emph{Inform, Questions, Directives, and Commissive}) and emotion (\emph{Anger, Disgust, Fear, Happiness, Sadness, and Surprise}), which may support the research on emotion interaction and dialog act modeling.
	
	%%%%这个数据暂不包括
	%%%%\noindent \textbf{Cornell Movie-Dialog Corpus} \cite{danescu2011chameleons}: The corpus contains short conversations extracted from movie scripts. The amount of metadata available for each conversation is rich in this dataset, including movie metadata such as genre, release year, and IMDB rating, as well as character metadata such as gender and position on movie credits. Although this corpus contains 220,000 dialogue excerpts, it only contains 300,000 utterances; therefore, many of the excerpts consist of single utterances.

	\noindent \textbf{Ubuntu Dialog Corpus}~\cite{lowe2015ubuntu}: This corpus contains two-party conversations that solve technical issues with Ubuntu. The data were extracted from online conversation logs in Ubuntu-related chat rooms on the Freenode Internet Relay Chat (IRC) network. In each log, a user may ask a technical question to be solved and other users can respond to the question. The log session will terminate until the problem is solved. A two-party conversation will be extracted from the chat log\footnote{Each chat log is a multi-party conversation, but only two-party sub-conversations which involve the same two users are retained.}. The corpus contains 930,000 human-human dialogs and 7,100,000 utterances, with an average of 7.71 turns per dialog and 10.34 words per utterance. Strictly speaking, this dataset is task-specific instead of open-domain conversation. This corpus is commonly used to evaluate retrieval-related models.

	\noindent \textbf{Persona-Chat}~\cite{Zhang2018Personalizing-dogpet}: This crowdsourced corpus is designed for personalized dialog modeling. In each conversation, each worker is given a persona which is defined by up to 5 sentences describing personal hobby or state (e.g., \emph{I like swimming}, or \emph{I need to lose weight}). Two workers are instructed to know each other through interaction. During the conversation, each worker should follow her own persona and try to know the partner's information. The dataset consists of 10,981 dialogs with 164,356 utterances.  
	%%%%%%%%%%%%%%%%%%%%%%%%%%%

	%%%%%%%%%%%%%%%%%%%%%%%%%%%%%
	\noindent \textbf{CMU Document-grounded conversation (CMU DOG)}~\cite{zhou2018DoG}: This corpus, designed for document or knowledge grounded dialog modeling, contains crowd-sourced conversations that are talking about 30 movies. The information about each movie is given through a correspondent Wikipedia article. There are two modes for data collection: only one worker has the movie document and both workers have the movie document during conversation. The dataset consists of 4,112 conversations with an average of 31.6 utterances per dialog and 10.8 words per utterance. 
	
	%%%%这个名字可能不太合适
	\noindent \textbf{}~\cite{moghe2018towards}: The corpus can be viewed as an expanded version of CMU DOG. The conversations discuss about 921 movies, and the knowledge about each movie is composed of a fact table\footnote{ which contains box office collection, similar movies (for recommendation), awards, and tag-lines.}, the plot description, and reviews and comments on the movie. 
	The corpus contains 9,071 conversations and 90,810 utterances with an average of 10 utterances per dialog and 15.3 words per utterance. The corpus is useful for studying the use of heterogeneous knowledge in conversation generation.
	
	\noindent \textbf{Wizard of Wikipedia} \cite{dinan2018WOW}: This corpus contains conversations that are grounded with knowledge retrieved from Wikipedia. The dataset covers 1,365 topics, each linked to a Wikipedia article. These topics include commuting, Gouda cheese, music festivals, podcasts, bowling, and Arnold Schwarzenegger.
	Each conversation is made between a knowledge expert and a curious learner, and the expert has full access to the Wikipedia article of a topic but the learner does not. The corpus consists of 22,311 dialogues and 201,999 utterances, with an average of 9 utterances per dialog. Each utterance is grounded to a selected knowledge sentence or indicated by that no knowledge is used.
	
	\noindent \textbf{Grounded Response Generation at DSTC7} 
	\cite{qin2019conversing}: The dataset, which is first released for the "sentence generation" task at the 7th Dialog System Technology Challenges (DSTC7) \cite{galley2019grounded}, is developed for grounded conversation modeling. It consists of conversation threads extracted from Reddit data. Each conversation contains exactly one URL to a web page (grounding) that defines the topic of the conversation. The dataset contains 2.8M conversation instances respectively divided into train, validation, and test based on date ranges: years 2011-2016 for train, Jan-Mar 2017 for validation, and the rest of 2017 for test, which consists of 2,208 conversational turns, each with 6 human responses. To access the human performance using the test set, one of the 6 human responses is set aside, and the remaining 5 responses serve as ground truths for evaluating different systems.
	
	\noindent \textbf{Topical-Chat} \cite{gopalakrishnan2019topical}: This corpus is designed towards building dialog systems that can converse with humans on various topics. It covers 300 popular topic entities spanning 8 domains including fashion, politics, books, and sports. For each entity, the authors fetched the Wikipedia lead section, and crowdsourced 8-10 fun facts. Furthermore, they fetched Washington Post articles in 2018 that each referenced 3 or more of the 300 entities. The authors then created a set of reading sets, each containing the wiki-information, several fun facts, and a Washington Post article. Workers were partnered up to converse, with symmetric or asymmetric settings where symmetric means two workers have the same reading set, and asymmetric with different sets. 
	The dataset contains 11,319 conversations and 248,014 utterances with an average of 22 turns per dialog and 19.8 words per turn.

	\noindent \textbf{OpenDialKG} \cite{moon2019opendialkg}: In this corpus, each dialog is paired with its corresponding "knowledge graph (KG) paths" that weave together the KG entities and relations. It was collected with a Wizard-of-Oz setting by connecting two crowd-workers to engage in a chat session. 
	The first worker is given a seed entity and asked to initiate a conversation about that entity. The second worker is provided with a list of facts relevant to that entity, and asked to choose the most natural and relevant facts and use them to frame a free-form conversational response. After the second worker sends her response, new multi-hop facts from KG are surfaced to include paths initiating from new entities introduced in the latest message. The circle continues for several rounds, which simulates a random walk over the knowledge graph. 
	The dataset covers four domains (movies, books, sports and music), with a KG of total 1,190,658 fact triples. It contains 15,673 dialogs and 91,209 turns with an average of 5.8 turns per dialog. 
	This corpus is useful in studying conversational reasoning, while it is not yet publicly available.

	\noindent \textbf{DuConv} \cite{wu2019proactive}: This corpus covers topics on movies and film stars whose related knowledge was crawled from the Web. Then, two linked entities were randomly sampled to construct a conversation goal like "[start]$\rightarrow entity_a \rightarrow entity_b$" where $entity_b$ is the final target of the conversation. Two annotators were asked to conduct knowledge-driven conversations with a leader-follower mode. The leader needs to change the conversation topics following the conversation goal and meanwhile keeps the conversation as engaging as possible. 
	The dataset contains 29,858 dialogs and 270,399 utterances with an average of 9.1 turns per dialog and 10.6 words per turn. This corpus is useful in constructing knowledge-driven proactive dialogue systems.
	
	%%%\noindent \textbf{ }:
	
	{
		\noindent \textbf{DyKgChat}~\cite{chendykgchat}: This corpus is collected for knowledge-grounded conversation modeling. The conversations are from the scripts of a Chinese palace drama (Hou Gong Zhen Huan Zhuang, with 76 episodes and hundreds of characters), and an English sitcom "Friends" (with 236 episodes and six main characters). The paired knowledge graphs are manually constructed. The corpus contains 1,247/3,092 dialogs, with 13.76/18.68 turns per dialog and 27.0/16.5 words per turn for the  Chinese and English TV series, respectively. The most interesting feature of this corpus is that it contains evolving knowlege graphs.}
	
	%%%%最重要的是清晰
	\noindent \textbf{EmpatheticDialogues}~\cite{rashkin2019empa}: This corpus is constructed toward building empathetic open-domain conversation models.
	%%%%先说有什么用，总括性的描述
	The data is collected by crowd workers with a speaker-listener mode. The speaker starts the conversation from a pre-set emotion state (e.g., \emph{afraid}) and a personal situation description (e.g., \emph{Speaker felt afraid when she has been hearing noises around the house at night}), and the listener becomes aware of the underlying situation through what the Speaker says and responds. 
	%%%%收集方式，渠道，模式等
	The corpus contains 24,850 conversations, and the average number of utterances per conversation and words per turn is 4.31/15.2 respectively.
	%%%%数据统计
	The corpus is useful in modeling emotion interactions in multi-turn conversation.
	%%%%最后能讲1-2句独特的地方最好，如果有的话
	
	\noindent \textbf{Target-Guided Conversation}~\cite{tang2019target}: This corpus is constructed towards building target-guided open-domain conversation models. It's derived from Persona-Chat \cite{Zhang2018Personalizing-dogpet} without the persona information. The keywords of each utterance, which indicate the targets in this task, are automatically extracted by a rule-based keyword extractor. The corpus contains 8,939/500/500 dialogs, 101,935/5,602/5,317 utterances and 2,678/2,080/1,571 keywords in the training/validation/test set, respectively. The average number of keywords in each utterance is about 2.0. This corpus is expected to model the turn-level keyword transition and the discourse-level target-guided dialogue strategy.

	\noindent \textbf{PERSUASION-FOR-GOOD}~\cite{wang2019persuasion}: This corpus contains persuasion conversations for charity donation where each speaker's  psychological profile attributes and sociodemographic backgrounds such as age and income were also collected. The data is collected with a persuader-persuadee mode in four steps. First, workers were asked to complete a pre-task survey to assess their psychological profile variables. Second, two workers were randomly assigned the roles of persuader and persuadee where the persuader needed to persuade the persuadee to donate part of his/her task earning to the charity, and the persuader could also choose to donate. Third, both the persuader and the persuadee were asked to input the intended donation amount privately though a text box when the conversation was ended. Last, workers were asked to complete a post-survey to assess their sociodemographic backgrounds. The corpus contains 1,017 dialogs, with an averge of 10.43 turns per dialog and 19.36 words per utterances. It also provides manual annotation in terms of persuasion strategy and dialog act for each sentence. This dataset is interesting for studying personalized dialog and complex strategy modeling.

	\section{Discussions and Future Trends}
	\label{sec:future}
	%%%%关于topic-grounded; knowledge-grounded conversation
	%%%%system-initiative; topic or knowledge; grouned 能力
	%%%%
	In this paper, we review the recent progress in developing open-domain dialog systems. We focus the discussion on neural approaches that have been proposed to deal with three key challenges: semantics, consistency, and interactiveness. We review open-domain dialog evaluation metrics for both manual and automatic evaluation, and share our thoughts on how to develop better automatic evaluation metrics. We survey frequently-used and recently-proposed corpora for the development of evaluation of open-domain dialog systems.
	%The works we have selected to present in this survey paper is based on our unique views and are by no means complete. Nevertheless, we hope that our colleagues would find the discussion instrumental to inspire new ideas and efforts for developing more intelligent dialog systems.
	
	%%%%【jianfeng】my thought is still again to focus on the three issues %%%%%%
	Differing from early generations of dialog assistants which are designed for simple tasks that require only short, domain-specific conversations, such as making reservation or asking for information, open-domain dialog systems are design to be AI companions that are able to have long, free-form social chats with human users.
	%that occur naturally in social and professional human interactions 
	\cite{ram2018conversational,zhou2018design}. 
	%the new frontier for the research and development of dialog systems.
	%%%%%
	%Current state-of-the-art systems are still a long way from being able to deliver natural daily conversations with humans. 
	Despite the recent progress as reviewed in this paper, achieving sustained, coherent, and engaging open-domain conversations remains very challenging.
	We conclude this paper by discussing some future research trends
	%, aiming to build more intelligent open-domain dialog systems:
	%\noindent $\bullet$ {\bf Topic-grounded or knowledge-grounded conversational agents}. 
	\paragraph{Topic and Knowledge Grounding}
	To deliver contentful conversations, it is important to ground conversations in real-world topics and entities (e.g., in knowledge bases). This is part of the semantics challenge we have discussed in Section \ref{sec:semantics}. Since natural language understanding in open-domain dialog systems is extremely challenging, knowledge grounding provides to some degree the ability of understanding language in dialog context, as shown in several preliminary studies \cite{Zhou2018Commonsense, liu2018knowledge, zhuwenya2017}.
	% However, open-domain named entity recognition and linking is another challenging problem, nevertheless, this is an important step for addressing the semantics issue in dialog systems. 
	Even though an open-domain dialog system has no access to annotated dialog acts (which are available only for task-oriented dialog) to learn to explicitly detect an user's intents (labeled by dialog acts), the system can still play a proactive role of leading the conversation by for example suggesting new topics, if the key concepts and entities are correctly recognized and linked to a knowledge base \cite{sounding2017,pichl2018alquist, askquestion18,zhou2018design}. Several recently proposed corpora, as described in Section \ref{sec:datasets}, provide new test beds for this research.
	
	% \noindent $\bullet$ {\bf Emotional social bot.} 
	\paragraph{Empathetic Computing}
	Sentiment and emotion form a key factor for making effective social interactions, and is crucial for building an empathetic social bot. 
	Existing studies \cite{Zhou2018EmotionalCM, asghar2018affective, zhou2018design, zhou2017mojitalk,rashkin2019empa} in this direction are still in the infant stage, as they only deal with superficial expression of emotion. 
	A future empathetic machine should be able to perceive a user's emotion state and change, deliver emotionally influential conversations, and evaluate the emotional impact of its action, much of which should be tightly aligned with psychological studies. 
	These become more important in more complicated scenarios such as psychological treatment, mental health, and emotional comforting. 
	Moreover, it is insufficient for an empathetic machine to use only text information. 
	The signals from other modalities such as facial expression and speech prosody should also be leveraged \cite{liao2018knowledge,zhang2019mnbt,cui2019user}. 
	To foster the research, \citet{saha2017multimodal} developed a conversational dataset consisting of multi-modal dialog sessions in a fashion domain where each turn contains a textual utterance, one or more images, or a mix of text and images.
	
	%\noindent $\bullet$ {\bf Personality of a social bot.} 
	\paragraph{Personality of a Social Bot}
	A coherent personality is important for a social bot to gain human trust, thereby improving the consistency and interactiveness of human-machine conversations. Personality (e.g., Big five traits) has been well-defined in psychology \cite{norman1963toward,gosling2003very}. However, existing studies \cite{Li2016_ACL-persona,Qian2017AssigningProfile,Zhang2018Personalizing-dogpet,zhou2018design} are yet to be significantly extended by incorporating the results of multidiscipline research covering psychology, cognitive science, computer science, etc. The central problem is how to ensure personality-coherent behaviors in conversations and evaluate such behaviors from the perspectives of multidisciplines, particularly via psychological studies.  
	
	%\noindent $\bullet$ {\bf Multi-modality dialog systems.} 
	%It is natural for a dialog system to not only perceive multi-modality input including text, voice, and vision, but also output with multiple modalities. Xiaoice \cite{shum2018xiaoice}  is a good exemplar for accepting multi-modality input and enabling multi-modality output, however, the strategy of output modality selection is rather heuristic. The major obstacle in this research line is the lack of multi-modality dialog data. The only corpus is available in \cite{saha2017multimodal} where multi-modality dialogues in a fashion domain are collected, and each turn contains a textual utterance, one or more images, or a mix of a text and several images.
	
	%\noindent $\bullet$ {\bf Controllability of dialog generation.} 
	\paragraph{Controllability of dialog generation}
	Most existing open-domain dialog systems are based on neural response generation models. Due to the essence of probabilistic sampling used in language generation, controllability is a challenging issue as repetitive, bland, illogical or even unethical responses are frequently observed. Controllability is closely related to the interpretability and robustness of neural network models. Achieving controllability requires new breakthroughs in modeling, such as the hybrid approaches that combine the strengths of both neural and symbolic methods.

	\section{Acknowledgement}
	This work was supported by the National Science Foundation of China (Grant No. 61936010/61876096), and the National Key R\&D Program of China (Grant No. 2018YFC0830200). We would like to thank THUNUS NExT Joint-Lab for the support.
	%%%%---需要更改

	%%We would like to thank Prof. Xiaoyan Zhu for her unreserved support. 
	We would like to thank Pei Ke, Qi Zhu, Chujie Zheng, Yaoqin Zhang, Hao Zhou, Chris Brockett, Bill Dolan, and Michel Galley for their discussions and contributions to this paper. We truly thank anonymous reviewers for their valuable reviews and comments.
	
	\bibliographystyle{ACM-Reference-Format}
	\bibliography{tois} %,acmart,sample-base}


\begin{thebibliography}{196}

%%% the Web address.

\ifx \showCODEN    \undefined \def \showCODEN     #1{\unskip}     \fi
\ifx \showDOI      \undefined \def \showDOI       #1{#1}\fi
\ifx \showISBNx    \undefined \def \showISBNx     #1{\unskip}     \fi
\ifx \showISBNxiii \undefined \def \showISBNxiii  #1{\unskip}     \fi
\ifx \showISSN     \undefined \def \showISSN      #1{\unskip}     \fi
\ifx \showLCCN     \undefined \def \showLCCN      #1{\unskip}     \fi
\ifx \shownote     \undefined \def \shownote      #1{#1}          \fi
\ifx \showarticletitle \undefined \def \showarticletitle #1{#1}   \fi
\ifx \showURL      \undefined \def \showURL       {\relax}        \fi
% invisible to TeX
\providecommand\bibfield[2]{#2}
\providecommand\bibinfo[2]{#2}
\providecommand\natexlab[1]{#1}
\providecommand\showeprint[2][]{arXiv:#2}

\bibitem[\protect\citeauthoryear{Adiwardana, Luong, So, Hall, Fiedel,
  Thoppilan, Yang, Kulshreshtha, Nemade, Lu, et~al\mbox{.}}{Adiwardana
  et~al\mbox{.}}{2020}]%
        {adiwardana2020towards}
\bibfield{author}{\bibinfo{person}{Daniel Adiwardana},
  \bibinfo{person}{Minh-Thang Luong}, \bibinfo{person}{David~R So},
  \bibinfo{person}{Jamie Hall}, \bibinfo{person}{Noah Fiedel},
  \bibinfo{person}{Romal Thoppilan}, \bibinfo{person}{Zi Yang},
  \bibinfo{person}{Apoorv Kulshreshtha}, \bibinfo{person}{Gaurav Nemade},
  \bibinfo{person}{Yifeng Lu}, {et~al\mbox{.}}}
  \bibinfo{year}{2020}\natexlab{}.
\newblock \showarticletitle{Towards a Human-like Open-Domain Chatbot}.
\newblock \bibinfo{journal}{\emph{arXiv preprint arXiv:2001.09977}}
  (\bibinfo{year}{2020}).
\newblock


\bibitem[\protect\citeauthoryear{Albrecht and Hwa}{Albrecht and Hwa}{2007}]%
        {albrecht07reex}
\bibfield{author}{\bibinfo{person}{Joshua Albrecht} {and}
  \bibinfo{person}{Rebecca Hwa}.} \bibinfo{year}{2007}\natexlab{}.
\newblock \showarticletitle{A Re-examination of Machine Learning Approaches for
  Sentence-Level MT Evaluation}. In \bibinfo{booktitle}{\emph{Proceedings of
  {ACL} 2007, June 23-30, 2007, Prague, Czech Republic}}.
  \bibinfo{pages}{880--887}.
\newblock


\bibitem[\protect\citeauthoryear{Althoff, Clark, and Leskovec}{Althoff
  et~al\mbox{.}}{2016}]%
        {althoff2016large}
\bibfield{author}{\bibinfo{person}{Tim Althoff}, \bibinfo{person}{Kevin Clark},
  {and} \bibinfo{person}{Jure Leskovec}.} \bibinfo{year}{2016}\natexlab{}.
\newblock \showarticletitle{Large-scale analysis of counseling conversations:
  An application of natural language processing to mental health}.
\newblock \bibinfo{journal}{\emph{Transactions of the Association for
  Computational Linguistics}}  \bibinfo{volume}{4} (\bibinfo{year}{2016}),
  \bibinfo{pages}{463--476}.
\newblock


\bibitem[\protect\citeauthoryear{Asghar, Poupart, Hoey, Jiang, and Mou}{Asghar
  et~al\mbox{.}}{2018}]%
        {asghar2018affective}
\bibfield{author}{\bibinfo{person}{Nabiha Asghar}, \bibinfo{person}{Pascal
  Poupart}, \bibinfo{person}{Jesse Hoey}, \bibinfo{person}{Xin Jiang}, {and}
  \bibinfo{person}{Lili Mou}.} \bibinfo{year}{2018}\natexlab{}.
\newblock \showarticletitle{Affective Neural Response Generation}. In
  \bibinfo{booktitle}{\emph{Advances in Information Retrieval - 40th European
  Conference on {IR} Research, {ECIR} 2018, Grenoble, France, March 26-29,
  2018, Proceedings}}, Vol.~\bibinfo{volume}{10772}. \bibinfo{pages}{154--166}.
\newblock


\bibitem[\protect\citeauthoryear{Bahdanau, Cho, and Bengio}{Bahdanau
  et~al\mbox{.}}{2015}]%
        {Bahdanau2015Neural}
\bibfield{author}{\bibinfo{person}{Dzmitry Bahdanau},
  \bibinfo{person}{Kyunghyun Cho}, {and} \bibinfo{person}{Yoshua Bengio}.}
  \bibinfo{year}{2015}\natexlab{}.
\newblock \showarticletitle{Neural Machine Translation by Jointly Learning to
  Align and Translate}. In \bibinfo{booktitle}{\emph{{ICLR} 2015, San Diego,
  CA, USA, May 7-9, 2015}}.
\newblock


\bibitem[\protect\citeauthoryear{Balakrishnan, Rao, Upasani, White, and
  Subba}{Balakrishnan et~al\mbox{.}}{2019}]%
        {balakrishnan2019constrained}
\bibfield{author}{\bibinfo{person}{Anusha Balakrishnan},
  \bibinfo{person}{Jinfeng Rao}, \bibinfo{person}{Kartikeya Upasani},
  \bibinfo{person}{Michael White}, {and} \bibinfo{person}{Rajen Subba}.}
  \bibinfo{year}{2019}\natexlab{}.
\newblock \showarticletitle{Constrained Decoding for Neural NLG from
  Compositional Representations in Task-Oriented Dialogue}.
\newblock \bibinfo{journal}{\emph{arXiv preprint arXiv:1906.07220}}
  (\bibinfo{year}{2019}).
\newblock


\bibitem[\protect\citeauthoryear{Banerjee and Lavie}{Banerjee and
  Lavie}{2005}]%
        {banerjee2005meteor}
\bibfield{author}{\bibinfo{person}{Satanjeev Banerjee} {and}
  \bibinfo{person}{Alon Lavie}.} \bibinfo{year}{2005}\natexlab{}.
\newblock \showarticletitle{METEOR: An automatic metric for MT evaluation with
  improved correlation with human judgments}. In
  \bibinfo{booktitle}{\emph{Proceedings of the acl workshop on intrinsic and
  extrinsic evaluation measures for machine translation and/or summarization}}.
  \bibinfo{pages}{65--72}.
\newblock


\bibitem[\protect\citeauthoryear{Bang, Noh, Kim, and Lee}{Bang
  et~al\mbox{.}}{2015}]%
        {Bang2015Example}
\bibfield{author}{\bibinfo{person}{Jeesoo Bang}, \bibinfo{person}{Hyungjong
  Noh}, \bibinfo{person}{Yonghee Kim}, {and} \bibinfo{person}{Gary~Geunbae
  Lee}.} \bibinfo{year}{2015}\natexlab{}.
\newblock \showarticletitle{Example-based chat-oriented dialogue system with
  personalized long-term memory}. In \bibinfo{booktitle}{\emph{2015
  International Conference on Big Data and Smart Computing, {BIGCOMP} 2015,
  Jeju, South Korea, February 9-11, 2015}}. \bibinfo{pages}{238--243}.
\newblock


\bibitem[\protect\citeauthoryear{Bordes, Boureau, and Weston}{Bordes
  et~al\mbox{.}}{2017}]%
        {bordes2016learning}
\bibfield{author}{\bibinfo{person}{Antoine Bordes}, \bibinfo{person}{Y.{-}Lan
  Boureau}, {and} \bibinfo{person}{Jason Weston}.}
  \bibinfo{year}{2017}\natexlab{}.
\newblock \showarticletitle{Learning End-to-End Goal-Oriented Dialog}. In
  \bibinfo{booktitle}{\emph{{ICLR} 2017, Toulon, France, April 24-26, 2017}}.
\newblock


\bibitem[\protect\citeauthoryear{Boussaha, Hernandez, Jacquin, and
  Morin}{Boussaha et~al\mbox{.}}{2019}]%
        {boussaha2019deep}
\bibfield{author}{\bibinfo{person}{Basma El~Amel Boussaha},
  \bibinfo{person}{Nicolas Hernandez}, \bibinfo{person}{Christine Jacquin},
  {and} \bibinfo{person}{Emmanuel Morin}.} \bibinfo{year}{2019}\natexlab{}.
\newblock \showarticletitle{Deep Retrieval-Based Dialogue Systems: A Short
  Review}.
\newblock \bibinfo{journal}{\emph{arXiv preprint arXiv:1907.12878}}
  (\bibinfo{year}{2019}).
\newblock


\bibitem[\protect\citeauthoryear{Budzianowski and Vuli{\'c}}{Budzianowski and
  Vuli{\'c}}{2019}]%
        {budzianowski2019hello}
\bibfield{author}{\bibinfo{person}{Pawe{\l} Budzianowski} {and}
  \bibinfo{person}{Ivan Vuli{\'c}}.} \bibinfo{year}{2019}\natexlab{}.
\newblock \showarticletitle{Hello, It's GPT-2--How Can I Help You? Towards the
  Use of Pretrained Language Models for Task-Oriented Dialogue Systems}.
\newblock \bibinfo{journal}{\emph{CoRR}}  \bibinfo{volume}{abs/1907.05774}
  (\bibinfo{year}{2019}).
\newblock


\bibitem[\protect\citeauthoryear{Cao, Tanana, Imel, Poitras, Atkins, and
  Srikumar}{Cao et~al\mbox{.}}{2019}]%
        {cao2019observing}
\bibfield{author}{\bibinfo{person}{Jie Cao}, \bibinfo{person}{Michael Tanana},
  \bibinfo{person}{Zac~E Imel}, \bibinfo{person}{Eric Poitras},
  \bibinfo{person}{David~C Atkins}, {and} \bibinfo{person}{Vivek Srikumar}.}
  \bibinfo{year}{2019}\natexlab{}.
\newblock \showarticletitle{Observing Dialogue in Therapy: Categorizing and
  Forecasting Behavioral Codes}.
\newblock \bibinfo{journal}{\emph{arXiv preprint arXiv:1907.00326}}
  (\bibinfo{year}{2019}).
\newblock


\bibitem[\protect\citeauthoryear{Carpenter}{Carpenter}{2011}]%
        {carpenter2011cleverbot}
\bibfield{author}{\bibinfo{person}{Rollo Carpenter}.}
  \bibinfo{year}{2011}\natexlab{}.
\newblock \bibinfo{title}{Cleverbot}.
\newblock
\newblock


\bibitem[\protect\citeauthoryear{Casanueva, Hain, Christensen, Marxer, and
  Green}{Casanueva et~al\mbox{.}}{2015}]%
        {Casanueva2015Knowledge}
\bibfield{author}{\bibinfo{person}{I{\~{n}}igo Casanueva},
  \bibinfo{person}{Thomas Hain}, \bibinfo{person}{Heidi Christensen},
  \bibinfo{person}{Ricard Marxer}, {and} \bibinfo{person}{Phil~D. Green}.}
  \bibinfo{year}{2015}\natexlab{}.
\newblock \showarticletitle{Knowledge transfer between speakers for
  personalised dialogue management}. In \bibinfo{booktitle}{\emph{Proceedings
  of {SIGDIAL} 2015, September 2-4, 2015, Prague, Czech Republic}}.
  \bibinfo{pages}{12--21}.
\newblock


\bibitem[\protect\citeauthoryear{Chaganty, Mussmann, and Liang}{Chaganty
  et~al\mbox{.}}{2018}]%
        {chaganty2018price}
\bibfield{author}{\bibinfo{person}{Arun Chaganty}, \bibinfo{person}{Stephen
  Mussmann}, {and} \bibinfo{person}{Percy Liang}.}
  \bibinfo{year}{2018}\natexlab{}.
\newblock \showarticletitle{The price of debiasing automatic metrics in natural
  language evalaution}. In \bibinfo{booktitle}{\emph{Proceedings of {ACL} 2018,
  Melbourne, Australia, July 15-20, 2018}}. \bibinfo{pages}{643--653}.
\newblock


\bibitem[\protect\citeauthoryear{Chen, Yu, Wen, Yang, Zhang, Zhou, Jesse, Chau,
  Bhowmick, Iyer, et~al\mbox{.}}{Chen et~al\mbox{.}}{2018}]%
        {chen2018gunrock}
\bibfield{author}{\bibinfo{person}{Chun-Yen Chen}, \bibinfo{person}{Dian Yu},
  \bibinfo{person}{Weiming Wen}, \bibinfo{person}{Yi~Mang Yang},
  \bibinfo{person}{Jiaping Zhang}, \bibinfo{person}{Mingyang Zhou},
  \bibinfo{person}{Kevin Jesse}, \bibinfo{person}{Austin Chau},
  \bibinfo{person}{Antara Bhowmick}, \bibinfo{person}{Shreenath Iyer},
  {et~al\mbox{.}}} \bibinfo{year}{2018}\natexlab{}.
\newblock \showarticletitle{Gunrock: Building A Human-Like Social Bot By
  Leveraging Large Scale Real User Data}.
\newblock \bibinfo{journal}{\emph{Alexa prize proceedings}}
  (\bibinfo{year}{2018}).
\newblock


\bibitem[\protect\citeauthoryear{Chen and Lee}{Chen and Lee}{2019}]%
        {chendykgchat}
\bibfield{author}{\bibinfo{person}{Yi-Lin Tuan Yun-Nung Chen} {and}
  \bibinfo{person}{Hung-yi Lee}.} \bibinfo{year}{2019}\natexlab{}.
\newblock \showarticletitle{DyKgChat: Benchmarking Dialogue Generation
  Grounding on Dynamic Knowledge Graphs}. In
  \bibinfo{booktitle}{\emph{Proceedings of {EMNLP} 2019, Hong Kong, China,
  November 3-7, 2019}}.
\newblock


\bibitem[\protect\citeauthoryear{Colby, Weber, and Hilf}{Colby
  et~al\mbox{.}}{1971}]%
        {colby1971artificial}
\bibfield{author}{\bibinfo{person}{Kenneth~Mark Colby}, \bibinfo{person}{Sylvia
  Weber}, {and} \bibinfo{person}{Franklin~Dennis Hilf}.}
  \bibinfo{year}{1971}\natexlab{}.
\newblock \showarticletitle{Artificial paranoia}.
\newblock \bibinfo{journal}{\emph{Artificial Intelligence}}
  \bibinfo{volume}{2}, \bibinfo{number}{1} (\bibinfo{year}{1971}),
  \bibinfo{pages}{1--25}.
\newblock


\bibitem[\protect\citeauthoryear{Colombo, Witon, Modi, Kennedy, and
  Kapadia}{Colombo et~al\mbox{.}}{2019}]%
        {colombo2019affect}
\bibfield{author}{\bibinfo{person}{Pierre Colombo}, \bibinfo{person}{Wojciech
  Witon}, \bibinfo{person}{Ashutosh Modi}, \bibinfo{person}{James Kennedy},
  {and} \bibinfo{person}{Mubbasir Kapadia}.} \bibinfo{year}{2019}\natexlab{}.
\newblock \showarticletitle{Affect-Driven Dialog Generation}. In
  \bibinfo{booktitle}{\emph{{NAACL} 2019, Minneapolis, USA, June 2-7, 2019}}.
  \bibinfo{pages}{3734--3743}.
\newblock


\bibitem[\protect\citeauthoryear{Corston-Oliver, Gamon, and
  Brockett}{Corston-Oliver et~al\mbox{.}}{2001}]%
        {corston01ml}
\bibfield{author}{\bibinfo{person}{Simon Corston-Oliver},
  \bibinfo{person}{Michael Gamon}, {and} \bibinfo{person}{Chris Brockett}.}
  \bibinfo{year}{2001}\natexlab{}.
\newblock \showarticletitle{A Machine Learning Approach to the Automatic
  Evaluation of Machine Translation}. In \bibinfo{booktitle}{\emph{Proceedings
  of {ACL} 2001, July 9-11, 2001, Toulouse, France}}.
  \bibinfo{pages}{148--155}.
\newblock


\bibitem[\protect\citeauthoryear{Cui, Wang, Song, Huang, Xu, and Nie}{Cui
  et~al\mbox{.}}{2019}]%
        {cui2019user}
\bibfield{author}{\bibinfo{person}{Chen Cui}, \bibinfo{person}{Wenjie Wang},
  \bibinfo{person}{Xuemeng Song}, \bibinfo{person}{Minlie Huang},
  \bibinfo{person}{Xin-Shun Xu}, {and} \bibinfo{person}{Liqiang Nie}.}
  \bibinfo{year}{2019}\natexlab{}.
\newblock \showarticletitle{User Attention-guided Multimodal Dialog Systems}.
\newblock  (\bibinfo{year}{2019}).
\newblock


\bibitem[\protect\citeauthoryear{Dagan, Glickman, and Magnini}{Dagan
  et~al\mbox{.}}{2005}]%
        {dagan2005pascal}
\bibfield{author}{\bibinfo{person}{Ido Dagan}, \bibinfo{person}{Oren Glickman},
  {and} \bibinfo{person}{Bernardo Magnini}.} \bibinfo{year}{2005}\natexlab{}.
\newblock \showarticletitle{The PASCAL recognising textual entailment
  challenge}. In \bibinfo{booktitle}{\emph{Machine Learning Challenges
  Workshop}}. Springer, \bibinfo{pages}{177--190}.
\newblock


\bibitem[\protect\citeauthoryear{Devlin, Chang, Lee, and Toutanova}{Devlin
  et~al\mbox{.}}{2019}]%
        {devlin2018bert}
\bibfield{author}{\bibinfo{person}{Jacob Devlin}, \bibinfo{person}{Ming-Wei
  Chang}, \bibinfo{person}{Kenton Lee}, {and} \bibinfo{person}{Kristina
  Toutanova}.} \bibinfo{year}{2019}\natexlab{}.
\newblock \showarticletitle{Bert: Pre-training of deep bidirectional
  transformers for language understanding}.
\newblock \bibinfo{journal}{\emph{NAACL}} (\bibinfo{year}{2019}),
  \bibinfo{pages}{4171--4186}.
\newblock


\bibitem[\protect\citeauthoryear{Dinan, Logacheva, Malykh, Miller, Shuster,
  Urbanek, Kiela, Szlam, Serban, Lowe, Prabhumoye, Black, Rudnicky, Williams,
  Pineau, Burtsev, and Weston}{Dinan et~al\mbox{.}}{2019}]%
        {dinan2019second}
\bibfield{author}{\bibinfo{person}{Emily Dinan}, \bibinfo{person}{Varvara
  Logacheva}, \bibinfo{person}{Valentin Malykh}, \bibinfo{person}{Alexander~H.
  Miller}, \bibinfo{person}{Kurt Shuster}, \bibinfo{person}{Jack Urbanek},
  \bibinfo{person}{Douwe Kiela}, \bibinfo{person}{Arthur Szlam},
  \bibinfo{person}{Iulian Serban}, \bibinfo{person}{Ryan Lowe},
  \bibinfo{person}{Shrimai Prabhumoye}, \bibinfo{person}{Alan~W. Black},
  \bibinfo{person}{Alexander~I. Rudnicky}, \bibinfo{person}{Jason Williams},
  \bibinfo{person}{Joelle Pineau}, \bibinfo{person}{Mikhail Burtsev}, {and}
  \bibinfo{person}{Jason Weston}.} \bibinfo{year}{2019}\natexlab{}.
\newblock \showarticletitle{The Second Conversational Intelligence Challenge
  (ConvAI2)}.
\newblock \bibinfo{journal}{\emph{CoRR}}  \bibinfo{volume}{abs/1902.00098}
  (\bibinfo{year}{2019}).
\newblock


\bibitem[\protect\citeauthoryear{Dinan, Roller, Shuster, Fan, Auli, and
  Weston}{Dinan et~al\mbox{.}}{2018}]%
        {dinan2018WOW}
\bibfield{author}{\bibinfo{person}{Emily Dinan}, \bibinfo{person}{Stephen
  Roller}, \bibinfo{person}{Kurt Shuster}, \bibinfo{person}{Angela Fan},
  \bibinfo{person}{Michael Auli}, {and} \bibinfo{person}{Jason Weston}.}
  \bibinfo{year}{2018}\natexlab{}.
\newblock \showarticletitle{Wizard of Wikipedia: Knowledge-Powered
  Conversational agents}.
\newblock \bibinfo{journal}{\emph{CoRR}}  \bibinfo{volume}{abs/1811.01241}
  (\bibinfo{year}{2018}).
\newblock


\bibitem[\protect\citeauthoryear{Du, Li, He, Xu, Bing, and Wang}{Du
  et~al\mbox{.}}{2018}]%
        {du18vae}
\bibfield{author}{\bibinfo{person}{Jiachen Du}, \bibinfo{person}{Wenjie Li},
  \bibinfo{person}{Yulan He}, \bibinfo{person}{Ruifeng Xu},
  \bibinfo{person}{Lidong Bing}, {and} \bibinfo{person}{Xuan Wang}.}
  \bibinfo{year}{2018}\natexlab{}.
\newblock \showarticletitle{Variational Autoregressive Decoder for Neural
  Response Generation}. In \bibinfo{booktitle}{\emph{Proceedings of {EMNLP}
  2018, Brussels, Belgium, October 31 - November 4, 2018}}.
  \bibinfo{pages}{3154--3163}.
\newblock


\bibitem[\protect\citeauthoryear{Dziri, Kamalloo, Mathewson, and Zaiane}{Dziri
  et~al\mbox{.}}{2019}]%
        {dziri2019NLI}
\bibfield{author}{\bibinfo{person}{Nouha Dziri}, \bibinfo{person}{Ehsan
  Kamalloo}, \bibinfo{person}{Kory Mathewson}, {and} \bibinfo{person}{Osmar~R
  Zaiane}.} \bibinfo{year}{2019}\natexlab{}.
\newblock \showarticletitle{Evaluating Coherence in Dialogue Systems using
  Entailment}. In \bibinfo{booktitle}{\emph{{NAACL} 2019, Minneapolis, USA,
  June 2-7, 2019}}. \bibinfo{pages}{3806--3812}.
\newblock


\bibitem[\protect\citeauthoryear{Fan, Pang, Hou, Guo, Lan, and Cheng}{Fan
  et~al\mbox{.}}{2017}]%
        {fan2017matchzoo}
\bibfield{author}{\bibinfo{person}{Yixing Fan}, \bibinfo{person}{Liang Pang},
  \bibinfo{person}{Jianpeng Hou}, \bibinfo{person}{Jiafeng Guo},
  \bibinfo{person}{Yanyan Lan}, {and} \bibinfo{person}{Xueqi Cheng}.}
  \bibinfo{year}{2017}\natexlab{}.
\newblock \showarticletitle{MatchZoo: {A} Toolkit for Deep Text Matching}.
\newblock \bibinfo{journal}{\emph{CoRR}}  \bibinfo{volume}{abs/1707.07270}
  (\bibinfo{year}{2017}).
\newblock


\bibitem[\protect\citeauthoryear{Fang, Cheng, Clark, Holtzman, Sap, Ostendorf,
  Choi, and Smith}{Fang et~al\mbox{.}}{2017}]%
        {fang2017sounding}
\bibfield{author}{\bibinfo{person}{Hao Fang}, \bibinfo{person}{Hao Cheng},
  \bibinfo{person}{Elizabeth Clark}, \bibinfo{person}{Ariel Holtzman},
  \bibinfo{person}{Maarten Sap}, \bibinfo{person}{Mari Ostendorf},
  \bibinfo{person}{Yejin Choi}, {and} \bibinfo{person}{Noah~A Smith}.}
  \bibinfo{year}{2017}\natexlab{}.
\newblock \showarticletitle{Sounding board--university of washington's alexa
  prize submission}.
\newblock \bibinfo{journal}{\emph{Alexa prize proceedings}}
  (\bibinfo{year}{2017}).
\newblock


\bibitem[\protect\citeauthoryear{Fang, Cheng, Sap, Clark, Holtzman, Choi,
  Smith, and Ostendorf}{Fang et~al\mbox{.}}{2018}]%
        {sounding2017}
\bibfield{author}{\bibinfo{person}{Hao Fang}, \bibinfo{person}{Hao Cheng},
  \bibinfo{person}{Maarten Sap}, \bibinfo{person}{Elizabeth Clark},
  \bibinfo{person}{Ari Holtzman}, \bibinfo{person}{Yejin Choi},
  \bibinfo{person}{Noah~A. Smith}, {and} \bibinfo{person}{Mari Ostendorf}.}
  \bibinfo{year}{2018}\natexlab{}.
\newblock \showarticletitle{Sounding Board: {A} User-Centric and Content-Driven
  Social Chatbot}. In \bibinfo{booktitle}{\emph{Proceedings of {NAACL-HLT}
  2018, New Orleans, Louisiana, USA, June 2-4, 2018, Demonstrations}}.
  \bibinfo{pages}{96--100}.
\newblock


\bibitem[\protect\citeauthoryear{Galley, Brockett, Gao, Gao, and Dolan}{Galley
  et~al\mbox{.}}{2019}]%
        {galley2019grounded}
\bibfield{author}{\bibinfo{person}{Michel Galley}, \bibinfo{person}{Chris
  Brockett}, \bibinfo{person}{Xiang Gao}, \bibinfo{person}{Jianfeng Gao}, {and}
  \bibinfo{person}{Bill Dolan}.} \bibinfo{year}{2019}\natexlab{}.
\newblock \showarticletitle{Grounded Response Generation Task at DSTC7}. In
  \bibinfo{booktitle}{\emph{AAAI Dialog System Technology Challenges
  Workshop}}.
\newblock


\bibitem[\protect\citeauthoryear{Galley, Brockett, Sordoni, Ji, Auli, Quirk,
  Mitchell, Gao, and Dolan}{Galley et~al\mbox{.}}{2015}]%
        {galley2015deltableu}
\bibfield{author}{\bibinfo{person}{Michel Galley}, \bibinfo{person}{Chris
  Brockett}, \bibinfo{person}{Alessandro Sordoni}, \bibinfo{person}{Yangfeng
  Ji}, \bibinfo{person}{Michael Auli}, \bibinfo{person}{Chris Quirk},
  \bibinfo{person}{Margaret Mitchell}, \bibinfo{person}{Jianfeng Gao}, {and}
  \bibinfo{person}{Bill Dolan}.} \bibinfo{year}{2015}\natexlab{}.
\newblock \showarticletitle{{deltaBLEU}: A Discriminative Metric for Generation
  Tasks with Intrinsically Diverse Targets}. In
  \bibinfo{booktitle}{\emph{Proceedings of {ACL-IJCNLP} 2015, July 26-31, 2015,
  Beijing, China}}. \bibinfo{pages}{445--450}.
\newblock


\bibitem[\protect\citeauthoryear{Gao, Galley, and Li}{Gao
  et~al\mbox{.}}{2019a}]%
        {gao2019neural}
\bibfield{author}{\bibinfo{person}{Jianfeng Gao}, \bibinfo{person}{Michel
  Galley}, {and} \bibinfo{person}{Lihong Li}.}
  \bibinfo{year}{2019}\natexlab{a}.
\newblock \showarticletitle{Neural approaches to conversational AI}.
\newblock \bibinfo{journal}{\emph{Foundations and Trends{\textregistered} in
  Information Retrieval}} \bibinfo{volume}{13}, \bibinfo{number}{2-3}
  (\bibinfo{year}{2019}), \bibinfo{pages}{127--298}.
\newblock


\bibitem[\protect\citeauthoryear{Gao, Pantel, Gamon, He, and Deng}{Gao
  et~al\mbox{.}}{2014}]%
        {gao2014modeling}
\bibfield{author}{\bibinfo{person}{Jianfeng Gao}, \bibinfo{person}{Patrick
  Pantel}, \bibinfo{person}{Michael Gamon}, \bibinfo{person}{Xiaodong He},
  {and} \bibinfo{person}{Li Deng}.} \bibinfo{year}{2014}\natexlab{}.
\newblock \showarticletitle{Modeling interestingness with deep neural
  networks}. In \bibinfo{booktitle}{\emph{Proceedings of the 2014 Conference on
  Empirical Methods in Natural Language Processing (EMNLP)}}.
  \bibinfo{pages}{2--13}.
\newblock


\bibitem[\protect\citeauthoryear{Gao, Lee, Zhang, Brockett, Galley, Gao, and
  Dolan}{Gao et~al\mbox{.}}{2019b}]%
        {gao2019jointly}
\bibfield{author}{\bibinfo{person}{Xiang Gao}, \bibinfo{person}{Sungjin Lee},
  \bibinfo{person}{Yizhe Zhang}, \bibinfo{person}{Chris Brockett},
  \bibinfo{person}{Michel Galley}, \bibinfo{person}{Jianfeng Gao}, {and}
  \bibinfo{person}{Bill Dolan}.} \bibinfo{year}{2019}\natexlab{b}.
\newblock \showarticletitle{Jointly Optimizing Diversity and Relevance in
  Neural Response Generation}.
\newblock \bibinfo{journal}{\emph{arXiv preprint arXiv:1902.11205}}
  (\bibinfo{year}{2019}).
\newblock


\bibitem[\protect\citeauthoryear{Ghazvininejad, Brockett, Chang, Dolan, Gao,
  Yih, and Galley}{Ghazvininejad et~al\mbox{.}}{2018}]%
        {ghazvininejad2017knowledge}
\bibfield{author}{\bibinfo{person}{Marjan Ghazvininejad},
  \bibinfo{person}{Chris Brockett}, \bibinfo{person}{Ming{-}Wei Chang},
  \bibinfo{person}{Bill Dolan}, \bibinfo{person}{Jianfeng Gao},
  \bibinfo{person}{Wen{-}tau Yih}, {and} \bibinfo{person}{Michel Galley}.}
  \bibinfo{year}{2018}\natexlab{}.
\newblock \showarticletitle{A Knowledge-Grounded Neural Conversation Model}. In
  \bibinfo{booktitle}{\emph{Proceedings of {AAAI} 2018, New Orleans, Louisiana,
  USA, February 2-7, 2018}}. \bibinfo{pages}{5110--5117}.
\newblock


\bibitem[\protect\citeauthoryear{Ghosh, Chollet, Laksana, Morency, and
  Scherer}{Ghosh et~al\mbox{.}}{2017}]%
        {ghosh2017affect}
\bibfield{author}{\bibinfo{person}{Sayan Ghosh}, \bibinfo{person}{Mathieu
  Chollet}, \bibinfo{person}{Eugene Laksana}, \bibinfo{person}{Louis{-}Philippe
  Morency}, {and} \bibinfo{person}{Stefan Scherer}.}
  \bibinfo{year}{2017}\natexlab{}.
\newblock \showarticletitle{Affect-LM: {A} Neural Language Model for
  Customizable Affective Text Generation}. In
  \bibinfo{booktitle}{\emph{Proceedings of {ACL} 2017, Vancouver, Canada, July
  30-August 4, 2017}}. \bibinfo{pages}{634--642}.
\newblock


\bibitem[\protect\citeauthoryear{Gim\'{e}nez and M\`{a}rquez}{Gim\'{e}nez and
  M\`{a}rquez}{2008}]%
        {gimenez08smor}
\bibfield{author}{\bibinfo{person}{Jes\'{u}s Gim\'{e}nez} {and}
  \bibinfo{person}{Llu\'{i}s M\`{a}rquez}.} \bibinfo{year}{2008}\natexlab{}.
\newblock \showarticletitle{A Smorgasbord of Features for Automatic {MT}
  Evaluation}. In \bibinfo{booktitle}{\emph{Proceedings of the Third Workshop
  on Statistical Machine Translation}}. \bibinfo{pages}{195--198}.
\newblock


\bibitem[\protect\citeauthoryear{Golovanov, Kurbanov, Nikolenko, Truskovskyi,
  Tselousov, and Wolf}{Golovanov et~al\mbox{.}}{2019}]%
        {golovanov-etal-2019-large}
\bibfield{author}{\bibinfo{person}{Sergey Golovanov}, \bibinfo{person}{Rauf
  Kurbanov}, \bibinfo{person}{Sergey Nikolenko}, \bibinfo{person}{Kyryl
  Truskovskyi}, \bibinfo{person}{Alexander Tselousov}, {and}
  \bibinfo{person}{Thomas Wolf}.} \bibinfo{year}{2019}\natexlab{}.
\newblock \showarticletitle{Large-Scale Transfer Learning for Natural Language
  Generation}. In \bibinfo{booktitle}{\emph{ACL}}. \bibinfo{pages}{6053--6058}.
\newblock
\urldef\tempurl%
\url{https://www.aclweb.org/anthology/P19-1608}
\showURL{%
\tempurl}


\bibitem[\protect\citeauthoryear{Gopalakrishnan, Hedayatnia, Chen, Gottardi,
  Kwatra, Venkatesh, Gabriel, Hakkani-T{\"u}r, and AI}{Gopalakrishnan
  et~al\mbox{.}}{2019}]%
        {gopalakrishnan2019topical}
\bibfield{author}{\bibinfo{person}{Karthik Gopalakrishnan},
  \bibinfo{person}{Behnam Hedayatnia}, \bibinfo{person}{Qinlang Chen},
  \bibinfo{person}{Anna Gottardi}, \bibinfo{person}{Sanjeev Kwatra},
  \bibinfo{person}{Anu Venkatesh}, \bibinfo{person}{Raefer Gabriel},
  \bibinfo{person}{Dilek Hakkani-T{\"u}r}, {and} \bibinfo{person}{Amazon~Alexa
  AI}.} \bibinfo{year}{2019}\natexlab{}.
\newblock \showarticletitle{Topical-Chat: Towards Knowledge-Grounded
  Open-Domain Conversations}.
\newblock \bibinfo{journal}{\emph{Proc. Interspeech 2019}}
  (\bibinfo{year}{2019}), \bibinfo{pages}{1891--1895}.
\newblock


\bibitem[\protect\citeauthoryear{Gosling, Rentfrow, and Swann}{Gosling
  et~al\mbox{.}}{2003}]%
        {gosling2003very}
\bibfield{author}{\bibinfo{person}{Samuel~D Gosling}, \bibinfo{person}{Peter~J
  Rentfrow}, {and} \bibinfo{person}{William~B Swann}.}
  \bibinfo{year}{2003}\natexlab{}.
\newblock \showarticletitle{A very brief measure of the Big-Five personality
  domains}.
\newblock \bibinfo{journal}{\emph{Journal of Research in personality}}
  \bibinfo{volume}{37}, \bibinfo{number}{6} (\bibinfo{year}{2003}),
  \bibinfo{pages}{504--528}.
\newblock


\bibitem[\protect\citeauthoryear{Gu, Bradbury, Xiong, Li, and Socher}{Gu
  et~al\mbox{.}}{2017}]%
        {gu2017NAD}
\bibfield{author}{\bibinfo{person}{Jiatao Gu}, \bibinfo{person}{James
  Bradbury}, \bibinfo{person}{Caiming Xiong}, \bibinfo{person}{Victor O.~K.
  Li}, {and} \bibinfo{person}{Richard Socher}.}
  \bibinfo{year}{2017}\natexlab{}.
\newblock \showarticletitle{Non-Autoregressive Neural Machine Translation}.
\newblock \bibinfo{journal}{\emph{CoRR}}  \bibinfo{volume}{abs/1711.02281}
  (\bibinfo{year}{2017}).
\newblock
\showeprint[arxiv]{1711.02281}
\urldef\tempurl%
\url{http://arxiv.org/abs/1711.02281}
\showURL{%
\tempurl}


\bibitem[\protect\citeauthoryear{Gu, Lu, Li, and Li}{Gu et~al\mbox{.}}{2016}]%
        {gu2016copy}
\bibfield{author}{\bibinfo{person}{Jiatao Gu}, \bibinfo{person}{Zhengdong Lu},
  \bibinfo{person}{Hang Li}, {and} \bibinfo{person}{Victor O.~K. Li}.}
  \bibinfo{year}{2016}\natexlab{}.
\newblock \showarticletitle{Incorporating Copying Mechanism in
  Sequence-to-Sequence Learning}. In \bibinfo{booktitle}{\emph{Proceedings of
  {ACL} 2016, Berlin, Germany, August 7-12, 2016}}.
\newblock


\bibitem[\protect\citeauthoryear{Han, Bang, Ryu, and Lee}{Han
  et~al\mbox{.}}{2015}]%
        {han2015exploiting}
\bibfield{author}{\bibinfo{person}{Sangdo Han}, \bibinfo{person}{Jeesoo Bang},
  \bibinfo{person}{Seonghan Ryu}, {and} \bibinfo{person}{Gary~Geunbae Lee}.}
  \bibinfo{year}{2015}\natexlab{}.
\newblock \showarticletitle{Exploiting knowledge base to generate responses for
  natural language dialog listening agents}. In
  \bibinfo{booktitle}{\emph{Proceedings of {SIGDIAL} 2015, September 2-4, 2015,
  Prague, Czech Republic}}. \bibinfo{pages}{129--133}.
\newblock


\bibitem[\protect\citeauthoryear{Hashimoto, Zhang, and Liang}{Hashimoto
  et~al\mbox{.}}{2019}]%
        {hashimoto2019unifying}
\bibfield{author}{\bibinfo{person}{Tatsunori Hashimoto}, \bibinfo{person}{Hugh
  Zhang}, {and} \bibinfo{person}{Percy Liang}.}
  \bibinfo{year}{2019}\natexlab{}.
\newblock \showarticletitle{Unifying Human and Statistical Evaluation for
  Natural Language Generation}. In \bibinfo{booktitle}{\emph{{NAACL} 2019,
  Minneapolis, USA, June 2-7, 2019}}. \bibinfo{pages}{1689--1701}.
\newblock


\bibitem[\protect\citeauthoryear{Henderson, Casanueva, Mrk{\v{s}}i{\'c}, Su,
  Vuli{\'c}, et~al\mbox{.}}{Henderson et~al\mbox{.}}{2019}]%
        {henderson2019convert}
\bibfield{author}{\bibinfo{person}{Matthew Henderson},
  \bibinfo{person}{I{\~n}igo Casanueva}, \bibinfo{person}{Nikola
  Mrk{\v{s}}i{\'c}}, \bibinfo{person}{Pei-Hao Su}, \bibinfo{person}{Ivan
  Vuli{\'c}}, {et~al\mbox{.}}} \bibinfo{year}{2019}\natexlab{}.
\newblock \showarticletitle{ConveRT: Efficient and Accurate Conversational
  Representations from Transformers}.
\newblock \bibinfo{journal}{\emph{arXiv preprint arXiv:1911.03688}}
  (\bibinfo{year}{2019}).
\newblock


\bibitem[\protect\citeauthoryear{Henderson, Thomson, and Young}{Henderson
  et~al\mbox{.}}{2013}]%
        {henderson2013deep}
\bibfield{author}{\bibinfo{person}{Matthew Henderson}, \bibinfo{person}{Blaise
  Thomson}, {and} \bibinfo{person}{Steve~J. Young}.}
  \bibinfo{year}{2013}\natexlab{}.
\newblock \showarticletitle{Deep Neural Network Approach for the Dialog State
  Tracking Challenge}. In \bibinfo{booktitle}{\emph{Proceedings of {SIGDIAL}
  2013, August 22-24, 2013, SUPELEC, Metz, France}}. \bibinfo{pages}{467--471}.
\newblock


\bibitem[\protect\citeauthoryear{Higashinaka, Imamura, Meguro, Miyazaki,
  Kobayashi, Sugiyama, Hirano, Makino, and Matsuo}{Higashinaka
  et~al\mbox{.}}{2014}]%
        {a32}
\bibfield{author}{\bibinfo{person}{Ryuichiro Higashinaka},
  \bibinfo{person}{Kenji Imamura}, \bibinfo{person}{Toyomi Meguro},
  \bibinfo{person}{Chiaki Miyazaki}, \bibinfo{person}{Nozomi Kobayashi},
  \bibinfo{person}{Hiroaki Sugiyama}, \bibinfo{person}{Toru Hirano},
  \bibinfo{person}{Toshiro Makino}, {and} \bibinfo{person}{Yoshihiro Matsuo}.}
  \bibinfo{year}{2014}\natexlab{}.
\newblock \showarticletitle{Towards an open-domain conversational system fully
  based on natural language processing}. In \bibinfo{booktitle}{\emph{{COLING}
  2014, August 23-29, 2014, Dublin, Ireland}}. \bibinfo{pages}{928--939}.
\newblock


\bibitem[\protect\citeauthoryear{Hu, Lu, Li, and Chen}{Hu
  et~al\mbox{.}}{2014}]%
        {arc}
\bibfield{author}{\bibinfo{person}{Baotian Hu}, \bibinfo{person}{Zhengdong Lu},
  \bibinfo{person}{Hang Li}, {and} \bibinfo{person}{Qingcai Chen}.}
  \bibinfo{year}{2014}\natexlab{}.
\newblock \showarticletitle{Convolutional Neural Network Architectures for
  Matching Natural Language Sentences}. In \bibinfo{booktitle}{\emph{{NIPS}
  2014, Montreal, Quebec, Canada, December 8-13, 2014}}.
  \bibinfo{pages}{2042--2050}.
\newblock


\bibitem[\protect\citeauthoryear{Hu, Yang, Liang, Salakhutdinov, and Xing}{Hu
  et~al\mbox{.}}{2017}]%
        {hu2017controllable}
\bibfield{author}{\bibinfo{person}{Zhiting Hu}, \bibinfo{person}{Zichao Yang},
  \bibinfo{person}{Xiaodan Liang}, \bibinfo{person}{Ruslan Salakhutdinov},
  {and} \bibinfo{person}{Eric~P. Xing}.} \bibinfo{year}{2017}\natexlab{}.
\newblock \showarticletitle{Toward Controlled Generation of Text}. In
  \bibinfo{booktitle}{\emph{Proceedings of {ICML} 2017, Sydney, NSW, Australia,
  6-11 August 2017}}, Vol.~\bibinfo{volume}{70}. \bibinfo{pages}{1587--1596}.
\newblock


\bibitem[\protect\citeauthoryear{Huang, He, Gao, Deng, Acero, and Heck}{Huang
  et~al\mbox{.}}{2013}]%
        {Huang:2013fz}
\bibfield{author}{\bibinfo{person}{Po{-}Sen Huang}, \bibinfo{person}{Xiaodong
  He}, \bibinfo{person}{Jianfeng Gao}, \bibinfo{person}{Li Deng},
  \bibinfo{person}{Alex Acero}, {and} \bibinfo{person}{Larry~P. Heck}.}
  \bibinfo{year}{2013}\natexlab{}.
\newblock \showarticletitle{Learning deep structured semantic models for web
  search using clickthrough data}. In \bibinfo{booktitle}{\emph{22nd {ACM}
  International Conference on Information and Knowledge Management, CIKM'13,
  San Francisco, CA, USA, October 27 - November 1, 2013}}.
  \bibinfo{pages}{2333--2338}.
\newblock


\bibitem[\protect\citeauthoryear{Huber, McDuff, Brockett, Galley, and
  Dolan}{Huber et~al\mbox{.}}{2018}]%
        {huber2018emotional}
\bibfield{author}{\bibinfo{person}{Bernd Huber}, \bibinfo{person}{Daniel
  McDuff}, \bibinfo{person}{Chris Brockett}, \bibinfo{person}{Michel Galley},
  {and} \bibinfo{person}{Bill Dolan}.} \bibinfo{year}{2018}\natexlab{}.
\newblock \showarticletitle{Emotional Dialogue Generation using Image-Grounded
  Language Models}. In \bibinfo{booktitle}{\emph{Proceedings of the 2018 {CHI}
  Conference on Human Factors in Computing Systems, {CHI} 2018, Montreal, QC,
  Canada, April 21-26, 2018}}. \bibinfo{pages}{277}.
\newblock


\bibitem[\protect\citeauthoryear{Jafarpour, Burges, and Ritter}{Jafarpour
  et~al\mbox{.}}{2010}]%
        {jafarpour2010filter}
\bibfield{author}{\bibinfo{person}{Sina Jafarpour},
  \bibinfo{person}{Christopher~JC Burges}, {and} \bibinfo{person}{Alan
  Ritter}.} \bibinfo{year}{2010}\natexlab{}.
\newblock \showarticletitle{Filter, rank, and transfer the knowledge: Learning
  to chat}.
\newblock \bibinfo{journal}{\emph{Advances in Ranking}}  \bibinfo{volume}{10}
  (\bibinfo{year}{2010}), \bibinfo{pages}{2329--9290}.
\newblock


\bibitem[\protect\citeauthoryear{Ji, Lu, and Li}{Ji et~al\mbox{.}}{2014}]%
        {ji2014information}
\bibfield{author}{\bibinfo{person}{Zongcheng Ji}, \bibinfo{person}{Zhengdong
  Lu}, {and} \bibinfo{person}{Hang Li}.} \bibinfo{year}{2014}\natexlab{}.
\newblock \showarticletitle{An information retrieval approach to short text
  conversation}.
\newblock \bibinfo{journal}{\emph{arXiv preprint arXiv:1408.6988}}
  (\bibinfo{year}{2014}).
\newblock


\bibitem[\protect\citeauthoryear{Kaiser, Bengio, Roy, Vaswani, Parmar,
  Uszkoreit, and Shazeer}{Kaiser et~al\mbox{.}}{2018}]%
        {kaiser2018lt}
\bibfield{author}{\bibinfo{person}{Lukasz Kaiser}, \bibinfo{person}{Samy
  Bengio}, \bibinfo{person}{Aurko Roy}, \bibinfo{person}{Ashish Vaswani},
  \bibinfo{person}{Niki Parmar}, \bibinfo{person}{Jakob Uszkoreit}, {and}
  \bibinfo{person}{Noam Shazeer}.} \bibinfo{year}{2018}\natexlab{}.
\newblock \showarticletitle{Fast Decoding in Sequence Models Using Discrete
  Latent Variables}. In \bibinfo{booktitle}{\emph{Proceedings of {ICML} 2018,
  Stockholmsm{\"{a}}ssan, Stockholm, Sweden, July 10-15, 2018}},
  Vol.~\bibinfo{volume}{80}. \bibinfo{pages}{2395--2404}.
\newblock


\bibitem[\protect\citeauthoryear{Ke, Guan, Huang, and Zhu}{Ke
  et~al\mbox{.}}{2018}]%
        {ke2018senfunc}
\bibfield{author}{\bibinfo{person}{Pei Ke}, \bibinfo{person}{Jian Guan},
  \bibinfo{person}{Minlie Huang}, {and} \bibinfo{person}{Xiaoyan Zhu}.}
  \bibinfo{year}{2018}\natexlab{}.
\newblock \showarticletitle{Generating Informative Responses with Controlled
  Sentence Function}. In \bibinfo{booktitle}{\emph{Proceedings of {ACL} 2018,
  Melbourne, Australia, July 15-20, 2018}}. \bibinfo{pages}{1499--1508}.
\newblock


\bibitem[\protect\citeauthoryear{Kim, Bang, Choi, Ryu, Koo, and Lee}{Kim
  et~al\mbox{.}}{2014}]%
        {Kim2014Acquisition}
\bibfield{author}{\bibinfo{person}{Yonghee Kim}, \bibinfo{person}{Jeesoo Bang},
  \bibinfo{person}{Junhwi Choi}, \bibinfo{person}{Seonghan Ryu},
  \bibinfo{person}{Sangjun Koo}, {and} \bibinfo{person}{Gary~Geunbae Lee}.}
  \bibinfo{year}{2014}\natexlab{}.
\newblock \showarticletitle{Acquisition and Use of Long-Term Memory for
  Personalized Dialog Systems}. In \bibinfo{booktitle}{\emph{Multimodal
  Analyses enabling Artificial Agents in Human-Machine Interaction - Second
  International Workshop, {MA3HMI} 2014, Held in Conjunction with {INTERSPEECH}
  2014, Singapore, Singapore, September 14, 2014}},
  Vol.~\bibinfo{volume}{8757}. \bibinfo{pages}{78--87}.
\newblock


\bibitem[\protect\citeauthoryear{Kulesza and Shieber}{Kulesza and
  Shieber}{2004}]%
        {kulesza04learning}
\bibfield{author}{\bibinfo{person}{Alex Kulesza} {and}
  \bibinfo{person}{Stuart~M. Shieber}.} \bibinfo{year}{2004}\natexlab{}.
\newblock \showarticletitle{A Learning Approach to Improving Sentence-Level
  {MT} Evaluation}. In \bibinfo{booktitle}{\emph{Proceedings of the 10th
  International Conference on Theoretical and Methodological Issues in Machine
  Translation}}. \bibinfo{address}{Baltimore, MD}.
\newblock


\bibitem[\protect\citeauthoryear{Lee, Mansimov, and Cho}{Lee
  et~al\mbox{.}}{2018}]%
        {lee2018refine}
\bibfield{author}{\bibinfo{person}{Jason Lee}, \bibinfo{person}{Elman
  Mansimov}, {and} \bibinfo{person}{Kyunghyun Cho}.}
  \bibinfo{year}{2018}\natexlab{}.
\newblock \showarticletitle{Deterministic Non-Autoregressive Neural Sequence
  Modeling by Iterative Refinement}. In \bibinfo{booktitle}{\emph{Proceedings
  of {EMNLP} 2018, Brussels, Belgium, October 31 - November 4, 2018}}.
  \bibinfo{pages}{1173--1182}.
\newblock


\bibitem[\protect\citeauthoryear{Leuski and Traum}{Leuski and Traum}{2011}]%
        {leuski2011npceditor}
\bibfield{author}{\bibinfo{person}{Anton Leuski} {and} \bibinfo{person}{David
  Traum}.} \bibinfo{year}{2011}\natexlab{}.
\newblock \showarticletitle{NPCEditor: Creating virtual human dialogue using
  information retrieval techniques}.
\newblock \bibinfo{journal}{\emph{Ai Magazine}} \bibinfo{volume}{32},
  \bibinfo{number}{2} (\bibinfo{year}{2011}), \bibinfo{pages}{42--56}.
\newblock


\bibitem[\protect\citeauthoryear{Li, Galley, Brockett, Gao, and Dolan}{Li
  et~al\mbox{.}}{2016a}]%
        {lidiversity}
\bibfield{author}{\bibinfo{person}{Jiwei Li}, \bibinfo{person}{Michel Galley},
  \bibinfo{person}{Chris Brockett}, \bibinfo{person}{Jianfeng Gao}, {and}
  \bibinfo{person}{Bill Dolan}.} \bibinfo{year}{2016}\natexlab{a}.
\newblock \showarticletitle{A Diversity-Promoting Objective Function for Neural
  Conversation Models}. In \bibinfo{booktitle}{\emph{{NAACL} {HLT} 2016, San
  Diego California, USA, June 12-17, 2016}}. \bibinfo{pages}{110--119}.
\newblock


\bibitem[\protect\citeauthoryear{Li, Galley, Brockett, Spithourakis, Gao, and
  Dolan}{Li et~al\mbox{.}}{2016b}]%
        {Li2016_ACL-persona}
\bibfield{author}{\bibinfo{person}{Jiwei Li}, \bibinfo{person}{Michel Galley},
  \bibinfo{person}{Chris Brockett}, \bibinfo{person}{Georgios~P. Spithourakis},
  \bibinfo{person}{Jianfeng Gao}, {and} \bibinfo{person}{William~B. Dolan}.}
  \bibinfo{year}{2016}\natexlab{b}.
\newblock \showarticletitle{A Persona-Based Neural Conversation Model}. In
  \bibinfo{booktitle}{\emph{Proceedings of {ACL} 2016, Berlin, Germany, August
  7-12, 2016}}.
\newblock


\bibitem[\protect\citeauthoryear{Li, Monroe, and Jurafsky}{Li
  et~al\mbox{.}}{2016c}]%
        {li2016simple}
\bibfield{author}{\bibinfo{person}{Jiwei Li}, \bibinfo{person}{Will Monroe},
  {and} \bibinfo{person}{Dan Jurafsky}.} \bibinfo{year}{2016}\natexlab{c}.
\newblock \showarticletitle{A Simple, Fast Diverse Decoding Algorithm for
  Neural Generation}.
\newblock \bibinfo{journal}{\emph{CoRR}}  \bibinfo{volume}{abs/1611.08562}
  (\bibinfo{year}{2016}).
\newblock


\bibitem[\protect\citeauthoryear{Li, Monroe, Shi, Jean, Ritter, and
  Jurafsky}{Li et~al\mbox{.}}{2017a}]%
        {li2017adversarial}
\bibfield{author}{\bibinfo{person}{Jiwei Li}, \bibinfo{person}{Will Monroe},
  \bibinfo{person}{Tianlin Shi}, \bibinfo{person}{S{\'{e}}bastien Jean},
  \bibinfo{person}{Alan Ritter}, {and} \bibinfo{person}{Dan Jurafsky}.}
  \bibinfo{year}{2017}\natexlab{a}.
\newblock \showarticletitle{Adversarial Learning for Neural Dialogue
  Generation}. In \bibinfo{booktitle}{\emph{Proceedings of {EMNLP} 2017,
  Copenhagen, Denmark, September 9-11, 2017}}. \bibinfo{pages}{2157--2169}.
\newblock


\bibitem[\protect\citeauthoryear{Li, Mou, Yan, and Zhang}{Li
  et~al\mbox{.}}{2016d}]%
        {li2016stalematebreakerIRbased}
\bibfield{author}{\bibinfo{person}{Xiang Li}, \bibinfo{person}{Lili Mou},
  \bibinfo{person}{Rui Yan}, {and} \bibinfo{person}{Ming Zhang}.}
  \bibinfo{year}{2016}\natexlab{d}.
\newblock \showarticletitle{StalemateBreaker: {A} Proactive Content-Introducing
  Approach to Automatic Human-Computer Conversation}. In
  \bibinfo{booktitle}{\emph{Proceedings of {IJCAI} 2016, New York, NY, USA,
  9-15 July 2016}}. \bibinfo{pages}{2845--2851}.
\newblock


\bibitem[\protect\citeauthoryear{Li, Su, Shen, Li, Cao, and Niu}{Li
  et~al\mbox{.}}{2017b}]%
        {li2017dailydialog}
\bibfield{author}{\bibinfo{person}{Yanran Li}, \bibinfo{person}{Hui Su},
  \bibinfo{person}{Xiaoyu Shen}, \bibinfo{person}{Wenjie Li},
  \bibinfo{person}{Ziqiang Cao}, {and} \bibinfo{person}{Shuzi Niu}.}
  \bibinfo{year}{2017}\natexlab{b}.
\newblock \showarticletitle{DailyDialog: A Manually Labelled Multi-turn
  Dialogue Dataset}. In \bibinfo{booktitle}{\emph{Proceedings of the Eighth
  International Joint Conference on Natural Language Processing (Volume 1: Long
  Papers)}}. \bibinfo{pages}{986--995}.
\newblock


\bibitem[\protect\citeauthoryear{Li, Niu, Meng, Feng, Li, and Zhou}{Li
  et~al\mbox{.}}{2019}]%
        {li2019incremental}
\bibfield{author}{\bibinfo{person}{Zekang Li}, \bibinfo{person}{Cheng Niu},
  \bibinfo{person}{Fandong Meng}, \bibinfo{person}{Yang Feng},
  \bibinfo{person}{Qian Li}, {and} \bibinfo{person}{Jie Zhou}.}
  \bibinfo{year}{2019}\natexlab{}.
\newblock \showarticletitle{Incremental Transformer with Deliberation Decoder
  for Document Grounded Conversations}. In
  \bibinfo{booktitle}{\emph{Proceedings of {ACL} 2019, Florence, Italy, July 28
  - Aug 2, 2019}}. \bibinfo{pages}{12--21}.
\newblock


\bibitem[\protect\citeauthoryear{Lian, Xie, Wang, Peng, and Wu}{Lian
  et~al\mbox{.}}{2019}]%
        {lian2019learning}
\bibfield{author}{\bibinfo{person}{Rongzhong Lian}, \bibinfo{person}{Min Xie},
  \bibinfo{person}{Fan Wang}, \bibinfo{person}{Jinhua Peng}, {and}
  \bibinfo{person}{Hua Wu}.} \bibinfo{year}{2019}\natexlab{}.
\newblock \showarticletitle{Learning to select knowledge for response
  generation in dialog systems}.
\newblock \bibinfo{journal}{\emph{arXiv preprint arXiv:1902.04911}}
  (\bibinfo{year}{2019}).
\newblock


\bibitem[\protect\citeauthoryear{Liao, Ma, He, Hong, and Chua}{Liao
  et~al\mbox{.}}{2018}]%
        {liao2018knowledge}
\bibfield{author}{\bibinfo{person}{Lizi Liao}, \bibinfo{person}{Yunshan Ma},
  \bibinfo{person}{Xiangnan He}, \bibinfo{person}{Richang Hong}, {and}
  \bibinfo{person}{Tat-seng Chua}.} \bibinfo{year}{2018}\natexlab{}.
\newblock \showarticletitle{Knowledge-aware Multimodal Dialogue Systems}. In
  \bibinfo{booktitle}{\emph{2018 ACM Multimedia Conference on Multimedia
  Conference}}. \bibinfo{pages}{801--809}.
\newblock


\bibitem[\protect\citeauthoryear{Lin}{Lin}{2004}]%
        {lin2004rouge}
\bibfield{author}{\bibinfo{person}{Chin-Yew Lin}.}
  \bibinfo{year}{2004}\natexlab{}.
\newblock \showarticletitle{Rouge: A package for automatic evaluation of
  summaries}.
\newblock \bibinfo{journal}{\emph{Text Summarization Branches Out}}
  (\bibinfo{year}{2004}).
\newblock


\bibitem[\protect\citeauthoryear{Lipton, Li, Gao, Li, Ahmed, and Deng}{Lipton
  et~al\mbox{.}}{2018}]%
        {lipton2018bbq}
\bibfield{author}{\bibinfo{person}{Zachary~C. Lipton}, \bibinfo{person}{Xiujun
  Li}, \bibinfo{person}{Jianfeng Gao}, \bibinfo{person}{Lihong Li},
  \bibinfo{person}{Faisal Ahmed}, {and} \bibinfo{person}{Li Deng}.}
  \bibinfo{year}{2018}\natexlab{}.
\newblock \showarticletitle{BBQ-Networks: Efficient Exploration in Deep
  Reinforcement Learning for Task-Oriented Dialogue Systems}. In
  \bibinfo{booktitle}{\emph{Proceedings of {AAAI} 2018, New Orleans, Louisiana,
  USA, February 2-7, 2018}}. \bibinfo{pages}{5237--5244}.
\newblock


\bibitem[\protect\citeauthoryear{Lita, Rogati, and Lavie}{Lita
  et~al\mbox{.}}{2005}]%
        {lita05blanc}
\bibfield{author}{\bibinfo{person}{Lucian~Vlad Lita}, \bibinfo{person}{Monica
  Rogati}, {and} \bibinfo{person}{Alon Lavie}.}
  \bibinfo{year}{2005}\natexlab{}.
\newblock \showarticletitle{{BLANC}: Learning Evaluation Metrics for {MT}}. In
  \bibinfo{booktitle}{\emph{Proceedings of the Conference on Human Language
  Technology and Empirical Methods in Natural Language Processing}} (Vancouver,
  British Columbia, Canada) \emph{(\bibinfo{series}{HLT '05})}.
  \bibinfo{pages}{740--747}.
\newblock


\bibitem[\protect\citeauthoryear{Liu, Lowe, Serban, Noseworthy, Charlin, and
  Pineau}{Liu et~al\mbox{.}}{2016}]%
        {liu2016not}
\bibfield{author}{\bibinfo{person}{Chia{-}Wei Liu}, \bibinfo{person}{Ryan
  Lowe}, \bibinfo{person}{Iulian Serban}, \bibinfo{person}{Michael Noseworthy},
  \bibinfo{person}{Laurent Charlin}, {and} \bibinfo{person}{Joelle Pineau}.}
  \bibinfo{year}{2016}\natexlab{}.
\newblock \showarticletitle{How {NOT} To Evaluate Your Dialogue System: An
  Empirical Study of Unsupervised Evaluation Metrics for Dialogue Response
  Generation}. In \bibinfo{booktitle}{\emph{Proceedings of {EMNLP} 2016,
  Austin, Texas, USA, November 1-4, 2016}}. \bibinfo{pages}{2122--2132}.
\newblock


\bibitem[\protect\citeauthoryear{Liu, Chen, Ren, Feng, Liu, and Yin}{Liu
  et~al\mbox{.}}{2018a}]%
        {liu2018knowledge}
\bibfield{author}{\bibinfo{person}{Shuman Liu}, \bibinfo{person}{Hongshen
  Chen}, \bibinfo{person}{Zhaochun Ren}, \bibinfo{person}{Yang Feng},
  \bibinfo{person}{Qun Liu}, {and} \bibinfo{person}{Dawei Yin}.}
  \bibinfo{year}{2018}\natexlab{a}.
\newblock \showarticletitle{Knowledge Diffusion for Neural Dialogue
  Generation}. In \bibinfo{booktitle}{\emph{Proceedings of {ACL} 2018,
  Melbourne, Australia, July 15-20, 2018}}. \bibinfo{pages}{1489--1498}.
\newblock


\bibitem[\protect\citeauthoryear{Liu}{Liu}{2010}]%
        {liutieyan2009l2r}
\bibfield{author}{\bibinfo{person}{Tie{-}Yan Liu}.}
  \bibinfo{year}{2010}\natexlab{}.
\newblock \showarticletitle{Learning to rank for information retrieval}. In
  \bibinfo{booktitle}{\emph{Proceeding of {SIGIR} 2010, Geneva, Switzerland,
  July 19-23, 2010}}. \bibinfo{pages}{904}.
\newblock


\bibitem[\protect\citeauthoryear{Liu, Shen, Duh, and Gao}{Liu
  et~al\mbox{.}}{2018b}]%
        {liu2018stochastic}
\bibfield{author}{\bibinfo{person}{Xiaodong Liu}, \bibinfo{person}{Yelong
  Shen}, \bibinfo{person}{Kevin Duh}, {and} \bibinfo{person}{Jianfeng Gao}.}
  \bibinfo{year}{2018}\natexlab{b}.
\newblock \showarticletitle{Stochastic Answer Networks for Machine Reading
  Comprehension}. In \bibinfo{booktitle}{\emph{Proceedings of the 56th Annual
  Meeting of the Association for Computational Linguistics (Volume 1: Long
  Papers)}}. \bibinfo{pages}{1694--1704}.
\newblock


\bibitem[\protect\citeauthoryear{Liu, Niu, Wu, and Wang}{Liu
  et~al\mbox{.}}{2019}]%
        {liu2019knowledge}
\bibfield{author}{\bibinfo{person}{Zhibin Liu}, \bibinfo{person}{Zheng-Yu Niu},
  \bibinfo{person}{Hua Wu}, {and} \bibinfo{person}{Haifeng Wang}.}
  \bibinfo{year}{2019}\natexlab{}.
\newblock \showarticletitle{Knowledge Aware Conversation Generation with
  Reasoning on Augmented Graph}.
\newblock \bibinfo{journal}{\emph{arXiv preprint arXiv:1903.10245}}
  (\bibinfo{year}{2019}).
\newblock


\bibitem[\protect\citeauthoryear{Lowe, Noseworthy, Serban, Angelard{-}Gontier,
  Bengio, and Pineau}{Lowe et~al\mbox{.}}{2017}]%
        {lowe2017towards}
\bibfield{author}{\bibinfo{person}{Ryan Lowe}, \bibinfo{person}{Michael
  Noseworthy}, \bibinfo{person}{Iulian~Vlad Serban}, \bibinfo{person}{Nicolas
  Angelard{-}Gontier}, \bibinfo{person}{Yoshua Bengio}, {and}
  \bibinfo{person}{Joelle Pineau}.} \bibinfo{year}{2017}\natexlab{}.
\newblock \showarticletitle{Towards an Automatic Turing Test: Learning to
  Evaluate Dialogue Responses}. In \bibinfo{booktitle}{\emph{Proceedings of
  {ACL} 2017, Vancouver, Canada, July 30-August 4, 2017}}.
  \bibinfo{pages}{1116--1126}.
\newblock


\bibitem[\protect\citeauthoryear{Lowe, Pow, Serban, and Pineau}{Lowe
  et~al\mbox{.}}{2015}]%
        {lowe2015ubuntu}
\bibfield{author}{\bibinfo{person}{Ryan Lowe}, \bibinfo{person}{Nissan Pow},
  \bibinfo{person}{Iulian Serban}, {and} \bibinfo{person}{Joelle Pineau}.}
  \bibinfo{year}{2015}\natexlab{}.
\newblock \showarticletitle{The Ubuntu Dialogue Corpus: A Large Dataset for
  Research in Unstructured Multi-Turn Dialogue Systems}. In
  \bibinfo{booktitle}{\emph{Proceedings of the 16th Annual Meeting of the
  Special Interest Group on Discourse and Dialogue}}.
  \bibinfo{pages}{285--294}.
\newblock


\bibitem[\protect\citeauthoryear{Lu and Li}{Lu and Li}{2013}]%
        {lu2013deep}
\bibfield{author}{\bibinfo{person}{Zhengdong Lu} {and} \bibinfo{person}{Hang
  Li}.} \bibinfo{year}{2013}\natexlab{}.
\newblock \showarticletitle{A Deep Architecture for Matching Short Texts}. In
  \bibinfo{booktitle}{\emph{{NIPS} 2013, December 5-8, 2013, Lake Tahoe,
  Nevada, United States.}} \bibinfo{pages}{1367--1375}.
\newblock


\bibitem[\protect\citeauthoryear{Luan, Brockett, Dolan, Gao, and Galley}{Luan
  et~al\mbox{.}}{2017}]%
        {ar1}
\bibfield{author}{\bibinfo{person}{Yi Luan}, \bibinfo{person}{Chris Brockett},
  \bibinfo{person}{Bill Dolan}, \bibinfo{person}{Jianfeng Gao}, {and}
  \bibinfo{person}{Michel Galley}.} \bibinfo{year}{2017}\natexlab{}.
\newblock \showarticletitle{Multi-Task Learning for Speaker-Role Adaptation in
  Neural Conversation Models}. In \bibinfo{booktitle}{\emph{Proceedings of the
  Eighth International Joint Conference on Natural Language Processing,
  {IJCNLP} 2017, Taipei, Taiwan, November 27 - December 1, 2017 - Volume 1:
  Long Papers}}. \bibinfo{pages}{605--614}.
\newblock


\bibitem[\protect\citeauthoryear{Majumder, Poria, Hazarika, Mihalcea, Gelbukh,
  and Cambria}{Majumder et~al\mbox{.}}{2019}]%
        {majumder2019dialoguernn}
\bibfield{author}{\bibinfo{person}{Navonil Majumder}, \bibinfo{person}{Soujanya
  Poria}, \bibinfo{person}{Devamanyu Hazarika}, \bibinfo{person}{Rada
  Mihalcea}, \bibinfo{person}{Alexander Gelbukh}, {and} \bibinfo{person}{Erik
  Cambria}.} \bibinfo{year}{2019}\natexlab{}.
\newblock \showarticletitle{Dialoguernn: An attentive rnn for emotion detection
  in conversations}. In \bibinfo{booktitle}{\emph{Proceedings of {AAAI} 2019,
  Honolulu, Hawaii, USA, January 27-February 1, 2019}}.
  \bibinfo{pages}{6818--6825}.
\newblock


\bibitem[\protect\citeauthoryear{Manning, Raghavan, and Sch{\"{u}}tze}{Manning
  et~al\mbox{.}}{2008}]%
        {manning2008introduction}
\bibfield{author}{\bibinfo{person}{Christopher~D. Manning},
  \bibinfo{person}{Prabhakar Raghavan}, {and} \bibinfo{person}{Hinrich
  Sch{\"{u}}tze}.} \bibinfo{year}{2008}\natexlab{}.
\newblock \bibinfo{booktitle}{\emph{Introduction to information retrieval}}.
\newblock \bibinfo{publisher}{Cambridge University Press}.
\newblock
\showISBNx{978-0-521-86571-5}


\bibitem[\protect\citeauthoryear{Mehri, Razumovskaia, Zhao, and Eskenazi}{Mehri
  et~al\mbox{.}}{2019}]%
        {mehri2019pretraining}
\bibfield{author}{\bibinfo{person}{Shikib Mehri}, \bibinfo{person}{Evgeniia
  Razumovskaia}, \bibinfo{person}{Tiancheng Zhao}, {and}
  \bibinfo{person}{Maxine Eskenazi}.} \bibinfo{year}{2019}\natexlab{}.
\newblock \showarticletitle{Pretraining Methods for Dialog Context
  Representation Learning}. In \bibinfo{booktitle}{\emph{ACL}}.
  \bibinfo{pages}{3836--3845}.
\newblock
\urldef\tempurl%
\url{https://www.aclweb.org/anthology/P19-1373}
\showURL{%
\tempurl}


\bibitem[\protect\citeauthoryear{Mo, Zhang, Li, Li, and Yang}{Mo
  et~al\mbox{.}}{2018}]%
        {mo2016personalizing}
\bibfield{author}{\bibinfo{person}{Kaixiang Mo}, \bibinfo{person}{Yu Zhang},
  \bibinfo{person}{Shuangyin Li}, \bibinfo{person}{Jiajun Li}, {and}
  \bibinfo{person}{Qiang Yang}.} \bibinfo{year}{2018}\natexlab{}.
\newblock \showarticletitle{Personalizing a Dialogue System With Transfer
  Reinforcement Learning}. In \bibinfo{booktitle}{\emph{Proceedings of {AAAI}
  2018, New Orleans, Louisiana, USA, February 2-7, 2018}}.
  \bibinfo{pages}{5317--5324}.
\newblock


\bibitem[\protect\citeauthoryear{Moghe, Arora, Banerjee, and Khapra}{Moghe
  et~al\mbox{.}}{2018}]%
        {moghe2018towards}
\bibfield{author}{\bibinfo{person}{Nikita Moghe}, \bibinfo{person}{Siddhartha
  Arora}, \bibinfo{person}{Suman Banerjee}, {and} \bibinfo{person}{Mitesh~M
  Khapra}.} \bibinfo{year}{2018}\natexlab{}.
\newblock \showarticletitle{Towards Exploiting Background Knowledge for
  Building Conversation Systems}. In \bibinfo{booktitle}{\emph{Proceedings of
  the 2018 Conference on Empirical Methods in Natural Language Processing}}.
  \bibinfo{pages}{2322--2332}.
\newblock


\bibitem[\protect\citeauthoryear{Moon, Shah, Kumar, and Subba}{Moon
  et~al\mbox{.}}{2019}]%
        {moon2019opendialkg}
\bibfield{author}{\bibinfo{person}{Seungwhan Moon}, \bibinfo{person}{Pararth
  Shah}, \bibinfo{person}{Anuj Kumar}, {and} \bibinfo{person}{Rajen Subba}.}
  \bibinfo{year}{2019}\natexlab{}.
\newblock \showarticletitle{OpenDialKG: Explainable Conversational Reasoning
  with Attention-based Walks over Knowledge Graphs}. In
  \bibinfo{booktitle}{\emph{Proceedings of {ACL} 2019, Florence, Italy, July 28
  - Aug 2, 2019}}. \bibinfo{pages}{845--854}.
\newblock


\bibitem[\protect\citeauthoryear{Mostafazadeh, Brockett, Dolan, Galley, Gao,
  Spithourakis, and Vanderwende}{Mostafazadeh et~al\mbox{.}}{2017}]%
        {mostafazadeh2017image}
\bibfield{author}{\bibinfo{person}{Nasrin Mostafazadeh}, \bibinfo{person}{Chris
  Brockett}, \bibinfo{person}{Bill Dolan}, \bibinfo{person}{Michel Galley},
  \bibinfo{person}{Jianfeng Gao}, \bibinfo{person}{Georgios Spithourakis},
  {and} \bibinfo{person}{Lucy Vanderwende}.} \bibinfo{year}{2017}\natexlab{}.
\newblock \showarticletitle{Image-Grounded Conversations: Multimodal Context
  for Natural Question and Response Generation}. In
  \bibinfo{booktitle}{\emph{IJCNLP}}. \bibinfo{pages}{462--472}.
\newblock


\bibitem[\protect\citeauthoryear{Mrksic, S{\'{e}}aghdha, Wen, Thomson, and
  Young}{Mrksic et~al\mbox{.}}{2017}]%
        {Mrk2017ACL}
\bibfield{author}{\bibinfo{person}{Nikola Mrksic},
  \bibinfo{person}{Diarmuid~{\'{O}} S{\'{e}}aghdha},
  \bibinfo{person}{Tsung{-}Hsien Wen}, \bibinfo{person}{Blaise Thomson}, {and}
  \bibinfo{person}{Steve~J. Young}.} \bibinfo{year}{2017}\natexlab{}.
\newblock \showarticletitle{Neural Belief Tracker: Data-Driven Dialogue State
  Tracking}. In \bibinfo{booktitle}{\emph{Proceedings of {ACL} 2017, Vancouver,
  Canada, July 30-August 4, 2017}}. \bibinfo{pages}{1777--1788}.
\newblock


\bibitem[\protect\citeauthoryear{Norman}{Norman}{1963}]%
        {norman1963toward}
\bibfield{author}{\bibinfo{person}{Warren~T Norman}.}
  \bibinfo{year}{1963}\natexlab{}.
\newblock \showarticletitle{Toward an adequate taxonomy of personality
  attributes: Replicated factor structure in peer nomination personality
  ratings.}
\newblock \bibinfo{journal}{\emph{The Journal of Abnormal and Social
  Psychology}} \bibinfo{volume}{66}, \bibinfo{number}{6}
  (\bibinfo{year}{1963}), \bibinfo{pages}{574}.
\newblock


\bibitem[\protect\citeauthoryear{Nothdurft, Ultes, and Minker}{Nothdurft
  et~al\mbox{.}}{2015}]%
        {openquest15}
\bibfield{author}{\bibinfo{person}{Florian Nothdurft}, \bibinfo{person}{Stefan
  Ultes}, {and} \bibinfo{person}{Wolfgang Minker}.}
  \bibinfo{year}{2015}\natexlab{}.
\newblock \showarticletitle{Finding appropriate interaction strategies for
  proactive dialogue systems-an open quest}. In
  \bibinfo{booktitle}{\emph{Proceedings of the 2nd European and the 5th Nordic
  Symposium on Multimodal Communication, August 6-8, 2014, Tartu, Estonia}}.
  \bibinfo{pages}{73--80}.
\newblock


\bibitem[\protect\citeauthoryear{Oraby, Reed, Tandon, S., Lukin, and
  Walker}{Oraby et~al\mbox{.}}{2018}]%
        {oraby2018controlling}
\bibfield{author}{\bibinfo{person}{Shereen Oraby}, \bibinfo{person}{Lena Reed},
  \bibinfo{person}{Shubhangi Tandon}, \bibinfo{person}{Sharath~T. S.},
  \bibinfo{person}{Stephanie~M. Lukin}, {and} \bibinfo{person}{Marilyn~A.
  Walker}.} \bibinfo{year}{2018}\natexlab{}.
\newblock \showarticletitle{Controlling Personality-Based Stylistic Variation
  with Neural Natural Language Generators}. In
  \bibinfo{booktitle}{\emph{Proceedings of {SIGDIAL} 2018, July 12-14, 2018,
  Melbourne, Australia}}. \bibinfo{pages}{180--190}.
\newblock


\bibitem[\protect\citeauthoryear{Ouchi and Tsuboi}{Ouchi and Tsuboi}{2016}]%
        {ouchi2016addressee}
\bibfield{author}{\bibinfo{person}{Hiroki Ouchi} {and} \bibinfo{person}{Yuta
  Tsuboi}.} \bibinfo{year}{2016}\natexlab{}.
\newblock \showarticletitle{Addressee and Response Selection for Multi-Party
  Conversation}. In \bibinfo{booktitle}{\emph{Proceedings of {EMNLP} 2016,
  Austin, Texas, USA, November 1-4, 2016}}. \bibinfo{pages}{2133--2143}.
\newblock


\bibitem[\protect\citeauthoryear{Pado, Cer, Galley, Jurafsky, and Manning}{Pado
  et~al\mbox{.}}{2009}]%
        {pado09measuring}
\bibfield{author}{\bibinfo{person}{Sebastian Pado}, \bibinfo{person}{Daniel
  Cer}, \bibinfo{person}{Michel Galley}, \bibinfo{person}{Dan Jurafsky}, {and}
  \bibinfo{person}{Christopher~D. Manning}.} \bibinfo{year}{2009}\natexlab{}.
\newblock \showarticletitle{Measuring Machine Translation Quality as Semantic
  Equivalence: A Metric Based on Entailment Features}.
\newblock \bibinfo{journal}{\emph{Machine Translation}} (\bibinfo{year}{2009}),
  \bibinfo{pages}{181--193}.
\newblock


\bibitem[\protect\citeauthoryear{Palangi, Deng, Shen, Gao, He, Chen, Song, and
  Ward}{Palangi et~al\mbox{.}}{2016}]%
        {palangi2015deep}
\bibfield{author}{\bibinfo{person}{Hamid Palangi}, \bibinfo{person}{Li Deng},
  \bibinfo{person}{Yelong Shen}, \bibinfo{person}{Jianfeng Gao},
  \bibinfo{person}{Xiaodong He}, \bibinfo{person}{Jianshu Chen},
  \bibinfo{person}{Xinying Song}, {and} \bibinfo{person}{Rabab~K. Ward}.}
  \bibinfo{year}{2016}\natexlab{}.
\newblock \showarticletitle{Deep Sentence Embedding Using Long Short-Term
  Memory Networks: Analysis and Application to Information Retrieval}.
\newblock \bibinfo{journal}{\emph{{IEEE/ACM} Trans. Audio, Speech {\&} Language
  Processing}} \bibinfo{volume}{24}, \bibinfo{number}{4}
  (\bibinfo{year}{2016}), \bibinfo{pages}{694--707}.
\newblock


\bibitem[\protect\citeauthoryear{Pandey, Contractor, Kumar, and Joshi}{Pandey
  et~al\mbox{.}}{2018}]%
        {Contractor:2018vl}
\bibfield{author}{\bibinfo{person}{Gaurav Pandey}, \bibinfo{person}{Danish
  Contractor}, \bibinfo{person}{Vineet Kumar}, {and} \bibinfo{person}{Sachindra
  Joshi}.} \bibinfo{year}{2018}\natexlab{}.
\newblock \showarticletitle{Exemplar Encoder-Decoder for Neural Conversation
  Generation}. In \bibinfo{booktitle}{\emph{Proceedings of {ACL} 2018,
  Melbourne, Australia, July 15-20, 2018}}. \bibinfo{pages}{1329--1338}.
\newblock


\bibitem[\protect\citeauthoryear{Pang, Lan, Guo, Xu, Wan, and Cheng}{Pang
  et~al\mbox{.}}{2016}]%
        {pang2016pyramid}
\bibfield{author}{\bibinfo{person}{Liang Pang}, \bibinfo{person}{Yanyan Lan},
  \bibinfo{person}{Jiafeng Guo}, \bibinfo{person}{Jun Xu},
  \bibinfo{person}{Shengxian Wan}, {and} \bibinfo{person}{Xueqi Cheng}.}
  \bibinfo{year}{2016}\natexlab{}.
\newblock \showarticletitle{Text Matching as Image Recognition}. In
  \bibinfo{booktitle}{\emph{Proceedings of {AAAI} 2016, February 12-17, 2016,
  Phoenix, Arizona, {USA.}}} \bibinfo{pages}{2793--2799}.
\newblock


\bibitem[\protect\citeauthoryear{Papineni, Roukos, Ward, and Zhu}{Papineni
  et~al\mbox{.}}{2002}]%
        {papineni2002bleu}
\bibfield{author}{\bibinfo{person}{Kishore Papineni}, \bibinfo{person}{Salim
  Roukos}, \bibinfo{person}{Todd Ward}, {and} \bibinfo{person}{Wei{-}Jing
  Zhu}.} \bibinfo{year}{2002}\natexlab{}.
\newblock \showarticletitle{Bleu: a Method for Automatic Evaluation of Machine
  Translation}. In \bibinfo{booktitle}{\emph{Proceedings of {ACL} 2002, July
  6-12, 2002, Philadelphia, PA, {USA.}}} \bibinfo{pages}{311--318}.
\newblock


\bibitem[\protect\citeauthoryear{Peng, Li, Li, Gao, {\c{C}}elikyilmaz, Lee, and
  Wong}{Peng et~al\mbox{.}}{2017}]%
        {Peng2017EMNLP}
\bibfield{author}{\bibinfo{person}{Baolin Peng}, \bibinfo{person}{Xiujun Li},
  \bibinfo{person}{Lihong Li}, \bibinfo{person}{Jianfeng Gao},
  \bibinfo{person}{Asli {\c{C}}elikyilmaz}, \bibinfo{person}{Sungjin Lee},
  {and} \bibinfo{person}{Kam{-}Fai Wong}.} \bibinfo{year}{2017}\natexlab{}.
\newblock \showarticletitle{Composite Task-Completion Dialogue Policy Learning
  via Hierarchical Deep Reinforcement Learning}. In
  \bibinfo{booktitle}{\emph{Proceedings of {EMNLP} 2017, Copenhagen, Denmark,
  September 9-11, 2017}}. \bibinfo{pages}{2231--2240}.
\newblock


\bibitem[\protect\citeauthoryear{Pennebaker, Francis, and Booth}{Pennebaker
  et~al\mbox{.}}{2001}]%
        {pennebaker2001linguistic}
\bibfield{author}{\bibinfo{person}{James~W Pennebaker},
  \bibinfo{person}{Martha~E Francis}, {and} \bibinfo{person}{Roger~J Booth}.}
  \bibinfo{year}{2001}\natexlab{}.
\newblock \showarticletitle{Linguistic inquiry and word count: LIWC 2001}.
\newblock \bibinfo{journal}{\emph{Mahway: Lawrence Erlbaum Associates}}
  \bibinfo{volume}{71}, \bibinfo{number}{2001} (\bibinfo{year}{2001}),
  \bibinfo{pages}{2001}.
\newblock


\bibitem[\protect\citeauthoryear{P{\'e}rez-Rosas, Mihalcea, Resnicow, Singh,
  and An}{P{\'e}rez-Rosas et~al\mbox{.}}{2017}]%
        {perez2017understanding}
\bibfield{author}{\bibinfo{person}{Ver{\'o}nica P{\'e}rez-Rosas},
  \bibinfo{person}{Rada Mihalcea}, \bibinfo{person}{Kenneth Resnicow},
  \bibinfo{person}{Satinder Singh}, {and} \bibinfo{person}{Lawrence An}.}
  \bibinfo{year}{2017}\natexlab{}.
\newblock \showarticletitle{Understanding and predicting empathic behavior in
  counseling therapy}. In \bibinfo{booktitle}{\emph{Proceedings of the 55th
  Annual Meeting of the Association for Computational Linguistics (Volume 1:
  Long Papers)}}. \bibinfo{pages}{1426--1435}.
\newblock


\bibitem[\protect\citeauthoryear{Pichl, Marek, Konr{\'{a}}d, Matul{\'{\i}}k,
  Nguyen, and Sediv{\'{y}}}{Pichl et~al\mbox{.}}{2018}]%
        {pichl2018alquist}
\bibfield{author}{\bibinfo{person}{Jan Pichl}, \bibinfo{person}{Petr Marek},
  \bibinfo{person}{Jakub Konr{\'{a}}d}, \bibinfo{person}{Martin
  Matul{\'{\i}}k}, \bibinfo{person}{Hoang~Long Nguyen}, {and}
  \bibinfo{person}{Jan Sediv{\'{y}}}.} \bibinfo{year}{2018}\natexlab{}.
\newblock \showarticletitle{Alquist: The Alexa Prize Socialbot}.
\newblock \bibinfo{journal}{\emph{CoRR}}  \bibinfo{volume}{abs/1804.06705}
  (\bibinfo{year}{2018}).
\newblock


\bibitem[\protect\citeauthoryear{Qian, Huang, Zhao, Xu, and Zhu}{Qian
  et~al\mbox{.}}{2018}]%
        {Qian2017AssigningProfile}
\bibfield{author}{\bibinfo{person}{Qiao Qian}, \bibinfo{person}{Minlie Huang},
  \bibinfo{person}{Haizhou Zhao}, \bibinfo{person}{Jingfang Xu}, {and}
  \bibinfo{person}{Xiaoyan Zhu}.} \bibinfo{year}{2018}\natexlab{}.
\newblock \showarticletitle{Assigning Personality/Profile to a Chatting Machine
  for Coherent Conversation Generation}. In
  \bibinfo{booktitle}{\emph{Proceedings of {IJCAI} 2018, Stockholm, Sweden,
  July 13-19, 2018}}. \bibinfo{pages}{4279--4285}.
\newblock


\bibitem[\protect\citeauthoryear{Qin, Galley, Brockett, Liu, Gao, Dolan, Choi,
  and Gao}{Qin et~al\mbox{.}}{2019}]%
        {qin2019conversing}
\bibfield{author}{\bibinfo{person}{Lianhui Qin}, \bibinfo{person}{Michel
  Galley}, \bibinfo{person}{Chris Brockett}, \bibinfo{person}{Xiaodong Liu},
  \bibinfo{person}{Xiang Gao}, \bibinfo{person}{Bill Dolan},
  \bibinfo{person}{Yejin Choi}, {and} \bibinfo{person}{Jianfeng Gao}.}
  \bibinfo{year}{2019}\natexlab{}.
\newblock \showarticletitle{Conversing by Reading: Contentful Neural
  Conversation with On-demand Machine Reading}.
\newblock \bibinfo{journal}{\emph{ACL}} (\bibinfo{year}{2019}).
\newblock


\bibitem[\protect\citeauthoryear{Radford, J{\'{o}}zefowicz, and
  Sutskever}{Radford et~al\mbox{.}}{2017}]%
        {radford2017learning}
\bibfield{author}{\bibinfo{person}{Alec Radford}, \bibinfo{person}{Rafal
  J{\'{o}}zefowicz}, {and} \bibinfo{person}{Ilya Sutskever}.}
  \bibinfo{year}{2017}\natexlab{}.
\newblock \showarticletitle{Learning to Generate Reviews and Discovering
  Sentiment}.
\newblock \bibinfo{journal}{\emph{CoRR}}  \bibinfo{volume}{abs/1704.01444}
  (\bibinfo{year}{2017}).
\newblock


\bibitem[\protect\citeauthoryear{Radford, Narasimhan, Salimans, and
  Sutskever}{Radford et~al\mbox{.}}{2018}]%
        {radford2018improving}
\bibfield{author}{\bibinfo{person}{Alec Radford}, \bibinfo{person}{Karthik
  Narasimhan}, \bibinfo{person}{Tim Salimans}, {and} \bibinfo{person}{Ilya
  Sutskever}.} \bibinfo{year}{2018}\natexlab{}.
\newblock \showarticletitle{Improving language understanding by generative
  pre-training}.
\newblock \bibinfo{journal}{\emph{OpenAI Blog}} (\bibinfo{year}{2018}).
\newblock


\bibitem[\protect\citeauthoryear{Ram, Prasad, Khatri, Venkatesh, Gabriel, Liu,
  Nunn, Hedayatnia, Cheng, Nagar, King, Bland, Wartick, Pan, Song, Jayadevan,
  Hwang, and Pettigrue}{Ram et~al\mbox{.}}{2018}]%
        {ram2018conversational}
\bibfield{author}{\bibinfo{person}{Ashwin Ram}, \bibinfo{person}{Rohit Prasad},
  \bibinfo{person}{Chandra Khatri}, \bibinfo{person}{Anu Venkatesh},
  \bibinfo{person}{Raefer Gabriel}, \bibinfo{person}{Qing Liu},
  \bibinfo{person}{Jeff Nunn}, \bibinfo{person}{Behnam Hedayatnia},
  \bibinfo{person}{Ming Cheng}, \bibinfo{person}{Ashish Nagar},
  \bibinfo{person}{Eric King}, \bibinfo{person}{Kate Bland},
  \bibinfo{person}{Amanda Wartick}, \bibinfo{person}{Yi Pan},
  \bibinfo{person}{Han Song}, \bibinfo{person}{Sk Jayadevan},
  \bibinfo{person}{Gene Hwang}, {and} \bibinfo{person}{Art Pettigrue}.}
  \bibinfo{year}{2018}\natexlab{}.
\newblock \showarticletitle{Conversational {AI:} The Science Behind the Alexa
  Prize}.
\newblock \bibinfo{journal}{\emph{CoRR}}  \bibinfo{volume}{abs/1801.03604}
  (\bibinfo{year}{2018}).
\newblock


\bibitem[\protect\citeauthoryear{Rao and Daum{\'e}~III}{Rao and
  Daum{\'e}~III}{2018}]%
        {rao2018learning}
\bibfield{author}{\bibinfo{person}{Sudha Rao} {and} \bibinfo{person}{Hal
  Daum{\'e}~III}.} \bibinfo{year}{2018}\natexlab{}.
\newblock \showarticletitle{Learning to Ask Good Questions: Ranking
  Clarification Questions using Neural Expected Value of Perfect Information}.
  In \bibinfo{booktitle}{\emph{Proceedings of the 56th Annual Meeting of the
  Association for Computational Linguistics (Volume 1: Long Papers)}}.
  \bibinfo{pages}{2737--2746}.
\newblock


\bibitem[\protect\citeauthoryear{Rashkin, Smith, Li, and Boureau}{Rashkin
  et~al\mbox{.}}{2019}]%
        {rashkin2019empa}
\bibfield{author}{\bibinfo{person}{Hannah Rashkin},
  \bibinfo{person}{Eric~Michael Smith}, \bibinfo{person}{Margaret Li}, {and}
  \bibinfo{person}{Y-Lan Boureau}.} \bibinfo{year}{2019}\natexlab{}.
\newblock \showarticletitle{Towards Empathetic Open-domain Conversation Models:
  A New Benchmark and Dataset}. In \bibinfo{booktitle}{\emph{Proceedings of
  {ACL} 2019, Florence, Italy, July 28 - Aug 2, 2019}}.
  \bibinfo{pages}{5370--5381}.
\newblock


\bibitem[\protect\citeauthoryear{Ren, Chen, Monz, Ma, and de~Rijke}{Ren
  et~al\mbox{.}}{2019}]%
        {ren2019thinking}
\bibfield{author}{\bibinfo{person}{Pengjie Ren}, \bibinfo{person}{Zhumin Chen},
  \bibinfo{person}{Christof Monz}, \bibinfo{person}{Jun Ma}, {and}
  \bibinfo{person}{Maarten de Rijke}.} \bibinfo{year}{2019}\natexlab{}.
\newblock \showarticletitle{Thinking Globally, Acting Locally: Distantly
  Supervised Global-to-Local Knowledge Selection for Background Based
  Conversation}.
\newblock \bibinfo{journal}{\emph{arXiv preprint arXiv:1908.09528}}
  (\bibinfo{year}{2019}).
\newblock


\bibitem[\protect\citeauthoryear{Ritter, Cherry, and Dolan}{Ritter
  et~al\mbox{.}}{2011}]%
        {a30}
\bibfield{author}{\bibinfo{person}{Alan Ritter}, \bibinfo{person}{Colin
  Cherry}, {and} \bibinfo{person}{William~B. Dolan}.}
  \bibinfo{year}{2011}\natexlab{}.
\newblock \showarticletitle{Data-Driven Response Generation in Social Media}.
  In \bibinfo{booktitle}{\emph{Proceedings of {EMNLP} 2011, John McIntyre
  Conference Centre, Edinburgh, UK, 27-31 July 2011}}.
  \bibinfo{pages}{583--593}.
\newblock


\bibitem[\protect\citeauthoryear{Rojas{-}Barahona, Gasic, Mrksic, Su, Ultes,
  Wen, Young, and Vandyke}{Rojas{-}Barahona et~al\mbox{.}}{2017}]%
        {wen2017network}
\bibfield{author}{\bibinfo{person}{Lina~Maria Rojas{-}Barahona},
  \bibinfo{person}{Milica Gasic}, \bibinfo{person}{Nikola Mrksic},
  \bibinfo{person}{Pei{-}Hao Su}, \bibinfo{person}{Stefan Ultes},
  \bibinfo{person}{Tsung{-}Hsien Wen}, \bibinfo{person}{Steve~J. Young}, {and}
  \bibinfo{person}{David Vandyke}.} \bibinfo{year}{2017}\natexlab{}.
\newblock \showarticletitle{A Network-based End-to-End Trainable Task-oriented
  Dialogue System}. In \bibinfo{booktitle}{\emph{Proceedings of the 15th
  Conference of the European Chapter of the Association for Computational
  Linguistics, {EACL} 2017, Valencia, Spain, April 3-7, 2017}}.
  \bibinfo{pages}{438--449}.
\newblock


\bibitem[\protect\citeauthoryear{Saha, Khapra, and Sankaranarayanan}{Saha
  et~al\mbox{.}}{2017}]%
        {saha2017multimodal}
\bibfield{author}{\bibinfo{person}{Amrita Saha}, \bibinfo{person}{Mitesh
  Khapra}, {and} \bibinfo{person}{Karthik Sankaranarayanan}.}
  \bibinfo{year}{2017}\natexlab{}.
\newblock \showarticletitle{Multimodal Dialogs (MMD): A large-scale dataset for
  studying multimodal domain-aware conversations}.
\newblock \bibinfo{journal}{\emph{arXiv preprint arXiv:1704.00200}}
  (\bibinfo{year}{2017}).
\newblock


\bibitem[\protect\citeauthoryear{Sai, Gupta, Khapra, and Srinivasan}{Sai
  et~al\mbox{.}}{2019}]%
        {sai2019re}
\bibfield{author}{\bibinfo{person}{Ananya Sai}, \bibinfo{person}{Mithun~Das
  Gupta}, \bibinfo{person}{Mitesh~M. Khapra}, {and} \bibinfo{person}{Mukundhan
  Srinivasan}.} \bibinfo{year}{2019}\natexlab{}.
\newblock \showarticletitle{Response Generation by Context-aware Prototype
  Editing}. In \bibinfo{booktitle}{\emph{Proceedings of {AAAI} 2019, Honolulu,
  Hawaii, USA, Jan 27-Feb 1, 2019}}.
\newblock


\bibitem[\protect\citeauthoryear{See, Liu, and Manning}{See
  et~al\mbox{.}}{2017}]%
        {see2017get}
\bibfield{author}{\bibinfo{person}{Abigail See}, \bibinfo{person}{Peter~J Liu},
  {and} \bibinfo{person}{Christopher~D Manning}.}
  \bibinfo{year}{2017}\natexlab{}.
\newblock \showarticletitle{Get To The Point: Summarization with
  Pointer-Generator Networks}. In \bibinfo{booktitle}{\emph{Proceedings of the
  55th Annual Meeting of the Association for Computational Linguistics (Volume
  1: Long Papers)}}. \bibinfo{pages}{1073--1083}.
\newblock


\bibitem[\protect\citeauthoryear{See, Roller, Kiela, and Weston}{See
  et~al\mbox{.}}{2019}]%
        {see2019makes}
\bibfield{author}{\bibinfo{person}{Abigail See}, \bibinfo{person}{Stephen
  Roller}, \bibinfo{person}{Douwe Kiela}, {and} \bibinfo{person}{Jason
  Weston}.} \bibinfo{year}{2019}\natexlab{}.
\newblock \showarticletitle{What makes a good conversation? How controllable
  attributes affect human judgments}. In \bibinfo{booktitle}{\emph{Proceedings
  of the 2019 Conference of the North American Chapter of the Association for
  Computational Linguistics: Human Language Technologies, Volume 1 (Long and
  Short Papers)}}. \bibinfo{pages}{1702--1723}.
\newblock


\bibitem[\protect\citeauthoryear{Seo, Kembhavi, Farhadi, and Hajishirzi}{Seo
  et~al\mbox{.}}{2016}]%
        {seo2016bidirectional}
\bibfield{author}{\bibinfo{person}{Minjoon Seo}, \bibinfo{person}{Aniruddha
  Kembhavi}, \bibinfo{person}{Ali Farhadi}, {and} \bibinfo{person}{Hannaneh
  Hajishirzi}.} \bibinfo{year}{2016}\natexlab{}.
\newblock \showarticletitle{Bidirectional attention flow for machine
  comprehension}.
\newblock \bibinfo{journal}{\emph{arXiv preprint arXiv:1611.01603}}
  (\bibinfo{year}{2016}).
\newblock


\bibitem[\protect\citeauthoryear{Serban, Lowe, Henderson, Charlin, and
  Pineau}{Serban et~al\mbox{.}}{2015a}]%
        {serban2015survey}
\bibfield{author}{\bibinfo{person}{Iulian~Vlad Serban}, \bibinfo{person}{Ryan
  Lowe}, \bibinfo{person}{Peter Henderson}, \bibinfo{person}{Laurent Charlin},
  {and} \bibinfo{person}{Joelle Pineau}.} \bibinfo{year}{2015}\natexlab{a}.
\newblock \showarticletitle{A survey of available corpora for building
  data-driven dialogue systems}.
\newblock \bibinfo{journal}{\emph{arXiv preprint arXiv:1512.05742}}
  (\bibinfo{year}{2015}).
\newblock


\bibitem[\protect\citeauthoryear{Serban, Sordoni, Bengio, Courville, and
  Pineau}{Serban et~al\mbox{.}}{2015b}]%
        {serban2015hierarchical}
\bibfield{author}{\bibinfo{person}{Iulian~Vlad Serban},
  \bibinfo{person}{Alessandro Sordoni}, \bibinfo{person}{Yoshua Bengio},
  \bibinfo{person}{Aaron Courville}, {and} \bibinfo{person}{Joelle Pineau}.}
  \bibinfo{year}{2015}\natexlab{b}.
\newblock \showarticletitle{Hierarchical Neural Network Generative Models for
  Movie Dialogues}.
\newblock \bibinfo{journal}{\emph{arXiv preprint arXiv:1507.04808}}
  (\bibinfo{year}{2015}).
\newblock


\bibitem[\protect\citeauthoryear{Serban, Sordoni, Bengio, Courville, and
  Pineau}{Serban et~al\mbox{.}}{2016}]%
        {hiContext}
\bibfield{author}{\bibinfo{person}{Iulian~Vlad Serban},
  \bibinfo{person}{Alessandro Sordoni}, \bibinfo{person}{Yoshua Bengio},
  \bibinfo{person}{Aaron~C. Courville}, {and} \bibinfo{person}{Joelle Pineau}.}
  \bibinfo{year}{2016}\natexlab{}.
\newblock \showarticletitle{Building End-To-End Dialogue Systems Using
  Generative Hierarchical Neural Network Models}. In
  \bibinfo{booktitle}{\emph{Proceedings of {AAAI} 2016, February 12-17, 2016,
  Phoenix, Arizona, {USA.}}} \bibinfo{pages}{3776--3784}.
\newblock


\bibitem[\protect\citeauthoryear{Serban, Sordoni, Lowe, Charlin, Pineau,
  Courville, and Bengio}{Serban et~al\mbox{.}}{2017}]%
        {serban2017hierarchical-1}
\bibfield{author}{\bibinfo{person}{Iulian~Vlad Serban},
  \bibinfo{person}{Alessandro Sordoni}, \bibinfo{person}{Ryan Lowe},
  \bibinfo{person}{Laurent Charlin}, \bibinfo{person}{Joelle Pineau},
  \bibinfo{person}{Aaron~C. Courville}, {and} \bibinfo{person}{Yoshua Bengio}.}
  \bibinfo{year}{2017}\natexlab{}.
\newblock \showarticletitle{A Hierarchical Latent Variable Encoder-Decoder
  Model for Generating Dialogues}. In \bibinfo{booktitle}{\emph{AAAI}}.
  \bibinfo{pages}{3295--3301}.
\newblock


\bibitem[\protect\citeauthoryear{Severyn and Moschitti}{Severyn and
  Moschitti}{2015}]%
        {Severyn:2015gm}
\bibfield{author}{\bibinfo{person}{Aliaksei Severyn} {and}
  \bibinfo{person}{Alessandro Moschitti}.} \bibinfo{year}{2015}\natexlab{}.
\newblock \showarticletitle{Learning to Rank Short Text Pairs with
  Convolutional Deep Neural Networks}. In \bibinfo{booktitle}{\emph{Proceedings
  of {SIGIR} 2015, Santiago, Chile, August 9-13, 2015}}.
  \bibinfo{pages}{373--382}.
\newblock


\bibitem[\protect\citeauthoryear{Shang, Lu, and Li}{Shang
  et~al\mbox{.}}{2015}]%
        {shang-2015-NRM}
\bibfield{author}{\bibinfo{person}{Lifeng Shang}, \bibinfo{person}{Zhengdong
  Lu}, {and} \bibinfo{person}{Hang Li}.} \bibinfo{year}{2015}\natexlab{}.
\newblock \showarticletitle{Neural Responding Machine for Short-Text
  Conversation}. In \bibinfo{booktitle}{\emph{Proceedings of {ACL} 2015, July
  26-31, 2015, Beijing, China}}. \bibinfo{pages}{1577--1586}.
\newblock


\bibitem[\protect\citeauthoryear{Shen, Su, Niu, and Demberg}{Shen
  et~al\mbox{.}}{2018}]%
        {shen18cvae}
\bibfield{author}{\bibinfo{person}{Xiaoyu Shen}, \bibinfo{person}{Hui Su},
  \bibinfo{person}{Shuzi Niu}, {and} \bibinfo{person}{Vera Demberg}.}
  \bibinfo{year}{2018}\natexlab{}.
\newblock \showarticletitle{Improving Variational Encoder-Decoders in Dialogue
  Generation}. In \bibinfo{booktitle}{\emph{Proceedings of {AAAI} 2018, New
  Orleans, Louisiana, USA, February 2-7, 2018}}. \bibinfo{pages}{5456--5463}.
\newblock


\bibitem[\protect\citeauthoryear{Shen, He, Gao, Deng, and Mesnil}{Shen
  et~al\mbox{.}}{2014}]%
        {Shen:2014id}
\bibfield{author}{\bibinfo{person}{Yelong Shen}, \bibinfo{person}{Xiaodong He},
  \bibinfo{person}{Jianfeng Gao}, \bibinfo{person}{Li Deng}, {and}
  \bibinfo{person}{Gr{\'{e}}goire Mesnil}.} \bibinfo{year}{2014}\natexlab{}.
\newblock \showarticletitle{A Latent Semantic Model with Convolutional-Pooling
  Structure for Information Retrieval}. In
  \bibinfo{booktitle}{\emph{Proceedings of the 23rd {ACM} International
  Conference on Conference on Information and Knowledge Management, {CIKM}
  2014, Shanghai, China, November 3-7, 2014}}. \bibinfo{pages}{101--110}.
\newblock


\bibitem[\protect\citeauthoryear{Shum, He, and Li}{Shum et~al\mbox{.}}{2018}]%
        {shum2018eliza}
\bibfield{author}{\bibinfo{person}{Heung{-}Yeung Shum},
  \bibinfo{person}{Xiaodong He}, {and} \bibinfo{person}{Di Li}.}
  \bibinfo{year}{2018}\natexlab{}.
\newblock \showarticletitle{From Eliza to XiaoIce: challenges and opportunities
  with social chatbots}.
\newblock \bibinfo{journal}{\emph{Frontiers of {IT} {\&} {EE}}}
  \bibinfo{volume}{19}, \bibinfo{number}{1} (\bibinfo{year}{2018}),
  \bibinfo{pages}{10--26}.
\newblock


\bibitem[\protect\citeauthoryear{Sohn, Lee, and Yan}{Sohn
  et~al\mbox{.}}{2015}]%
        {sohn2015learning}
\bibfield{author}{\bibinfo{person}{Kihyuk Sohn}, \bibinfo{person}{Honglak Lee},
  {and} \bibinfo{person}{Xinchen Yan}.} \bibinfo{year}{2015}\natexlab{}.
\newblock \showarticletitle{Learning Structured Output Representation using
  Deep Conditional Generative Models}. In \bibinfo{booktitle}{\emph{{NIPS}
  2015, December 7-12, 2015, Montreal, Quebec, Canada}}.
  \bibinfo{pages}{3483--3491}.
\newblock


\bibitem[\protect\citeauthoryear{Song, Li, Nie, Zhang, Zhao, and Yan}{Song
  et~al\mbox{.}}{2018}]%
        {Song:2018ve}
\bibfield{author}{\bibinfo{person}{Yiping Song}, \bibinfo{person}{Cheng{-}Te
  Li}, \bibinfo{person}{Jian{-}Yun Nie}, \bibinfo{person}{Ming Zhang},
  \bibinfo{person}{Dongyan Zhao}, {and} \bibinfo{person}{Rui Yan}.}
  \bibinfo{year}{2018}\natexlab{}.
\newblock \showarticletitle{An Ensemble of Retrieval-Based and Generation-Based
  Human-Computer Conversation Systems}. In
  \bibinfo{booktitle}{\emph{Proceedings of {IJCAI} 2018, Stockholm, Sweden,
  July 13-19, 2018}}. \bibinfo{pages}{4382--4388}.
\newblock


\bibitem[\protect\citeauthoryear{Sordoni, Galley, Auli, Brockett, Ji, Mitchell,
  Nie, Gao, and Dolan}{Sordoni et~al\mbox{.}}{2015}]%
        {sordoni2015neural}
\bibfield{author}{\bibinfo{person}{Alessandro Sordoni}, \bibinfo{person}{Michel
  Galley}, \bibinfo{person}{Michael Auli}, \bibinfo{person}{Chris Brockett},
  \bibinfo{person}{Yangfeng Ji}, \bibinfo{person}{Margaret Mitchell},
  \bibinfo{person}{Jian-Yun Nie}, \bibinfo{person}{Jianfeng Gao}, {and}
  \bibinfo{person}{Bill Dolan}.} \bibinfo{year}{2015}\natexlab{}.
\newblock \showarticletitle{A Neural Network Approach to Context-Sensitive
  Generation of Conversational Responses}. In \bibinfo{booktitle}{\emph{{NAACL}
  {HLT} 2015, Denver, Colorado, USA, May 31 - June 5, 2015}}.
  \bibinfo{pages}{196--205}.
\newblock


\bibitem[\protect\citeauthoryear{Speer, Chin, and Havasi}{Speer
  et~al\mbox{.}}{2017}]%
        {speer2013conceptnet}
\bibfield{author}{\bibinfo{person}{Robyn Speer}, \bibinfo{person}{Joshua Chin},
  {and} \bibinfo{person}{Catherine Havasi}.} \bibinfo{year}{2017}\natexlab{}.
\newblock \showarticletitle{ConceptNet 5.5: An Open Multilingual Graph of
  General Knowledge}. In \bibinfo{booktitle}{\emph{Proceedings of {AAAI} 2017,
  San Francisco, California, {USA.}, February 4-9, 2017}}.
  \bibinfo{pages}{4444--4451}.
\newblock


\bibitem[\protect\citeauthoryear{Stanojevi\'{c} and Sima'an}{Stanojevi\'{c} and
  Sima'an}{2014}]%
        {stanojevic14fitting}
\bibfield{author}{\bibinfo{person}{Milo\v{s} Stanojevi\'{c}} {and}
  \bibinfo{person}{Khalil Sima'an}.} \bibinfo{year}{2014}\natexlab{}.
\newblock \showarticletitle{Fitting Sentence Level Translation Evaluation with
  Many Dense Features}. In \bibinfo{booktitle}{\emph{Proceedings of {EMNLP}
  2014, Doha, Qatar, October 25-29, 2014}}. \bibinfo{pages}{202--206}.
\newblock


\bibitem[\protect\citeauthoryear{Su, Gasic, Mrksic, Rojas{-}Barahona, Ultes,
  Vandyke, Wen, and Young}{Su et~al\mbox{.}}{2016}]%
        {Su2016ACL}
\bibfield{author}{\bibinfo{person}{Pei{-}Hao Su}, \bibinfo{person}{Milica
  Gasic}, \bibinfo{person}{Nikola Mrksic}, \bibinfo{person}{Lina~Maria
  Rojas{-}Barahona}, \bibinfo{person}{Stefan Ultes}, \bibinfo{person}{David
  Vandyke}, \bibinfo{person}{Tsung{-}Hsien Wen}, {and}
  \bibinfo{person}{Steve~J. Young}.} \bibinfo{year}{2016}\natexlab{}.
\newblock \showarticletitle{On-line Active Reward Learning for Policy
  Optimisation in Spoken Dialogue Systems}. In
  \bibinfo{booktitle}{\emph{Proceedings of {ACL} 2016, Berlin, Germany, August
  7-12, 2016}}.
\newblock


\bibitem[\protect\citeauthoryear{Sutskever, Vinyals, and Le}{Sutskever
  et~al\mbox{.}}{2014}]%
        {2014sequence}
\bibfield{author}{\bibinfo{person}{Ilya Sutskever}, \bibinfo{person}{Oriol
  Vinyals}, {and} \bibinfo{person}{Quoc~V. Le}.}
  \bibinfo{year}{2014}\natexlab{}.
\newblock \showarticletitle{Sequence to Sequence Learning with Neural
  Networks}. In \bibinfo{booktitle}{\emph{{NIPS} 2014, Montreal, Quebec,
  Canada, December 8-13, 2014}}. \bibinfo{pages}{3104--3112}.
\newblock


\bibitem[\protect\citeauthoryear{Szegedy, Zaremba, Sutskever, Bruna, Erhan,
  Goodfellow, and Fergus}{Szegedy et~al\mbox{.}}{2013}]%
        {szegedy13intriging}
\bibfield{author}{\bibinfo{person}{Christian Szegedy},
  \bibinfo{person}{Wojciech Zaremba}, \bibinfo{person}{Ilya Sutskever},
  \bibinfo{person}{Joan Bruna}, \bibinfo{person}{Dumitru Erhan},
  \bibinfo{person}{Ian~J. Goodfellow}, {and} \bibinfo{person}{Rob Fergus}.}
  \bibinfo{year}{2013}\natexlab{}.
\newblock \showarticletitle{Intriguing properties of neural networks}.
\newblock \bibinfo{journal}{\emph{CoRR}}  \bibinfo{volume}{abs/1312.6199}
  (\bibinfo{year}{2013}).
\newblock


\bibitem[\protect\citeauthoryear{Tang, Zhao, Xiong, Liang, Xing, and Hu}{Tang
  et~al\mbox{.}}{2019}]%
        {tang2019target}
\bibfield{author}{\bibinfo{person}{Jianheng Tang}, \bibinfo{person}{Tiancheng
  Zhao}, \bibinfo{person}{Chenyan Xiong}, \bibinfo{person}{Xiaodan Liang},
  \bibinfo{person}{Eric~P. Xing}, {and} \bibinfo{person}{Zhiting Hu}.}
  \bibinfo{year}{2019}\natexlab{}.
\newblock \showarticletitle{Target-Guided Open-Domain Conversation}. In
  \bibinfo{booktitle}{\emph{Proceedings of {ACL} 2019, Florence, Italy, July 28
  - Aug 2, 2019}}. \bibinfo{pages}{5624--5634}.
\newblock


\bibitem[\protect\citeauthoryear{Tao, Mou, Zhao, and Yan}{Tao
  et~al\mbox{.}}{2018}]%
        {tao2018ruber}
\bibfield{author}{\bibinfo{person}{Chongyang Tao}, \bibinfo{person}{Lili Mou},
  \bibinfo{person}{Dongyan Zhao}, {and} \bibinfo{person}{Rui Yan}.}
  \bibinfo{year}{2018}\natexlab{}.
\newblock \showarticletitle{{RUBER:} An Unsupervised Method for Automatic
  Evaluation of Open-Domain Dialog Systems}. In
  \bibinfo{booktitle}{\emph{Proceedings of {AAAI} 2018, New Orleans, Louisiana,
  USA, February 2-7, 2018}}. \bibinfo{pages}{722--729}.
\newblock


\bibitem[\protect\citeauthoryear{Tao, Wu, Xu, Hu, Zhao, and Yan}{Tao
  et~al\mbox{.}}{2019}]%
        {tao2019multi}
\bibfield{author}{\bibinfo{person}{Chongyang Tao}, \bibinfo{person}{Wei Wu},
  \bibinfo{person}{Can Xu}, \bibinfo{person}{Wenpeng Hu},
  \bibinfo{person}{Dongyan Zhao}, {and} \bibinfo{person}{Rui Yan}.}
  \bibinfo{year}{2019}\natexlab{}.
\newblock \showarticletitle{Multi-Representation Fusion Network for Multi-Turn
  Response Selection in Retrieval-Based Chatbots}. In
  \bibinfo{booktitle}{\emph{Proceedings of the Twelfth ACM International
  Conference on Web Search and Data Mining}}. ACM, \bibinfo{pages}{267--275}.
\newblock


\bibitem[\protect\citeauthoryear{Turing}{Turing}{1950}]%
        {alan1950}
\bibfield{author}{\bibinfo{person}{Alan~M Turing}.}
  \bibinfo{year}{1950}\natexlab{}.
\newblock \showarticletitle{Computing machinery and intelligence}.
\newblock \bibinfo{journal}{\emph{Mind}} \bibinfo{volume}{59},
  \bibinfo{number}{236} (\bibinfo{year}{1950}), \bibinfo{pages}{433--460}.
\newblock


\bibitem[\protect\citeauthoryear{Vaswani, Shazeer, Parmar, Uszkoreit, Jones,
  Gomez, Kaiser, and Polosukhin}{Vaswani et~al\mbox{.}}{2017}]%
        {vaswani2017attention}
\bibfield{author}{\bibinfo{person}{Ashish Vaswani}, \bibinfo{person}{Noam
  Shazeer}, \bibinfo{person}{Niki Parmar}, \bibinfo{person}{Jakob Uszkoreit},
  \bibinfo{person}{Llion Jones}, \bibinfo{person}{Aidan~N Gomez},
  \bibinfo{person}{{\L}ukasz Kaiser}, {and} \bibinfo{person}{Illia
  Polosukhin}.} \bibinfo{year}{2017}\natexlab{}.
\newblock \showarticletitle{Attention is all you need}. In
  \bibinfo{booktitle}{\emph{Advances in neural information processing
  systems}}. \bibinfo{pages}{5998--6008}.
\newblock


\bibitem[\protect\citeauthoryear{Vijayakumar, Cogswell, Selvaraju, Sun, Lee,
  Crandall, and Batra}{Vijayakumar et~al\mbox{.}}{2018}]%
        {Ashwin2018DBS}
\bibfield{author}{\bibinfo{person}{Ashwin~K. Vijayakumar},
  \bibinfo{person}{Michael Cogswell}, \bibinfo{person}{Ramprasaath~R.
  Selvaraju}, \bibinfo{person}{Qing Sun}, \bibinfo{person}{Stefan Lee},
  \bibinfo{person}{David~J. Crandall}, {and} \bibinfo{person}{Dhruv Batra}.}
  \bibinfo{year}{2018}\natexlab{}.
\newblock \showarticletitle{Diverse Beam Search for Improved Description of
  Complex Scenes}. In \bibinfo{booktitle}{\emph{Proceedings of {AAAI} 2018, New
  Orleans, Louisiana, USA, February 2-7, 2018}}. \bibinfo{pages}{7371--7379}.
\newblock


\bibitem[\protect\citeauthoryear{Vinyals and Le}{Vinyals and Le}{2015}]%
        {vinyals2015neural}
\bibfield{author}{\bibinfo{person}{Oriol Vinyals} {and}
  \bibinfo{person}{Quoc~V. Le}.} \bibinfo{year}{2015}\natexlab{}.
\newblock \showarticletitle{A Neural Conversational Model}.
\newblock \bibinfo{journal}{\emph{CoRR}}  \bibinfo{volume}{abs/1506.05869}
  (\bibinfo{year}{2015}).
\newblock


\bibitem[\protect\citeauthoryear{Wallace}{Wallace}{2009}]%
        {wallace2009anatomy}
\bibfield{author}{\bibinfo{person}{Richard~S Wallace}.}
  \bibinfo{year}{2009}\natexlab{}.
\newblock \showarticletitle{The anatomy of ALICE}.
\newblock In \bibinfo{booktitle}{\emph{Parsing the Turing Test}}.
  \bibinfo{publisher}{Springer}, \bibinfo{pages}{181--210}.
\newblock


\bibitem[\protect\citeauthoryear{Wang, Jojic, Brockett, and Nyberg}{Wang
  et~al\mbox{.}}{2017b}]%
        {wang2017steering}
\bibfield{author}{\bibinfo{person}{Di Wang}, \bibinfo{person}{Nebojsa Jojic},
  \bibinfo{person}{Chris Brockett}, {and} \bibinfo{person}{Eric Nyberg}.}
  \bibinfo{year}{2017}\natexlab{b}.
\newblock \showarticletitle{Steering Output Style and Topic in Neural Response
  Generation}. In \bibinfo{booktitle}{\emph{Proceedings of {EMNLP} 2017,
  Copenhagen, Denmark, September 9-11, 2017}}. \bibinfo{pages}{2140--2150}.
\newblock


\bibitem[\protect\citeauthoryear{Wang, Wang, Li, Xu, Wang, and Wang}{Wang
  et~al\mbox{.}}{2017c}]%
        {wang2017groupbias}
\bibfield{author}{\bibinfo{person}{Jianan Wang}, \bibinfo{person}{Xin Wang},
  \bibinfo{person}{Fang Li}, \bibinfo{person}{Zhen Xu},
  \bibinfo{person}{Zhuoran Wang}, {and} \bibinfo{person}{Baoxun Wang}.}
  \bibinfo{year}{2017}\natexlab{c}.
\newblock \showarticletitle{Group Linguistic Bias Aware Neural Response
  Generation}. In \bibinfo{booktitle}{\emph{Proceedings of the 9th {SIGHAN}
  Workshop on Chinese Language Processing, SIGHAN@IJCNLP 2017, Taipei, Taiwan,
  December 1, 2017}}. \bibinfo{pages}{1--10}.
\newblock


\bibitem[\protect\citeauthoryear{Wang and Wan}{Wang and Wan}{2018}]%
        {wang2018sentigan}
\bibfield{author}{\bibinfo{person}{Ke Wang} {and} \bibinfo{person}{Xiaojun
  Wan}.} \bibinfo{year}{2018}\natexlab{}.
\newblock \showarticletitle{SentiGAN: Generating Sentimental Texts via Mixture
  Adversarial Networks}. In \bibinfo{booktitle}{\emph{Proceedings of {IJCAI}
  2018, Stockholm, Sweden, July 13-19, 2018}}. \bibinfo{pages}{4446--4452}.
\newblock


\bibitem[\protect\citeauthoryear{Wang, Huang, Xu, Shen, and Nie}{Wang
  et~al\mbox{.}}{2018a}]%
        {wang2018chat}
\bibfield{author}{\bibinfo{person}{Wenjie Wang}, \bibinfo{person}{Minlie
  Huang}, \bibinfo{person}{Xin{-}Shun Xu}, \bibinfo{person}{Fumin Shen}, {and}
  \bibinfo{person}{Liqiang Nie}.} \bibinfo{year}{2018}\natexlab{a}.
\newblock \showarticletitle{Chat More: Deepening and Widening the Chatting
  Topic via {A} Deep Model}. In \bibinfo{booktitle}{\emph{{SIGIR} 2018, Ann
  Arbor, MI, USA, July 08-12, 2018}}. \bibinfo{pages}{255--264}.
\newblock


\bibitem[\protect\citeauthoryear{Wang, Shi, Kim, Oh, Yang, Zhang, and Yu}{Wang
  et~al\mbox{.}}{2019}]%
        {wang2019persuasion}
\bibfield{author}{\bibinfo{person}{Xuewei Wang}, \bibinfo{person}{Weiyan Shi},
  \bibinfo{person}{Richard Kim}, \bibinfo{person}{Yoojung Oh},
  \bibinfo{person}{Sijia Yang}, \bibinfo{person}{Jingwen Zhang}, {and}
  \bibinfo{person}{Zhou Yu}.} \bibinfo{year}{2019}\natexlab{}.
\newblock \showarticletitle{Persuasion for Good: Towards a Personalized
  Persuasive Dialogue System for Social Good}. In
  \bibinfo{booktitle}{\emph{Proceedings of {ACL} 2019, Florence, Italy, July 28
  - Aug 2, 2019}}. \bibinfo{pages}{5635--5649}.
\newblock


\bibitem[\protect\citeauthoryear{Wang, Liu, Huang, and Nie}{Wang
  et~al\mbox{.}}{2018b}]%
        {askquestion18}
\bibfield{author}{\bibinfo{person}{Yansen Wang}, \bibinfo{person}{Chenyi Liu},
  \bibinfo{person}{Minlie Huang}, {and} \bibinfo{person}{Liqiang Nie}.}
  \bibinfo{year}{2018}\natexlab{b}.
\newblock \showarticletitle{Learning to Ask Questions in Open-domain
  Conversational Systems with Typed Decoders}. In
  \bibinfo{booktitle}{\emph{Proceedings of {ACL} 2018, Melbourne, Australia,
  July 15-20, 2018}}. \bibinfo{pages}{2193--2203}.
\newblock


\bibitem[\protect\citeauthoryear{Wang, Hamza, and Florian}{Wang
  et~al\mbox{.}}{2017a}]%
        {Wang2017bimpm}
\bibfield{author}{\bibinfo{person}{Zhiguo Wang}, \bibinfo{person}{Wael Hamza},
  {and} \bibinfo{person}{Radu Florian}.} \bibinfo{year}{2017}\natexlab{a}.
\newblock \showarticletitle{Bilateral Multi-Perspective Matching for Natural
  Language Sentences}. In \bibinfo{booktitle}{\emph{Proceedings of {IJCAI}
  2017, Melbourne, Australia, August 19-25, 2017}}.
  \bibinfo{pages}{4144--4150}.
\newblock


\bibitem[\protect\citeauthoryear{Warriner, Kuperman, and Brysbaert}{Warriner
  et~al\mbox{.}}{2013}]%
        {warriner2013norms}
\bibfield{author}{\bibinfo{person}{Amy~Beth Warriner}, \bibinfo{person}{Victor
  Kuperman}, {and} \bibinfo{person}{Marc Brysbaert}.}
  \bibinfo{year}{2013}\natexlab{}.
\newblock \showarticletitle{Norms of valence, arousal, and dominance for 13,915
  English lemmas}.
\newblock \bibinfo{journal}{\emph{Behavior research methods}}
  \bibinfo{volume}{45}, \bibinfo{number}{4} (\bibinfo{year}{2013}),
  \bibinfo{pages}{1191--1207}.
\newblock


\bibitem[\protect\citeauthoryear{Weizenbaum}{Weizenbaum}{1966}]%
        {weizenbaum1966eliza}
\bibfield{author}{\bibinfo{person}{Joseph Weizenbaum}.}
  \bibinfo{year}{1966}\natexlab{}.
\newblock \showarticletitle{{ELIZA} - a computer program for the study of
  natural language communication between man and machine}.
\newblock \bibinfo{journal}{\emph{Commun. {ACM}}} \bibinfo{volume}{9},
  \bibinfo{number}{1} (\bibinfo{year}{1966}), \bibinfo{pages}{36--45}.
\newblock


\bibitem[\protect\citeauthoryear{Welleck, Weston, Szlam, and Cho}{Welleck
  et~al\mbox{.}}{2018}]%
        {welleck2018NLI}
\bibfield{author}{\bibinfo{person}{Sean Welleck}, \bibinfo{person}{Jason
  Weston}, \bibinfo{person}{Arthur Szlam}, {and} \bibinfo{person}{Kyunghyun
  Cho}.} \bibinfo{year}{2018}\natexlab{}.
\newblock \showarticletitle{Dialogue natural language inference}.
\newblock \bibinfo{journal}{\emph{arXiv preprint arXiv:1811.00671}}
  (\bibinfo{year}{2018}).
\newblock


\bibitem[\protect\citeauthoryear{Weston, Dinan, and Miller}{Weston
  et~al\mbox{.}}{2018}]%
        {weston2018retrieve}
\bibfield{author}{\bibinfo{person}{Jason Weston}, \bibinfo{person}{Emily
  Dinan}, {and} \bibinfo{person}{Alexander~H Miller}.}
  \bibinfo{year}{2018}\natexlab{}.
\newblock \showarticletitle{Retrieve and Refine: Improved Sequence Generation
  Models For Dialogue}.
\newblock \bibinfo{journal}{\emph{EMNLP 2018}} (\bibinfo{year}{2018}),
  \bibinfo{pages}{87}.
\newblock


\bibitem[\protect\citeauthoryear{Winata, Kampman, Yang, Dey, and Fung}{Winata
  et~al\mbox{.}}{2017}]%
        {winata2017nora}
\bibfield{author}{\bibinfo{person}{Genta~Indra Winata}, \bibinfo{person}{Onno
  Kampman}, \bibinfo{person}{Yang Yang}, \bibinfo{person}{Anik Dey}, {and}
  \bibinfo{person}{Pascale Fung}.} \bibinfo{year}{2017}\natexlab{}.
\newblock \showarticletitle{Nora the empathetic psychologist}. In
  \bibinfo{booktitle}{\emph{Proc. Interspeech}}. \bibinfo{pages}{3437--3438}.
\newblock


\bibitem[\protect\citeauthoryear{Wolf, Sanh, Chaumond, and Delangue}{Wolf
  et~al\mbox{.}}{2018}]%
        {wolf2018transfer}
\bibfield{author}{\bibinfo{person}{Thomas Wolf}, \bibinfo{person}{Victor Sanh},
  \bibinfo{person}{Julien Chaumond}, {and} \bibinfo{person}{Clement Delangue}.}
  \bibinfo{year}{2018}\natexlab{}.
\newblock \showarticletitle{TransferTransfo: A Transfer Learning Approach for
  Neural Network Based Conversational Agents}. In
  \bibinfo{booktitle}{\emph{NIPS2018 CAI Workshop}}.
\newblock


\bibitem[\protect\citeauthoryear{Wu, Guo, Zhou, Wu, Zhang, Lian, and Wang}{Wu
  et~al\mbox{.}}{2019a}]%
        {wu2019proactive}
\bibfield{author}{\bibinfo{person}{Wenquan Wu}, \bibinfo{person}{Zhen Guo},
  \bibinfo{person}{Xiangyang Zhou}, \bibinfo{person}{Hua Wu},
  \bibinfo{person}{Xiyuan Zhang}, \bibinfo{person}{Rongzhong Lian}, {and}
  \bibinfo{person}{Haifeng Wang}.} \bibinfo{year}{2019}\natexlab{a}.
\newblock \showarticletitle{Proactive Human-Machine Conversation with Explicit
  Conversation Goals}.
\newblock \bibinfo{journal}{\emph{arXiv preprint arXiv:1906.05572}}
  (\bibinfo{year}{2019}).
\newblock


\bibitem[\protect\citeauthoryear{Wu, Li, Wu, and Zhou}{Wu
  et~al\mbox{.}}{2018}]%
        {wu2018response}
\bibfield{author}{\bibinfo{person}{Yu Wu}, \bibinfo{person}{Zhoujun Li},
  \bibinfo{person}{Wei Wu}, {and} \bibinfo{person}{Ming Zhou}.}
  \bibinfo{year}{2018}\natexlab{}.
\newblock \showarticletitle{Response selection with topic clues for
  retrieval-based chatbots}.
\newblock \bibinfo{journal}{\emph{Neurocomputing}}  \bibinfo{volume}{316}
  (\bibinfo{year}{2018}), \bibinfo{pages}{251--261}.
\newblock


\bibitem[\protect\citeauthoryear{Wu, Wei, Huang, Li, and Zhou}{Wu
  et~al\mbox{.}}{2019b}]%
        {wu2019prototype}
\bibfield{author}{\bibinfo{person}{Yu Wu}, \bibinfo{person}{Furu Wei},
  \bibinfo{person}{Shaohan Huang}, \bibinfo{person}{Zhoujun Li}, {and}
  \bibinfo{person}{Ming Zhou}.} \bibinfo{year}{2019}\natexlab{b}.
\newblock \showarticletitle{Response Generation by Context-aware Prototype
  Editing}. In \bibinfo{booktitle}{\emph{Proceedings of {AAAI} 2019, Honolulu,
  Hawaii, USA, January 27-February 1, 2019}}.
\newblock


\bibitem[\protect\citeauthoryear{Wu, Wu, Xing, Zhou, and Li}{Wu
  et~al\mbox{.}}{2017}]%
        {wu2017sequential}
\bibfield{author}{\bibinfo{person}{Yu Wu}, \bibinfo{person}{Wei Wu},
  \bibinfo{person}{Chen Xing}, \bibinfo{person}{Ming Zhou}, {and}
  \bibinfo{person}{Zhoujun Li}.} \bibinfo{year}{2017}\natexlab{}.
\newblock \showarticletitle{Sequential Matching Network: {A} New Architecture
  for Multi-turn Response Selection in Retrieval-Based Chatbots}. In
  \bibinfo{booktitle}{\emph{Proceedings of {ACL} 2017, Vancouver, Canada, July
  30-August 4, 2017}}. \bibinfo{pages}{496--505}.
\newblock


\bibitem[\protect\citeauthoryear{Xing, Wu, Wu, Liu, Huang, Zhou, and Ma}{Xing
  et~al\mbox{.}}{2017}]%
        {xing2017topic}
\bibfield{author}{\bibinfo{person}{Chen Xing}, \bibinfo{person}{Wei Wu},
  \bibinfo{person}{Yu Wu}, \bibinfo{person}{Jie Liu}, \bibinfo{person}{Yalou
  Huang}, \bibinfo{person}{Ming Zhou}, {and} \bibinfo{person}{Wei{-}Ying Ma}.}
  \bibinfo{year}{2017}\natexlab{}.
\newblock \showarticletitle{Topic Aware Neural Response Generation}. In
  \bibinfo{booktitle}{\emph{Proceedings of {AAAI} 2017, San Francisco,
  California, {USA.}, February 4-9, 2017}}. \bibinfo{pages}{3351--3357}.
\newblock


\bibitem[\protect\citeauthoryear{Xu, Ren, Lin, and Sun}{Xu
  et~al\mbox{.}}{2018b}]%
        {xu2018dpgan}
\bibfield{author}{\bibinfo{person}{Jingjing Xu}, \bibinfo{person}{Xuancheng
  Ren}, \bibinfo{person}{Junyang Lin}, {and} \bibinfo{person}{Xu Sun}.}
  \bibinfo{year}{2018}\natexlab{b}.
\newblock \showarticletitle{Diversity-Promoting {GAN:} {A} Cross-Entropy Based
  Generative Adversarial Network for Diversified Text Generation}. In
  \bibinfo{booktitle}{\emph{Proceedings of {EMNLP} 2018, Brussels, Belgium,
  October 31 - November 4, 2018}}. \bibinfo{pages}{3940--3949}.
\newblock


\bibitem[\protect\citeauthoryear{Xu, Madotto, Wu, Park, and Fung}{Xu
  et~al\mbox{.}}{2018a}]%
        {xu2018emo2vec}
\bibfield{author}{\bibinfo{person}{Peng Xu}, \bibinfo{person}{Andrea Madotto},
  \bibinfo{person}{Chien-Sheng Wu}, \bibinfo{person}{Ji~Ho Park}, {and}
  \bibinfo{person}{Pascale Fung}.} \bibinfo{year}{2018}\natexlab{a}.
\newblock \showarticletitle{Emo2Vec: Learning Generalized Emotion
  Representation by Multi-task Training}.
\newblock \bibinfo{journal}{\emph{arXiv preprint arXiv:1809.04505}}
  (\bibinfo{year}{2018}).
\newblock


\bibitem[\protect\citeauthoryear{Yan, Song, and Wu}{Yan et~al\mbox{.}}{2016}]%
        {DL2R}
\bibfield{author}{\bibinfo{person}{Rui Yan}, \bibinfo{person}{Yiping Song},
  {and} \bibinfo{person}{Hua Wu}.} \bibinfo{year}{2016}\natexlab{}.
\newblock \showarticletitle{Learning to Respond with Deep Neural Networks for
  Retrieval-Based Human-Computer Conversation System}. In
  \bibinfo{booktitle}{\emph{Proceedings of {SIGIR} 2016, Pisa, Italy, July
  17-21, 2016}}. \bibinfo{pages}{55--64}.
\newblock


\bibitem[\protect\citeauthoryear{Yan and Zhao}{Yan and Zhao}{2018}]%
        {prosuggestion18}
\bibfield{author}{\bibinfo{person}{Rui Yan} {and} \bibinfo{person}{Dongyan
  Zhao}.} \bibinfo{year}{2018}\natexlab{}.
\newblock \showarticletitle{Smarter Response with Proactive Suggestion: {A} New
  Generative Neural Conversation Paradigm}. In
  \bibinfo{booktitle}{\emph{Proceedings of {IJCAI} 2018, Stockholm, Sweden,
  July 13-19, 2018}}. \bibinfo{pages}{4525--4531}.
\newblock


\bibitem[\protect\citeauthoryear{Yang, Hu, Qiu, Qu, Gao, Croft, Liu, Shen, and
  Liu}{Yang et~al\mbox{.}}{2019}]%
        {yang2019hybrid}
\bibfield{author}{\bibinfo{person}{Liu Yang}, \bibinfo{person}{Junjie Hu},
  \bibinfo{person}{Minghui Qiu}, \bibinfo{person}{Chen Qu},
  \bibinfo{person}{Jianfeng Gao}, \bibinfo{person}{W~Bruce Croft},
  \bibinfo{person}{Xiaodong Liu}, \bibinfo{person}{Yelong Shen}, {and}
  \bibinfo{person}{Jingjing Liu}.} \bibinfo{year}{2019}\natexlab{}.
\newblock \showarticletitle{A Hybrid Retrieval-Generation Neural Conversation
  Model}.
\newblock \bibinfo{journal}{\emph{arXiv preprint arXiv:1904.09068}}
  (\bibinfo{year}{2019}).
\newblock


\bibitem[\protect\citeauthoryear{Yang, Qiu, Qu, Guo, Zhang, Croft, Huang, and
  Chen}{Yang et~al\mbox{.}}{2018}]%
        {yang2018response}
\bibfield{author}{\bibinfo{person}{Liu Yang}, \bibinfo{person}{Minghui Qiu},
  \bibinfo{person}{Chen Qu}, \bibinfo{person}{Jiafeng Guo},
  \bibinfo{person}{Yongfeng Zhang}, \bibinfo{person}{W~Bruce Croft},
  \bibinfo{person}{Jun Huang}, {and} \bibinfo{person}{Haiqing Chen}.}
  \bibinfo{year}{2018}\natexlab{}.
\newblock \showarticletitle{Response ranking with deep matching networks and
  external knowledge in information-seeking conversation systems}. In
  \bibinfo{booktitle}{\emph{Proceedings of {SIGIR} 2018, Ann Arbor Micigan,
  {USA}, July 8-12, 2018}}. \bibinfo{pages}{245--254}.
\newblock


\bibitem[\protect\citeauthoryear{Yang, Zhao, Zhao, Chen, Zhu, Zhou, and
  Cao}{Yang et~al\mbox{.}}{2017}]%
        {npm}
\bibfield{author}{\bibinfo{person}{Min Yang}, \bibinfo{person}{Zhou Zhao},
  \bibinfo{person}{Wei Zhao}, \bibinfo{person}{Xiaojun Chen},
  \bibinfo{person}{Jia Zhu}, \bibinfo{person}{Lianqiang Zhou}, {and}
  \bibinfo{person}{Zigang Cao}.} \bibinfo{year}{2017}\natexlab{}.
\newblock \showarticletitle{Personalized Response Generation via Domain
  adaptation}. In \bibinfo{booktitle}{\emph{Proceedings of {SIGIR} 2017, Tokyo,
  Japan, August 7-11, 2017}}. \bibinfo{pages}{1021--1024}.
\newblock


\bibitem[\protect\citeauthoryear{Young, Cambria, Chaturvedi, Zhou, Biswas, and
  Huang}{Young et~al\mbox{.}}{2018}]%
        {young2018augmenting}
\bibfield{author}{\bibinfo{person}{Tom Young}, \bibinfo{person}{Erik Cambria},
  \bibinfo{person}{Iti Chaturvedi}, \bibinfo{person}{Hao Zhou},
  \bibinfo{person}{Subham Biswas}, {and} \bibinfo{person}{Minlie Huang}.}
  \bibinfo{year}{2018}\natexlab{}.
\newblock \showarticletitle{Augmenting End-to-End Dialogue Systems With
  Commonsense Knowledge}. In \bibinfo{booktitle}{\emph{Proceedings of {AAAI}
  2018, New Orleans, Louisiana, USA, February 2-7, 2018}}.
  \bibinfo{pages}{4970--4977}.
\newblock


\bibitem[\protect\citeauthoryear{Yu, Xu, Black, and Rudnicky}{Yu
  et~al\mbox{.}}{2016}]%
        {Yu2016strategy}
\bibfield{author}{\bibinfo{person}{Zhou Yu}, \bibinfo{person}{Ziyu Xu},
  \bibinfo{person}{Alan~W. Black}, {and} \bibinfo{person}{Alexander~I.
  Rudnicky}.} \bibinfo{year}{2016}\natexlab{}.
\newblock \showarticletitle{Strategy and Policy Learning for Non-Task-Oriented
  Conversational Systems}. In \bibinfo{booktitle}{\emph{Proceedings of
  {SIGDIAL} 2016, Los Angeles, CA, {USA}, September 13-15, 2016}}.
  \bibinfo{pages}{404--412}.
\newblock


\bibitem[\protect\citeauthoryear{Zhang, Chang, Danescu{-}Niculescu{-}Mizil,
  Dixon, Hua, Taraborelli, and Thain}{Zhang et~al\mbox{.}}{2018a}]%
        {earlysign18}
\bibfield{author}{\bibinfo{person}{Justine Zhang}, \bibinfo{person}{Jonathan~P.
  Chang}, \bibinfo{person}{Cristian Danescu{-}Niculescu{-}Mizil},
  \bibinfo{person}{Lucas Dixon}, \bibinfo{person}{Yiqing Hua},
  \bibinfo{person}{Dario Taraborelli}, {and} \bibinfo{person}{Nithum Thain}.}
  \bibinfo{year}{2018}\natexlab{a}.
\newblock \showarticletitle{Conversations Gone Awry: Detecting Early Signs of
  Conversational Failure}. In \bibinfo{booktitle}{\emph{Proceedings of {ACL}
  2018, Melbourne, Australia, July 15-20, 2018}}. \bibinfo{pages}{1350--1361}.
\newblock


\bibitem[\protect\citeauthoryear{Zhang, Filbin, Morrison, Weiser, and
  Danescu-Niculescu-Mizil}{Zhang et~al\mbox{.}}{2019a}]%
        {zhang2019finding}
\bibfield{author}{\bibinfo{person}{Justine Zhang}, \bibinfo{person}{Robert
  Filbin}, \bibinfo{person}{Christine Morrison}, \bibinfo{person}{Jaclyn
  Weiser}, {and} \bibinfo{person}{Cristian Danescu-Niculescu-Mizil}.}
  \bibinfo{year}{2019}\natexlab{a}.
\newblock \showarticletitle{Finding Your Voice: The Linguistic Development of
  Mental Health Counselors}.
\newblock \bibinfo{journal}{\emph{arXiv preprint arXiv:1906.07194}}
  (\bibinfo{year}{2019}).
\newblock


\bibitem[\protect\citeauthoryear{Zhang, Tao, Xu, Xie, Chen, and Yan}{Zhang
  et~al\mbox{.}}{2019g}]%
        {zhang2019ensemblegan}
\bibfield{author}{\bibinfo{person}{Jiayi Zhang}, \bibinfo{person}{Chongyang
  Tao}, \bibinfo{person}{Zhenjing Xu}, \bibinfo{person}{Qiaojing Xie},
  \bibinfo{person}{Wei Chen}, {and} \bibinfo{person}{Rui Yan}.}
  \bibinfo{year}{2019}\natexlab{g}.
\newblock \showarticletitle{EnsembleGAN: Adversarial Learning for
  Retrieval-Generation Ensemble Model on Short-Text Conversation}. In
  \bibinfo{booktitle}{\emph{Proceedings of the 42nd International ACM SIGIR
  Conference on Research and Development in Information Retrieval}}. ACM,
  \bibinfo{pages}{435--444}.
\newblock


\bibitem[\protect\citeauthoryear{Zhang, Guo, Fan, Lan, Xu, and Cheng}{Zhang
  et~al\mbox{.}}{2018c}]%
        {zhang2018learning}
\bibfield{author}{\bibinfo{person}{Ruqing Zhang}, \bibinfo{person}{Jiafeng
  Guo}, \bibinfo{person}{Yixing Fan}, \bibinfo{person}{Yanyan Lan},
  \bibinfo{person}{Jun Xu}, {and} \bibinfo{person}{Xueqi Cheng}.}
  \bibinfo{year}{2018}\natexlab{c}.
\newblock \showarticletitle{Learning to Control the Specificity in Neural
  Response Generation}. In \bibinfo{booktitle}{\emph{Proceedings of {ACL} 2018,
  Melbourne, Australia, July 15-20, 2018}}. \bibinfo{pages}{1108--1117}.
\newblock


\bibitem[\protect\citeauthoryear{Zhang, Lee, Polymenakos, and Radev}{Zhang
  et~al\mbox{.}}{2018d}]%
        {zhang2018addressee}
\bibfield{author}{\bibinfo{person}{Rui Zhang}, \bibinfo{person}{Honglak Lee},
  \bibinfo{person}{Lazaros Polymenakos}, {and} \bibinfo{person}{Dragomir~R.
  Radev}.} \bibinfo{year}{2018}\natexlab{d}.
\newblock \showarticletitle{Addressee and Response Selection in Multi-Party
  Conversations With Speaker Interaction RNNs}. In
  \bibinfo{booktitle}{\emph{Proceedings of {AAAI} 2018, New Orleans, Louisiana,
  USA, February 2-7, 2018}}. \bibinfo{pages}{5690--5697}.
\newblock


\bibitem[\protect\citeauthoryear{Zhang, Dinan, Urbanek, Szlam, Kiela, and
  Weston}{Zhang et~al\mbox{.}}{2018b}]%
        {Zhang2018Personalizing-dogpet}
\bibfield{author}{\bibinfo{person}{Saizheng Zhang}, \bibinfo{person}{Emily
  Dinan}, \bibinfo{person}{Jack Urbanek}, \bibinfo{person}{Arthur Szlam},
  \bibinfo{person}{Douwe Kiela}, {and} \bibinfo{person}{Jason Weston}.}
  \bibinfo{year}{2018}\natexlab{b}.
\newblock \showarticletitle{Personalizing Dialogue Agents: {I} have a dog, do
  you have pets too?}. In \bibinfo{booktitle}{\emph{Proceedings of {ACL} 2018,
  Melbourne, Australia, July 15-20, 2018}}. \bibinfo{pages}{2204--2213}.
\newblock


\bibitem[\protect\citeauthoryear{Zhang, Zhu, Wang, Zhao, and Liu}{Zhang
  et~al\mbox{.}}{2017}]%
        {zhang2017neural}
\bibfield{author}{\bibinfo{person}{Wei-Nan Zhang}, \bibinfo{person}{Qingfu
  Zhu}, \bibinfo{person}{Yifa Wang}, \bibinfo{person}{Yanyan Zhao}, {and}
  \bibinfo{person}{Ting Liu}.} \bibinfo{year}{2017}\natexlab{}.
\newblock \showarticletitle{Neural personalized response generation as domain
  adaptation}.
\newblock \bibinfo{journal}{\emph{World Wide Web}} (\bibinfo{year}{2017}),
  \bibinfo{pages}{1--20}.
\newblock


\bibitem[\protect\citeauthoryear{Zhang, Gao, Lee, Brockett, Galley, Gao, and
  Dolan}{Zhang et~al\mbox{.}}{2019b}]%
        {zhang2019consistent}
\bibfield{author}{\bibinfo{person}{Yizhe Zhang}, \bibinfo{person}{Xiang Gao},
  \bibinfo{person}{Sungjin Lee}, \bibinfo{person}{Chris Brockett},
  \bibinfo{person}{Michel Galley}, \bibinfo{person}{Jianfeng Gao}, {and}
  \bibinfo{person}{Bill Dolan}.} \bibinfo{year}{2019}\natexlab{b}.
\newblock \showarticletitle{Consistent Dialogue Generation with Self-supervised
  Feature Learning}.
\newblock \bibinfo{journal}{\emph{arXiv preprint arXiv:1903.05759}}
  (\bibinfo{year}{2019}).
\newblock


\bibitem[\protect\citeauthoryear{Zhang, Ren, and de~Rijke}{Zhang
  et~al\mbox{.}}{2019e}]%
        {zhang2019improving}
\bibfield{author}{\bibinfo{person}{Yangjun Zhang}, \bibinfo{person}{Pengjie
  Ren}, {and} \bibinfo{person}{Maarten de Rijke}.}
  \bibinfo{year}{2019}\natexlab{e}.
\newblock \showarticletitle{Improving Background Based Conversation with
  Context-aware Knowledge Pre-selection}.
\newblock \bibinfo{journal}{\emph{arXiv preprint arXiv:1906.06685}}
  (\bibinfo{year}{2019}).
\newblock


\bibitem[\protect\citeauthoryear{Zhang, Sun, Galley, Chen, Brockett, Gao, Gao,
  Liu, and Dolan}{Zhang et~al\mbox{.}}{2019f}]%
        {zhang2019dialogpt}
\bibfield{author}{\bibinfo{person}{Yizhe Zhang}, \bibinfo{person}{Siqi Sun},
  \bibinfo{person}{Michel Galley}, \bibinfo{person}{Yen-Chun Chen},
  \bibinfo{person}{Chris Brockett}, \bibinfo{person}{Xiang Gao},
  \bibinfo{person}{Jianfeng Gao}, \bibinfo{person}{Jingjing Liu}, {and}
  \bibinfo{person}{Bill Dolan}.} \bibinfo{year}{2019}\natexlab{f}.
\newblock \bibinfo{title}{DialoGPT: Large-Scale Generative Pre-training for
  Conversational Response Generation}.
\newblock
\newblock
\showeprint[arxiv]{cs.CL/1911.00536}


\bibitem[\protect\citeauthoryear{Zhang, Huang, Zhao, Ji, Chen, and Zhu}{Zhang
  et~al\mbox{.}}{2019c}]%
        {zhang2019tois}
\bibfield{author}{\bibinfo{person}{Zheng Zhang}, \bibinfo{person}{Minlie
  Huang}, \bibinfo{person}{Zhongzhou Zhao}, \bibinfo{person}{Feng Ji},
  \bibinfo{person}{Haiqing Chen}, {and} \bibinfo{person}{Xiaoyan Zhu}.}
  \bibinfo{year}{2019}\natexlab{c}.
\newblock \showarticletitle{Memory-augmented Dialogue Management for
  Task-oriented Dialogue Systems}.
\newblock \bibinfo{journal}{\emph{ACM Transactions on Information Systems}}
  \bibinfo{volume}{1} (\bibinfo{year}{2019}).
\newblock


\bibitem[\protect\citeauthoryear{Zhang, Li, Zhu, Zhao, and Liu}{Zhang
  et~al\mbox{.}}{2018e}]%
        {zhang2018modeling}
\bibfield{author}{\bibinfo{person}{Zhuosheng Zhang}, \bibinfo{person}{Jiangtong
  Li}, \bibinfo{person}{Pengfei Zhu}, \bibinfo{person}{Hai Zhao}, {and}
  \bibinfo{person}{Gongshen Liu}.} \bibinfo{year}{2018}\natexlab{e}.
\newblock \showarticletitle{Modeling Multi-turn Conversation with Deep
  Utterance Aggregation}. In \bibinfo{booktitle}{\emph{Proceedings of the 27th
  International Conference on Computational Linguistics}}.
  \bibinfo{pages}{3740--3752}.
\newblock


\bibitem[\protect\citeauthoryear{Zhang, Liao, Huang, Zhu, and Chua}{Zhang
  et~al\mbox{.}}{2019d}]%
        {zhang2019mnbt}
\bibfield{author}{\bibinfo{person}{Zheng Zhang}, \bibinfo{person}{Lizi Liao},
  \bibinfo{person}{Minlie Huang}, \bibinfo{person}{Xiaoyan Zhu}, {and}
  \bibinfo{person}{Tat-Seng Chua}.} \bibinfo{year}{2019}\natexlab{d}.
\newblock \showarticletitle{Neural Multimodal Belief Tracker with Adaptive
  Attention for Dialogue Systems}. In \bibinfo{booktitle}{\emph{The World Wide
  Web Conference}}. ACM, \bibinfo{pages}{2401--2412}.
\newblock


\bibitem[\protect\citeauthoryear{Zhao and Esk{\'{e}}nazi}{Zhao and
  Esk{\'{e}}nazi}{2016}]%
        {zhao2016towards}
\bibfield{author}{\bibinfo{person}{Tiancheng Zhao} {and}
  \bibinfo{person}{Maxine Esk{\'{e}}nazi}.} \bibinfo{year}{2016}\natexlab{}.
\newblock \showarticletitle{Towards End-to-End Learning for Dialog State
  Tracking and Management using Deep Reinforcement Learning}. In
  \bibinfo{booktitle}{\emph{Proceedings of {SIGDIAL} 2016, Los Angeles, CA,
  {USA}, September 13-15, 2016}}. \bibinfo{pages}{1--10}.
\newblock


\bibitem[\protect\citeauthoryear{Zhao, Lee, and Esk{\'{e}}nazi}{Zhao
  et~al\mbox{.}}{2018}]%
        {zhao18discrete}
\bibfield{author}{\bibinfo{person}{Tiancheng Zhao}, \bibinfo{person}{Kyusong
  Lee}, {and} \bibinfo{person}{Maxine Esk{\'{e}}nazi}.}
  \bibinfo{year}{2018}\natexlab{}.
\newblock \showarticletitle{Unsupervised Discrete Sentence Representation
  Learning for Interpretable Neural Dialog Generation}. In
  \bibinfo{booktitle}{\emph{Proceedings of {ACL} 2018, Melbourne, Australia,
  July 15-20, 2018}}. \bibinfo{pages}{1098--1107}.
\newblock


\bibitem[\protect\citeauthoryear{Zhao, Zhao, and Esk{\'{e}}nazi}{Zhao
  et~al\mbox{.}}{2017}]%
        {zhao2017cvae}
\bibfield{author}{\bibinfo{person}{Tiancheng Zhao}, \bibinfo{person}{Ran Zhao},
  {and} \bibinfo{person}{Maxine Esk{\'{e}}nazi}.}
  \bibinfo{year}{2017}\natexlab{}.
\newblock \showarticletitle{Learning Discourse-level Diversity for Neural
  Dialog Models using Conditional Variational Autoencoders}. In
  \bibinfo{booktitle}{\emph{Proceedings of {ACL} 2017, Vancouver, Canada, July
  30-August 4, 2017}}. \bibinfo{pages}{654--664}.
\newblock


\bibitem[\protect\citeauthoryear{Zheng, Chen, Huang, Liu, and Zhu}{Zheng
  et~al\mbox{.}}{2019}]%
        {zheng2019personalized}
\bibfield{author}{\bibinfo{person}{Yinhe Zheng}, \bibinfo{person}{Guanyi Chen},
  \bibinfo{person}{Minlie Huang}, \bibinfo{person}{Song Liu}, {and}
  \bibinfo{person}{Xuan Zhu}.} \bibinfo{year}{2019}\natexlab{}.
\newblock \showarticletitle{Personalized Dialogue Generation with Diversified
  Traits}.
\newblock \bibinfo{journal}{\emph{CoRR}}  \bibinfo{volume}{abs/1901.09672}
  (\bibinfo{year}{2019}).
\newblock


\bibitem[\protect\citeauthoryear{Zhou, Luo, Cao, Lin, Chen, and He}{Zhou
  et~al\mbox{.}}{2017}]%
        {zhou2017mechanism}
\bibfield{author}{\bibinfo{person}{Ganbin Zhou}, \bibinfo{person}{Ping Luo},
  \bibinfo{person}{Rongyu Cao}, \bibinfo{person}{Fen Lin}, \bibinfo{person}{Bo
  Chen}, {and} \bibinfo{person}{Qing He}.} \bibinfo{year}{2017}\natexlab{}.
\newblock \showarticletitle{Mechanism-Aware Neural Machine for Dialogue
  Response Generation}. In \bibinfo{booktitle}{\emph{Proceedings of {AAAI}
  2017, San Francisco, California, {USA.}, February 4-9, 2017}}.
  \bibinfo{pages}{3400--3407}.
\newblock


\bibitem[\protect\citeauthoryear{Zhou, Luo, Xiao, Lin, Chen, and He}{Zhou
  et~al\mbox{.}}{2018d}]%
        {zhou2018elastic}
\bibfield{author}{\bibinfo{person}{Ganbin Zhou}, \bibinfo{person}{Ping Luo},
  \bibinfo{person}{Yijun Xiao}, \bibinfo{person}{Fen Lin}, \bibinfo{person}{Bo
  Chen}, {and} \bibinfo{person}{Qing He}.} \bibinfo{year}{2018}\natexlab{d}.
\newblock \showarticletitle{Elastic Responding Machine for Dialog Generation
  with Dynamically Mechanism Selecting}. In
  \bibinfo{booktitle}{\emph{Proceedings of {AAAI} 2018, New Orleans, Louisiana,
  USA, February 2-7, 2018}}. \bibinfo{pages}{5730--5737}.
\newblock


\bibitem[\protect\citeauthoryear{Zhou, Huang, Zhang, Zhu, and Liu}{Zhou
  et~al\mbox{.}}{2018b}]%
        {Zhou2018EmotionalCM}
\bibfield{author}{\bibinfo{person}{Hao Zhou}, \bibinfo{person}{Minlie Huang},
  \bibinfo{person}{Tianyang Zhang}, \bibinfo{person}{Xiaoyan Zhu}, {and}
  \bibinfo{person}{Bing Liu}.} \bibinfo{year}{2018}\natexlab{b}.
\newblock \showarticletitle{Emotional Chatting Machine: Emotional Conversation
  Generation with Internal and External Memory}. In
  \bibinfo{booktitle}{\emph{Proceedings of {AAAI} 2018, New Orleans, Louisiana,
  USA, February 2-7, 2018}}. \bibinfo{pages}{730--739}.
\newblock


\bibitem[\protect\citeauthoryear{Zhou, Young, Huang, Zhao, Xu, and Zhu}{Zhou
  et~al\mbox{.}}{2018f}]%
        {Zhou2018Commonsense}
\bibfield{author}{\bibinfo{person}{Hao Zhou}, \bibinfo{person}{Tom Young},
  \bibinfo{person}{Minlie Huang}, \bibinfo{person}{Haizhou Zhao},
  \bibinfo{person}{Jingfang Xu}, {and} \bibinfo{person}{Xiaoyan Zhu}.}
  \bibinfo{year}{2018}\natexlab{f}.
\newblock \showarticletitle{Commonsense Knowledge Aware Conversation Generation
  with Graph Attention}. In \bibinfo{booktitle}{\emph{Proceedings of {IJCAI}
  2018, Stockholm, Sweden, July 13-19, 2018}}. \bibinfo{pages}{4623--4629}.
\newblock


\bibitem[\protect\citeauthoryear{Zhou, Prabhumoye, and Black}{Zhou
  et~al\mbox{.}}{2018e}]%
        {zhou2018DoG}
\bibfield{author}{\bibinfo{person}{Kangyan Zhou}, \bibinfo{person}{Shrimai
  Prabhumoye}, {and} \bibinfo{person}{Alan~W. Black}.}
  \bibinfo{year}{2018}\natexlab{e}.
\newblock \showarticletitle{A Dataset for Document Grounded Conversations}. In
  \bibinfo{booktitle}{\emph{Proceedings of {EMNLP} 2018, Brussels, Belgium,
  October 31 - November 4, 2018}}. \bibinfo{pages}{708--713}.
\newblock


\bibitem[\protect\citeauthoryear{Zhou, Gao, Li, and Shum}{Zhou
  et~al\mbox{.}}{2018a}]%
        {zhou2018design}
\bibfield{author}{\bibinfo{person}{Li Zhou}, \bibinfo{person}{Jianfeng Gao},
  \bibinfo{person}{Di Li}, {and} \bibinfo{person}{Heung{-}Yeung Shum}.}
  \bibinfo{year}{2018}\natexlab{a}.
\newblock \showarticletitle{The Design and Implementation of XiaoIce, an
  Empathetic Social Chatbot}.
\newblock \bibinfo{journal}{\emph{CoRR}}  \bibinfo{volume}{abs/1812.08989}
  (\bibinfo{year}{2018}).
\newblock


\bibitem[\protect\citeauthoryear{Zhou, Dong, Wu, Zhao, Yu, Tian, Liu, and
  Yan}{Zhou et~al\mbox{.}}{2016}]%
        {zhou2016multi}
\bibfield{author}{\bibinfo{person}{Xiangyang Zhou}, \bibinfo{person}{Daxiang
  Dong}, \bibinfo{person}{Hua Wu}, \bibinfo{person}{Shiqi Zhao},
  \bibinfo{person}{Dianhai Yu}, \bibinfo{person}{Hao Tian},
  \bibinfo{person}{Xuan Liu}, {and} \bibinfo{person}{Rui Yan}.}
  \bibinfo{year}{2016}\natexlab{}.
\newblock \showarticletitle{Multi-view Response Selection for Human-Computer
  Conversation}. In \bibinfo{booktitle}{\emph{Proceedings of {EMNLP} 2016,
  Austin, Texas, USA, November 1-4, 2016}}. \bibinfo{pages}{372--381}.
\newblock


\bibitem[\protect\citeauthoryear{Zhou, Li, Dong, Liu, Chen, Zhao, Yu, and
  Wu}{Zhou et~al\mbox{.}}{2018c}]%
        {zhou-etal-2018-multi}
\bibfield{author}{\bibinfo{person}{Xiangyang Zhou}, \bibinfo{person}{Lu Li},
  \bibinfo{person}{Daxiang Dong}, \bibinfo{person}{Yi Liu},
  \bibinfo{person}{Ying Chen}, \bibinfo{person}{Wayne~Xin Zhao},
  \bibinfo{person}{Dianhai Yu}, {and} \bibinfo{person}{Hua Wu}.}
  \bibinfo{year}{2018}\natexlab{c}.
\newblock \showarticletitle{Multi-Turn Response Selection for Chatbots with
  Deep Attention Matching Network}. In \bibinfo{booktitle}{\emph{Proceedings of
  {ACL} 2018, Melbourne, Australia, July 15-20, 2018}}.
  \bibinfo{pages}{1118--1127}.
\newblock


\bibitem[\protect\citeauthoryear{Zhou and Wang}{Zhou and Wang}{2018}]%
        {zhou2017mojitalk}
\bibfield{author}{\bibinfo{person}{Xianda Zhou} {and}
  \bibinfo{person}{William~Yang Wang}.} \bibinfo{year}{2018}\natexlab{}.
\newblock \showarticletitle{MojiTalk: Generating Emotional Responses at Scale}.
  In \bibinfo{booktitle}{\emph{Proceedings of {ACL} 2018, Melbourne, Australia,
  July 15-20, 2018}}. \bibinfo{pages}{1128--1137}.
\newblock


\bibitem[\protect\citeauthoryear{Zhu, Mo, Zhang, Zhu, Peng, and Yang}{Zhu
  et~al\mbox{.}}{2017}]%
        {zhuwenya2017}
\bibfield{author}{\bibinfo{person}{Wenya Zhu}, \bibinfo{person}{Kaixiang Mo},
  \bibinfo{person}{Yu Zhang}, \bibinfo{person}{Zhangbin Zhu},
  \bibinfo{person}{Xuezheng Peng}, {and} \bibinfo{person}{Qiang Yang}.}
  \bibinfo{year}{2017}\natexlab{}.
\newblock \showarticletitle{Flexible End-to-End Dialogue System for Knowledge
  Grounded Conversation}.
\newblock \bibinfo{journal}{\emph{CoRR}}  \bibinfo{volume}{abs/1709.04264}
  (\bibinfo{year}{2017}).
\newblock


\end{thebibliography}



\begin{thebibliography}{138}

%%% the Web address.

\ifx \showCODEN    \undefined \def \showCODEN     #1{\unskip}     \fi
\ifx \showDOI      \undefined \def \showDOI       #1{#1}\fi
\ifx \showISBNx    \undefined \def \showISBNx     #1{\unskip}     \fi
\ifx \showISBNxiii \undefined \def \showISBNxiii  #1{\unskip}     \fi
\ifx \showISSN     \undefined \def \showISSN      #1{\unskip}     \fi
\ifx \showLCCN     \undefined \def \showLCCN      #1{\unskip}     \fi
\ifx \shownote     \undefined \def \shownote      #1{#1}          \fi
\ifx \showarticletitle \undefined \def \showarticletitle #1{#1}   \fi
\ifx \showURL      \undefined \def \showURL       {\relax}        \fi
% invisible to TeX
\providecommand\bibfield[2]{#2}
\providecommand\bibinfo[2]{#2}
\providecommand\natexlab[1]{#1}
\providecommand\showeprint[2][]{arXiv:#2}

\bibitem[\protect\citeauthoryear{Albrecht and Hwa}{Albrecht and Hwa}{2007}]%
        {albrecht07reex}
\bibfield{author}{\bibinfo{person}{Joshua Albrecht} {and}
  \bibinfo{person}{Rebecca Hwa}.} \bibinfo{year}{2007}\natexlab{}.
\newblock \showarticletitle{A Re-examination of Machine Learning Approaches for
  Sentence-Level MT Evaluation}. In \bibinfo{booktitle}{\emph{Proceedings of
  {ACL} 2007, June 23-30, 2007, Prague, Czech Republic}}.
  \bibinfo{pages}{880--887}.
\newblock


\bibitem[\protect\citeauthoryear{Asghar, Poupart, Hoey, Jiang, and Mou}{Asghar
  et~al\mbox{.}}{2018}]%
        {asghar2018affective}
\bibfield{author}{\bibinfo{person}{Nabiha Asghar}, \bibinfo{person}{Pascal
  Poupart}, \bibinfo{person}{Jesse Hoey}, \bibinfo{person}{Xin Jiang}, {and}
  \bibinfo{person}{Lili Mou}.} \bibinfo{year}{2018}\natexlab{}.
\newblock \showarticletitle{Affective Neural Response Generation}. In
  \bibinfo{booktitle}{\emph{Advances in Information Retrieval - 40th European
  Conference on {IR} Research, {ECIR} 2018, Grenoble, France, March 26-29,
  2018, Proceedings}}, Vol.~\bibinfo{volume}{10772}. \bibinfo{pages}{154--166}.
\newblock


\bibitem[\protect\citeauthoryear{Bahdanau, Cho, and Bengio}{Bahdanau
  et~al\mbox{.}}{2015}]%
        {Bahdanau2015Neural}
\bibfield{author}{\bibinfo{person}{Dzmitry Bahdanau},
  \bibinfo{person}{Kyunghyun Cho}, {and} \bibinfo{person}{Yoshua Bengio}.}
  \bibinfo{year}{2015}\natexlab{}.
\newblock \showarticletitle{Neural Machine Translation by Jointly Learning to
  Align and Translate}. In \bibinfo{booktitle}{\emph{{ICLR} 2015, San Diego,
  CA, USA, May 7-9, 2015}}.
\newblock


\bibitem[\protect\citeauthoryear{Banerjee and Lavie}{Banerjee and
  Lavie}{2005}]%
        {banerjee2005meteor}
\bibfield{author}{\bibinfo{person}{Satanjeev Banerjee} {and}
  \bibinfo{person}{Alon Lavie}.} \bibinfo{year}{2005}\natexlab{}.
\newblock \showarticletitle{METEOR: An automatic metric for MT evaluation with
  improved correlation with human judgments}. In
  \bibinfo{booktitle}{\emph{Proceedings of the acl workshop on intrinsic and
  extrinsic evaluation measures for machine translation and/or summarization}}.
  \bibinfo{pages}{65--72}.
\newblock


\bibitem[\protect\citeauthoryear{Bang, Noh, Kim, and Lee}{Bang
  et~al\mbox{.}}{2015}]%
        {Bang2015Example}
\bibfield{author}{\bibinfo{person}{Jeesoo Bang}, \bibinfo{person}{Hyungjong
  Noh}, \bibinfo{person}{Yonghee Kim}, {and} \bibinfo{person}{Gary~Geunbae
  Lee}.} \bibinfo{year}{2015}\natexlab{}.
\newblock \showarticletitle{Example-based chat-oriented dialogue system with
  personalized long-term memory}. In \bibinfo{booktitle}{\emph{2015
  International Conference on Big Data and Smart Computing, {BIGCOMP} 2015,
  Jeju, South Korea, February 9-11, 2015}}. \bibinfo{pages}{238--243}.
\newblock


\bibitem[\protect\citeauthoryear{Bordes, Boureau, and Weston}{Bordes
  et~al\mbox{.}}{2017}]%
        {bordes2016learning}
\bibfield{author}{\bibinfo{person}{Antoine Bordes}, \bibinfo{person}{Y.{-}Lan
  Boureau}, {and} \bibinfo{person}{Jason Weston}.}
  \bibinfo{year}{2017}\natexlab{}.
\newblock \showarticletitle{Learning End-to-End Goal-Oriented Dialog}. In
  \bibinfo{booktitle}{\emph{{ICLR} 2017, Toulon, France, April 24-26, 2017}}.
\newblock


\bibitem[\protect\citeauthoryear{Casanueva, Hain, Christensen, Marxer, and
  Green}{Casanueva et~al\mbox{.}}{2015}]%
        {Casanueva2015Knowledge}
\bibfield{author}{\bibinfo{person}{I{\~{n}}igo Casanueva},
  \bibinfo{person}{Thomas Hain}, \bibinfo{person}{Heidi Christensen},
  \bibinfo{person}{Ricard Marxer}, {and} \bibinfo{person}{Phil~D. Green}.}
  \bibinfo{year}{2015}\natexlab{}.
\newblock \showarticletitle{Knowledge transfer between speakers for
  personalised dialogue management}. In \bibinfo{booktitle}{\emph{Proceedings
  of {SIGDIAL} 2015, September 2-4, 2015, Prague, Czech Republic}}.
  \bibinfo{pages}{12--21}.
\newblock


\bibitem[\protect\citeauthoryear{Chaganty, Mussmann, and Liang}{Chaganty
  et~al\mbox{.}}{2018}]%
        {chaganty2018price}
\bibfield{author}{\bibinfo{person}{Arun Chaganty}, \bibinfo{person}{Stephen
  Mussmann}, {and} \bibinfo{person}{Percy Liang}.}
  \bibinfo{year}{2018}\natexlab{}.
\newblock \showarticletitle{The price of debiasing automatic metrics in natural
  language evalaution}. In \bibinfo{booktitle}{\emph{Proceedings of {ACL} 2018,
  Melbourne, Australia, July 15-20, 2018}}. \bibinfo{pages}{643--653}.
\newblock


\bibitem[\protect\citeauthoryear{Colby, Weber, and Hilf}{Colby
  et~al\mbox{.}}{1971}]%
        {colby1971artificial}
\bibfield{author}{\bibinfo{person}{Kenneth~Mark Colby}, \bibinfo{person}{Sylvia
  Weber}, {and} \bibinfo{person}{Franklin~Dennis Hilf}.}
  \bibinfo{year}{1971}\natexlab{}.
\newblock \showarticletitle{Artificial paranoia}.
\newblock \bibinfo{journal}{\emph{Artificial Intelligence}}
  \bibinfo{volume}{2}, \bibinfo{number}{1} (\bibinfo{year}{1971}),
  \bibinfo{pages}{1--25}.
\newblock


\bibitem[\protect\citeauthoryear{Corston-Oliver, Gamon, and
  Brockett}{Corston-Oliver et~al\mbox{.}}{2001}]%
        {corston01ml}
\bibfield{author}{\bibinfo{person}{Simon Corston-Oliver},
  \bibinfo{person}{Michael Gamon}, {and} \bibinfo{person}{Chris Brockett}.}
  \bibinfo{year}{2001}\natexlab{}.
\newblock \showarticletitle{A Machine Learning Approach to the Automatic
  Evaluation of Machine Translation}. In \bibinfo{booktitle}{\emph{Proceedings
  of {ACL} 2001, July 9-11, 2001, Toulouse, France}}.
  \bibinfo{pages}{148--155}.
\newblock


\bibitem[\protect\citeauthoryear{Devlin, Chang, Lee, and Toutanova}{Devlin
  et~al\mbox{.}}{2019}]%
        {devlin2018bert}
\bibfield{author}{\bibinfo{person}{Jacob Devlin}, \bibinfo{person}{Ming-Wei
  Chang}, \bibinfo{person}{Kenton Lee}, {and} \bibinfo{person}{Kristina
  Toutanova}.} \bibinfo{year}{2019}\natexlab{}.
\newblock \showarticletitle{Bert: Pre-training of deep bidirectional
  transformers for language understanding}.
\newblock \bibinfo{journal}{\emph{NAACL}} (\bibinfo{year}{2019}),
  \bibinfo{pages}{4171--4186}.
\newblock


\bibitem[\protect\citeauthoryear{Dinan, Logacheva, Malykh, Miller, Shuster,
  Urbanek, Kiela, Szlam, Serban, Lowe, Prabhumoye, Black, Rudnicky, Williams,
  Pineau, Burtsev, and Weston}{Dinan et~al\mbox{.}}{2019}]%
        {dinan2019second}
\bibfield{author}{\bibinfo{person}{Emily Dinan}, \bibinfo{person}{Varvara
  Logacheva}, \bibinfo{person}{Valentin Malykh}, \bibinfo{person}{Alexander~H.
  Miller}, \bibinfo{person}{Kurt Shuster}, \bibinfo{person}{Jack Urbanek},
  \bibinfo{person}{Douwe Kiela}, \bibinfo{person}{Arthur Szlam},
  \bibinfo{person}{Iulian Serban}, \bibinfo{person}{Ryan Lowe},
  \bibinfo{person}{Shrimai Prabhumoye}, \bibinfo{person}{Alan~W. Black},
  \bibinfo{person}{Alexander~I. Rudnicky}, \bibinfo{person}{Jason Williams},
  \bibinfo{person}{Joelle Pineau}, \bibinfo{person}{Mikhail Burtsev}, {and}
  \bibinfo{person}{Jason Weston}.} \bibinfo{year}{2019}\natexlab{}.
\newblock \showarticletitle{The Second Conversational Intelligence Challenge
  (ConvAI2)}.
\newblock \bibinfo{journal}{\emph{CoRR}}  \bibinfo{volume}{abs/1902.00098}
  (\bibinfo{year}{2019}).
\newblock


\bibitem[\protect\citeauthoryear{Dinan, Roller, Shuster, Fan, Auli, and
  Weston}{Dinan et~al\mbox{.}}{2018}]%
        {dinan2018WOW}
\bibfield{author}{\bibinfo{person}{Emily Dinan}, \bibinfo{person}{Stephen
  Roller}, \bibinfo{person}{Kurt Shuster}, \bibinfo{person}{Angela Fan},
  \bibinfo{person}{Michael Auli}, {and} \bibinfo{person}{Jason Weston}.}
  \bibinfo{year}{2018}\natexlab{}.
\newblock \showarticletitle{Wizard of Wikipedia: Knowledge-Powered
  Conversational agents}.
\newblock \bibinfo{journal}{\emph{CoRR}}  \bibinfo{volume}{abs/1811.01241}
  (\bibinfo{year}{2018}).
\newblock


\bibitem[\protect\citeauthoryear{Du, Li, He, Xu, Bing, and Wang}{Du
  et~al\mbox{.}}{2018}]%
        {du18vae}
\bibfield{author}{\bibinfo{person}{Jiachen Du}, \bibinfo{person}{Wenjie Li},
  \bibinfo{person}{Yulan He}, \bibinfo{person}{Ruifeng Xu},
  \bibinfo{person}{Lidong Bing}, {and} \bibinfo{person}{Xuan Wang}.}
  \bibinfo{year}{2018}\natexlab{}.
\newblock \showarticletitle{Variational Autoregressive Decoder for Neural
  Response Generation}. In \bibinfo{booktitle}{\emph{Proceedings of {EMNLP}
  2018, Brussels, Belgium, October 31 - November 4, 2018}}.
  \bibinfo{pages}{3154--3163}.
\newblock


\bibitem[\protect\citeauthoryear{Fan, Pang, Hou, Guo, Lan, and Cheng}{Fan
  et~al\mbox{.}}{2017}]%
        {fan2017matchzoo}
\bibfield{author}{\bibinfo{person}{Yixing Fan}, \bibinfo{person}{Liang Pang},
  \bibinfo{person}{Jianpeng Hou}, \bibinfo{person}{Jiafeng Guo},
  \bibinfo{person}{Yanyan Lan}, {and} \bibinfo{person}{Xueqi Cheng}.}
  \bibinfo{year}{2017}\natexlab{}.
\newblock \showarticletitle{MatchZoo: {A} Toolkit for Deep Text Matching}.
\newblock \bibinfo{journal}{\emph{CoRR}}  \bibinfo{volume}{abs/1707.07270}
  (\bibinfo{year}{2017}).
\newblock


\bibitem[\protect\citeauthoryear{Fang, Cheng, Sap, Clark, Holtzman, Choi,
  Smith, and Ostendorf}{Fang et~al\mbox{.}}{2018}]%
        {sounding2017}
\bibfield{author}{\bibinfo{person}{Hao Fang}, \bibinfo{person}{Hao Cheng},
  \bibinfo{person}{Maarten Sap}, \bibinfo{person}{Elizabeth Clark},
  \bibinfo{person}{Ari Holtzman}, \bibinfo{person}{Yejin Choi},
  \bibinfo{person}{Noah~A. Smith}, {and} \bibinfo{person}{Mari Ostendorf}.}
  \bibinfo{year}{2018}\natexlab{}.
\newblock \showarticletitle{Sounding Board: {A} User-Centric and Content-Driven
  Social Chatbot}. In \bibinfo{booktitle}{\emph{Proceedings of {NAACL-HLT}
  2018, New Orleans, Louisiana, USA, June 2-4, 2018, Demonstrations}}.
  \bibinfo{pages}{96--100}.
\newblock


\bibitem[\protect\citeauthoryear{Galley, Brockett, Sordoni, Ji, Auli, Quirk,
  Mitchell, Gao, and Dolan}{Galley et~al\mbox{.}}{2015}]%
        {galley2015deltableu}
\bibfield{author}{\bibinfo{person}{Michel Galley}, \bibinfo{person}{Chris
  Brockett}, \bibinfo{person}{Alessandro Sordoni}, \bibinfo{person}{Yangfeng
  Ji}, \bibinfo{person}{Michael Auli}, \bibinfo{person}{Chris Quirk},
  \bibinfo{person}{Margaret Mitchell}, \bibinfo{person}{Jianfeng Gao}, {and}
  \bibinfo{person}{Bill Dolan}.} \bibinfo{year}{2015}\natexlab{}.
\newblock \showarticletitle{{deltaBLEU}: A Discriminative Metric for Generation
  Tasks with Intrinsically Diverse Targets}. In
  \bibinfo{booktitle}{\emph{Proceedings of {ACL-IJCNLP} 2015, July 26-31, 2015,
  Beijing, China}}. \bibinfo{pages}{445--450}.
\newblock


\bibitem[\protect\citeauthoryear{Gao, Galley, and Li}{Gao
  et~al\mbox{.}}{2019a}]%
        {gao2019neural}
\bibfield{author}{\bibinfo{person}{Jianfeng Gao}, \bibinfo{person}{Michel
  Galley}, {and} \bibinfo{person}{Lihong Li}.}
  \bibinfo{year}{2019}\natexlab{a}.
\newblock \showarticletitle{Neural approaches to conversational AI}.
\newblock \bibinfo{journal}{\emph{Foundations and Trends{\textregistered} in
  Information Retrieval}} \bibinfo{volume}{13}, \bibinfo{number}{2-3}
  (\bibinfo{year}{2019}), \bibinfo{pages}{127--298}.
\newblock


\bibitem[\protect\citeauthoryear{Gao, Lee, Zhang, Brockett, Galley, Gao, and
  Dolan}{Gao et~al\mbox{.}}{2019b}]%
        {gao2019jointly}
\bibfield{author}{\bibinfo{person}{Xiang Gao}, \bibinfo{person}{Sungjin Lee},
  \bibinfo{person}{Yizhe Zhang}, \bibinfo{person}{Chris Brockett},
  \bibinfo{person}{Michel Galley}, \bibinfo{person}{Jianfeng Gao}, {and}
  \bibinfo{person}{Bill Dolan}.} \bibinfo{year}{2019}\natexlab{b}.
\newblock \showarticletitle{Jointly Optimizing Diversity and Relevance in
  Neural Response Generation}.
\newblock \bibinfo{journal}{\emph{arXiv preprint arXiv:1902.11205}}
  (\bibinfo{year}{2019}).
\newblock


\bibitem[\protect\citeauthoryear{Ghazvininejad, Brockett, Chang, Dolan, Gao,
  Yih, and Galley}{Ghazvininejad et~al\mbox{.}}{2018}]%
        {ghazvininejad2017knowledge}
\bibfield{author}{\bibinfo{person}{Marjan Ghazvininejad},
  \bibinfo{person}{Chris Brockett}, \bibinfo{person}{Ming{-}Wei Chang},
  \bibinfo{person}{Bill Dolan}, \bibinfo{person}{Jianfeng Gao},
  \bibinfo{person}{Wen{-}tau Yih}, {and} \bibinfo{person}{Michel Galley}.}
  \bibinfo{year}{2018}\natexlab{}.
\newblock \showarticletitle{A Knowledge-Grounded Neural Conversation Model}. In
  \bibinfo{booktitle}{\emph{Proceedings of {AAAI} 2018, New Orleans, Louisiana,
  USA, February 2-7, 2018}}. \bibinfo{pages}{5110--5117}.
\newblock


\bibitem[\protect\citeauthoryear{Ghosh, Chollet, Laksana, Morency, and
  Scherer}{Ghosh et~al\mbox{.}}{2017}]%
        {ghosh2017affect}
\bibfield{author}{\bibinfo{person}{Sayan Ghosh}, \bibinfo{person}{Mathieu
  Chollet}, \bibinfo{person}{Eugene Laksana}, \bibinfo{person}{Louis{-}Philippe
  Morency}, {and} \bibinfo{person}{Stefan Scherer}.}
  \bibinfo{year}{2017}\natexlab{}.
\newblock \showarticletitle{Affect-LM: {A} Neural Language Model for
  Customizable Affective Text Generation}. In
  \bibinfo{booktitle}{\emph{Proceedings of {ACL} 2017, Vancouver, Canada, July
  30-August 4, 2017}}. \bibinfo{pages}{634--642}.
\newblock


\bibitem[\protect\citeauthoryear{Gim\'{e}nez and M\`{a}rquez}{Gim\'{e}nez and
  M\`{a}rquez}{2008}]%
        {gimenez08smor}
\bibfield{author}{\bibinfo{person}{Jes\'{u}s Gim\'{e}nez} {and}
  \bibinfo{person}{Llu\'{i}s M\`{a}rquez}.} \bibinfo{year}{2008}\natexlab{}.
\newblock \showarticletitle{A Smorgasbord of Features for Automatic {MT}
  Evaluation}. In \bibinfo{booktitle}{\emph{Proceedings of the Third Workshop
  on Statistical Machine Translation}}. \bibinfo{pages}{195--198}.
\newblock


\bibitem[\protect\citeauthoryear{Gosling, Rentfrow, and Swann}{Gosling
  et~al\mbox{.}}{2003}]%
        {gosling2003very}
\bibfield{author}{\bibinfo{person}{Samuel~D Gosling}, \bibinfo{person}{Peter~J
  Rentfrow}, {and} \bibinfo{person}{William~B Swann}.}
  \bibinfo{year}{2003}\natexlab{}.
\newblock \showarticletitle{A very brief measure of the Big-Five personality
  domains}.
\newblock \bibinfo{journal}{\emph{Journal of Research in personality}}
  \bibinfo{volume}{37}, \bibinfo{number}{6} (\bibinfo{year}{2003}),
  \bibinfo{pages}{504--528}.
\newblock


\bibitem[\protect\citeauthoryear{Gu, Lu, Li, and Li}{Gu et~al\mbox{.}}{2016}]%
        {gu2016copy}
\bibfield{author}{\bibinfo{person}{Jiatao Gu}, \bibinfo{person}{Zhengdong Lu},
  \bibinfo{person}{Hang Li}, {and} \bibinfo{person}{Victor O.~K. Li}.}
  \bibinfo{year}{2016}\natexlab{}.
\newblock \showarticletitle{Incorporating Copying Mechanism in
  Sequence-to-Sequence Learning}. In \bibinfo{booktitle}{\emph{Proceedings of
  {ACL} 2016, Berlin, Germany, August 7-12, 2016}}.
\newblock


\bibitem[\protect\citeauthoryear{Han, Bang, Ryu, and Lee}{Han
  et~al\mbox{.}}{2015}]%
        {han2015exploiting}
\bibfield{author}{\bibinfo{person}{Sangdo Han}, \bibinfo{person}{Jeesoo Bang},
  \bibinfo{person}{Seonghan Ryu}, {and} \bibinfo{person}{Gary~Geunbae Lee}.}
  \bibinfo{year}{2015}\natexlab{}.
\newblock \showarticletitle{Exploiting knowledge base to generate responses for
  natural language dialog listening agents}. In
  \bibinfo{booktitle}{\emph{Proceedings of {SIGDIAL} 2015, September 2-4, 2015,
  Prague, Czech Republic}}. \bibinfo{pages}{129--133}.
\newblock


\bibitem[\protect\citeauthoryear{Hashimoto, Zhang, and Liang}{Hashimoto
  et~al\mbox{.}}{2019}]%
        {hashimoto2019unifying}
\bibfield{author}{\bibinfo{person}{Tatsunori Hashimoto}, \bibinfo{person}{Hugh
  Zhang}, {and} \bibinfo{person}{Percy Liang}.}
  \bibinfo{year}{2019}\natexlab{}.
\newblock \showarticletitle{Unifying Human and Statistical Evaluation for
  Natural Language Generation}. In \bibinfo{booktitle}{\emph{{NAACL} 2019,
  Minneapolis, USA, June 2-7, 2019}}. \bibinfo{pages}{1689--1701}.
\newblock


\bibitem[\protect\citeauthoryear{Henderson, Thomson, and Young}{Henderson
  et~al\mbox{.}}{2013}]%
        {henderson2013deep}
\bibfield{author}{\bibinfo{person}{Matthew Henderson}, \bibinfo{person}{Blaise
  Thomson}, {and} \bibinfo{person}{Steve~J. Young}.}
  \bibinfo{year}{2013}\natexlab{}.
\newblock \showarticletitle{Deep Neural Network Approach for the Dialog State
  Tracking Challenge}. In \bibinfo{booktitle}{\emph{Proceedings of {SIGDIAL}
  2013, August 22-24, 2013, SUPELEC, Metz, France}}. \bibinfo{pages}{467--471}.
\newblock


\bibitem[\protect\citeauthoryear{Higashinaka, Imamura, Meguro, Miyazaki,
  Kobayashi, Sugiyama, Hirano, Makino, and Matsuo}{Higashinaka
  et~al\mbox{.}}{2014}]%
        {a32}
\bibfield{author}{\bibinfo{person}{Ryuichiro Higashinaka},
  \bibinfo{person}{Kenji Imamura}, \bibinfo{person}{Toyomi Meguro},
  \bibinfo{person}{Chiaki Miyazaki}, \bibinfo{person}{Nozomi Kobayashi},
  \bibinfo{person}{Hiroaki Sugiyama}, \bibinfo{person}{Toru Hirano},
  \bibinfo{person}{Toshiro Makino}, {and} \bibinfo{person}{Yoshihiro Matsuo}.}
  \bibinfo{year}{2014}\natexlab{}.
\newblock \showarticletitle{Towards an open-domain conversational system fully
  based on natural language processing}. In \bibinfo{booktitle}{\emph{{COLING}
  2014, August 23-29, 2014, Dublin, Ireland}}. \bibinfo{pages}{928--939}.
\newblock


\bibitem[\protect\citeauthoryear{Hu, Lu, Li, and Chen}{Hu
  et~al\mbox{.}}{2014}]%
        {arc}
\bibfield{author}{\bibinfo{person}{Baotian Hu}, \bibinfo{person}{Zhengdong Lu},
  \bibinfo{person}{Hang Li}, {and} \bibinfo{person}{Qingcai Chen}.}
  \bibinfo{year}{2014}\natexlab{}.
\newblock \showarticletitle{Convolutional Neural Network Architectures for
  Matching Natural Language Sentences}. In \bibinfo{booktitle}{\emph{{NIPS}
  2014, Montreal, Quebec, Canada, December 8-13, 2014}}.
  \bibinfo{pages}{2042--2050}.
\newblock


\bibitem[\protect\citeauthoryear{Hu, Yang, Liang, Salakhutdinov, and Xing}{Hu
  et~al\mbox{.}}{2017}]%
        {hu2017controllable}
\bibfield{author}{\bibinfo{person}{Zhiting Hu}, \bibinfo{person}{Zichao Yang},
  \bibinfo{person}{Xiaodan Liang}, \bibinfo{person}{Ruslan Salakhutdinov},
  {and} \bibinfo{person}{Eric~P. Xing}.} \bibinfo{year}{2017}\natexlab{}.
\newblock \showarticletitle{Toward Controlled Generation of Text}. In
  \bibinfo{booktitle}{\emph{Proceedings of {ICML} 2017, Sydney, NSW, Australia,
  6-11 August 2017}}, Vol.~\bibinfo{volume}{70}. \bibinfo{pages}{1587--1596}.
\newblock


\bibitem[\protect\citeauthoryear{Huang, He, Gao, Deng, Acero, and Heck}{Huang
  et~al\mbox{.}}{2013}]%
        {Huang:2013fz}
\bibfield{author}{\bibinfo{person}{Po{-}Sen Huang}, \bibinfo{person}{Xiaodong
  He}, \bibinfo{person}{Jianfeng Gao}, \bibinfo{person}{Li Deng},
  \bibinfo{person}{Alex Acero}, {and} \bibinfo{person}{Larry~P. Heck}.}
  \bibinfo{year}{2013}\natexlab{}.
\newblock \showarticletitle{Learning deep structured semantic models for web
  search using clickthrough data}. In \bibinfo{booktitle}{\emph{22nd {ACM}
  International Conference on Information and Knowledge Management, CIKM'13,
  San Francisco, CA, USA, October 27 - November 1, 2013}}.
  \bibinfo{pages}{2333--2338}.
\newblock


\bibitem[\protect\citeauthoryear{Huber, McDuff, Brockett, Galley, and
  Dolan}{Huber et~al\mbox{.}}{2018}]%
        {huber2018emotional}
\bibfield{author}{\bibinfo{person}{Bernd Huber}, \bibinfo{person}{Daniel
  McDuff}, \bibinfo{person}{Chris Brockett}, \bibinfo{person}{Michel Galley},
  {and} \bibinfo{person}{Bill Dolan}.} \bibinfo{year}{2018}\natexlab{}.
\newblock \showarticletitle{Emotional Dialogue Generation using Image-Grounded
  Language Models}. In \bibinfo{booktitle}{\emph{Proceedings of the 2018 {CHI}
  Conference on Human Factors in Computing Systems, {CHI} 2018, Montreal, QC,
  Canada, April 21-26, 2018}}. \bibinfo{pages}{277}.
\newblock


\bibitem[\protect\citeauthoryear{Joshi, Mi, and Faltings}{Joshi
  et~al\mbox{.}}{2017}]%
        {joshi2017personalization-goaldialog}
\bibfield{author}{\bibinfo{person}{Chaitanya~K. Joshi}, \bibinfo{person}{Fei
  Mi}, {and} \bibinfo{person}{Boi Faltings}.} \bibinfo{year}{2017}\natexlab{}.
\newblock \showarticletitle{Personalization in Goal-Oriented Dialog}.
\newblock \bibinfo{journal}{\emph{CoRR}}  \bibinfo{volume}{abs/1706.07503}
  (\bibinfo{year}{2017}).
\newblock


\bibitem[\protect\citeauthoryear{Kaiser, Bengio, Roy, Vaswani, Parmar,
  Uszkoreit, and Shazeer}{Kaiser et~al\mbox{.}}{2018}]%
        {kaiser2018lt}
\bibfield{author}{\bibinfo{person}{Lukasz Kaiser}, \bibinfo{person}{Samy
  Bengio}, \bibinfo{person}{Aurko Roy}, \bibinfo{person}{Ashish Vaswani},
  \bibinfo{person}{Niki Parmar}, \bibinfo{person}{Jakob Uszkoreit}, {and}
  \bibinfo{person}{Noam Shazeer}.} \bibinfo{year}{2018}\natexlab{}.
\newblock \showarticletitle{Fast Decoding in Sequence Models Using Discrete
  Latent Variables}. In \bibinfo{booktitle}{\emph{Proceedings of {ICML} 2018,
  Stockholmsm{\"{a}}ssan, Stockholm, Sweden, July 10-15, 2018}},
  Vol.~\bibinfo{volume}{80}. \bibinfo{pages}{2395--2404}.
\newblock


\bibitem[\protect\citeauthoryear{Ke, Guan, Huang, and Zhu}{Ke
  et~al\mbox{.}}{2018}]%
        {ke2018senfunc}
\bibfield{author}{\bibinfo{person}{Pei Ke}, \bibinfo{person}{Jian Guan},
  \bibinfo{person}{Minlie Huang}, {and} \bibinfo{person}{Xiaoyan Zhu}.}
  \bibinfo{year}{2018}\natexlab{}.
\newblock \showarticletitle{Generating Informative Responses with Controlled
  Sentence Function}. In \bibinfo{booktitle}{\emph{Proceedings of {ACL} 2018,
  Melbourne, Australia, July 15-20, 2018}}. \bibinfo{pages}{1499--1508}.
\newblock


\bibitem[\protect\citeauthoryear{Kim, Bang, Choi, Ryu, Koo, and Lee}{Kim
  et~al\mbox{.}}{2014}]%
        {Kim2014Acquisition}
\bibfield{author}{\bibinfo{person}{Yonghee Kim}, \bibinfo{person}{Jeesoo Bang},
  \bibinfo{person}{Junhwi Choi}, \bibinfo{person}{Seonghan Ryu},
  \bibinfo{person}{Sangjun Koo}, {and} \bibinfo{person}{Gary~Geunbae Lee}.}
  \bibinfo{year}{2014}\natexlab{}.
\newblock \showarticletitle{Acquisition and Use of Long-Term Memory for
  Personalized Dialog Systems}. In \bibinfo{booktitle}{\emph{Multimodal
  Analyses enabling Artificial Agents in Human-Machine Interaction - Second
  International Workshop, {MA3HMI} 2014, Held in Conjunction with {INTERSPEECH}
  2014, Singapore, Singapore, September 14, 2014}},
  Vol.~\bibinfo{volume}{8757}. \bibinfo{pages}{78--87}.
\newblock


\bibitem[\protect\citeauthoryear{Kulesza and Shieber}{Kulesza and
  Shieber}{2004}]%
        {kulesza04learning}
\bibfield{author}{\bibinfo{person}{Alex Kulesza} {and}
  \bibinfo{person}{Stuart~M. Shieber}.} \bibinfo{year}{2004}\natexlab{}.
\newblock \showarticletitle{A Learning Approach to Improving Sentence-Level
  {MT} Evaluation}. In \bibinfo{booktitle}{\emph{Proceedings of the 10th
  International Conference on Theoretical and Methodological Issues in Machine
  Translation}}. \bibinfo{address}{Baltimore, MD}.
\newblock


\bibitem[\protect\citeauthoryear{Lee, Mansimov, and Cho}{Lee
  et~al\mbox{.}}{2018}]%
        {lee2018refine}
\bibfield{author}{\bibinfo{person}{Jason Lee}, \bibinfo{person}{Elman
  Mansimov}, {and} \bibinfo{person}{Kyunghyun Cho}.}
  \bibinfo{year}{2018}\natexlab{}.
\newblock \showarticletitle{Deterministic Non-Autoregressive Neural Sequence
  Modeling by Iterative Refinement}. In \bibinfo{booktitle}{\emph{Proceedings
  of {EMNLP} 2018, Brussels, Belgium, October 31 - November 4, 2018}}.
  \bibinfo{pages}{1173--1182}.
\newblock


\bibitem[\protect\citeauthoryear{Li, Galley, Brockett, Gao, and Dolan}{Li
  et~al\mbox{.}}{2016a}]%
        {lidiversity}
\bibfield{author}{\bibinfo{person}{Jiwei Li}, \bibinfo{person}{Michel Galley},
  \bibinfo{person}{Chris Brockett}, \bibinfo{person}{Jianfeng Gao}, {and}
  \bibinfo{person}{Bill Dolan}.} \bibinfo{year}{2016}\natexlab{a}.
\newblock \showarticletitle{A Diversity-Promoting Objective Function for Neural
  Conversation Models}. In \bibinfo{booktitle}{\emph{{NAACL} {HLT} 2016, San
  Diego California, USA, June 12-17, 2016}}. \bibinfo{pages}{110--119}.
\newblock


\bibitem[\protect\citeauthoryear{Li, Galley, Brockett, Spithourakis, Gao, and
  Dolan}{Li et~al\mbox{.}}{2016b}]%
        {Li2016_ACL-persona}
\bibfield{author}{\bibinfo{person}{Jiwei Li}, \bibinfo{person}{Michel Galley},
  \bibinfo{person}{Chris Brockett}, \bibinfo{person}{Georgios~P. Spithourakis},
  \bibinfo{person}{Jianfeng Gao}, {and} \bibinfo{person}{William~B. Dolan}.}
  \bibinfo{year}{2016}\natexlab{b}.
\newblock \showarticletitle{A Persona-Based Neural Conversation Model}. In
  \bibinfo{booktitle}{\emph{Proceedings of {ACL} 2016, Berlin, Germany, August
  7-12, 2016}}.
\newblock


\bibitem[\protect\citeauthoryear{Li, Monroe, and Jurafsky}{Li
  et~al\mbox{.}}{2016c}]%
        {li2016simple}
\bibfield{author}{\bibinfo{person}{Jiwei Li}, \bibinfo{person}{Will Monroe},
  {and} \bibinfo{person}{Dan Jurafsky}.} \bibinfo{year}{2016}\natexlab{c}.
\newblock \showarticletitle{A Simple, Fast Diverse Decoding Algorithm for
  Neural Generation}.
\newblock \bibinfo{journal}{\emph{CoRR}}  \bibinfo{volume}{abs/1611.08562}
  (\bibinfo{year}{2016}).
\newblock


\bibitem[\protect\citeauthoryear{Li, Monroe, Shi, Jean, Ritter, and
  Jurafsky}{Li et~al\mbox{.}}{2017}]%
        {li2017adversarial}
\bibfield{author}{\bibinfo{person}{Jiwei Li}, \bibinfo{person}{Will Monroe},
  \bibinfo{person}{Tianlin Shi}, \bibinfo{person}{S{\'{e}}bastien Jean},
  \bibinfo{person}{Alan Ritter}, {and} \bibinfo{person}{Dan Jurafsky}.}
  \bibinfo{year}{2017}\natexlab{}.
\newblock \showarticletitle{Adversarial Learning for Neural Dialogue
  Generation}. In \bibinfo{booktitle}{\emph{Proceedings of {EMNLP} 2017,
  Copenhagen, Denmark, September 9-11, 2017}}. \bibinfo{pages}{2157--2169}.
\newblock


\bibitem[\protect\citeauthoryear{Li, Mou, Yan, and Zhang}{Li
  et~al\mbox{.}}{2016d}]%
        {li2016stalematebreakerIRbased}
\bibfield{author}{\bibinfo{person}{Xiang Li}, \bibinfo{person}{Lili Mou},
  \bibinfo{person}{Rui Yan}, {and} \bibinfo{person}{Ming Zhang}.}
  \bibinfo{year}{2016}\natexlab{d}.
\newblock \showarticletitle{StalemateBreaker: {A} Proactive Content-Introducing
  Approach to Automatic Human-Computer Conversation}. In
  \bibinfo{booktitle}{\emph{Proceedings of {IJCAI} 2016, New York, NY, USA,
  9-15 July 2016}}. \bibinfo{pages}{2845--2851}.
\newblock


\bibitem[\protect\citeauthoryear{Liao, Ma, He, Hong, and Chua}{Liao
  et~al\mbox{.}}{2018}]%
        {liao2018knowledge}
\bibfield{author}{\bibinfo{person}{Lizi Liao}, \bibinfo{person}{Yunshan Ma},
  \bibinfo{person}{Xiangnan He}, \bibinfo{person}{Richang Hong}, {and}
  \bibinfo{person}{Tat-seng Chua}.} \bibinfo{year}{2018}\natexlab{}.
\newblock \showarticletitle{Knowledge-aware Multimodal Dialogue Systems}. In
  \bibinfo{booktitle}{\emph{2018 ACM Multimedia Conference on Multimedia
  Conference}}. \bibinfo{pages}{801--809}.
\newblock


\bibitem[\protect\citeauthoryear{Lin}{Lin}{2004}]%
        {lin2004rouge}
\bibfield{author}{\bibinfo{person}{Chin-Yew Lin}.}
  \bibinfo{year}{2004}\natexlab{}.
\newblock \showarticletitle{Rouge: A package for automatic evaluation of
  summaries}.
\newblock \bibinfo{journal}{\emph{Text Summarization Branches Out}}
  (\bibinfo{year}{2004}).
\newblock


\bibitem[\protect\citeauthoryear{Lipton, Li, Gao, Li, Ahmed, and Deng}{Lipton
  et~al\mbox{.}}{2018}]%
        {lipton2018bbq}
\bibfield{author}{\bibinfo{person}{Zachary~C. Lipton}, \bibinfo{person}{Xiujun
  Li}, \bibinfo{person}{Jianfeng Gao}, \bibinfo{person}{Lihong Li},
  \bibinfo{person}{Faisal Ahmed}, {and} \bibinfo{person}{Li Deng}.}
  \bibinfo{year}{2018}\natexlab{}.
\newblock \showarticletitle{BBQ-Networks: Efficient Exploration in Deep
  Reinforcement Learning for Task-Oriented Dialogue Systems}. In
  \bibinfo{booktitle}{\emph{Proceedings of {AAAI} 2018, New Orleans, Louisiana,
  USA, February 2-7, 2018}}. \bibinfo{pages}{5237--5244}.
\newblock


\bibitem[\protect\citeauthoryear{Lita, Rogati, and Lavie}{Lita
  et~al\mbox{.}}{2005}]%
        {lita05blanc}
\bibfield{author}{\bibinfo{person}{Lucian~Vlad Lita}, \bibinfo{person}{Monica
  Rogati}, {and} \bibinfo{person}{Alon Lavie}.}
  \bibinfo{year}{2005}\natexlab{}.
\newblock \showarticletitle{{BLANC}: Learning Evaluation Metrics for {MT}}. In
  \bibinfo{booktitle}{\emph{Proceedings of the Conference on Human Language
  Technology and Empirical Methods in Natural Language Processing}} (Vancouver,
  British Columbia, Canada) \emph{(\bibinfo{series}{HLT '05})}.
  \bibinfo{pages}{740--747}.
\newblock


\bibitem[\protect\citeauthoryear{Liu, Lowe, Serban, Noseworthy, Charlin, and
  Pineau}{Liu et~al\mbox{.}}{2016}]%
        {liu2016not}
\bibfield{author}{\bibinfo{person}{Chia{-}Wei Liu}, \bibinfo{person}{Ryan
  Lowe}, \bibinfo{person}{Iulian Serban}, \bibinfo{person}{Michael Noseworthy},
  \bibinfo{person}{Laurent Charlin}, {and} \bibinfo{person}{Joelle Pineau}.}
  \bibinfo{year}{2016}\natexlab{}.
\newblock \showarticletitle{How {NOT} To Evaluate Your Dialogue System: An
  Empirical Study of Unsupervised Evaluation Metrics for Dialogue Response
  Generation}. In \bibinfo{booktitle}{\emph{Proceedings of {EMNLP} 2016,
  Austin, Texas, USA, November 1-4, 2016}}. \bibinfo{pages}{2122--2132}.
\newblock


\bibitem[\protect\citeauthoryear{Liu, Chen, Ren, Feng, Liu, and Yin}{Liu
  et~al\mbox{.}}{2018}]%
        {liu2018knowledge}
\bibfield{author}{\bibinfo{person}{Shuman Liu}, \bibinfo{person}{Hongshen
  Chen}, \bibinfo{person}{Zhaochun Ren}, \bibinfo{person}{Yang Feng},
  \bibinfo{person}{Qun Liu}, {and} \bibinfo{person}{Dawei Yin}.}
  \bibinfo{year}{2018}\natexlab{}.
\newblock \showarticletitle{Knowledge Diffusion for Neural Dialogue
  Generation}. In \bibinfo{booktitle}{\emph{Proceedings of {ACL} 2018,
  Melbourne, Australia, July 15-20, 2018}}. \bibinfo{pages}{1489--1498}.
\newblock


\bibitem[\protect\citeauthoryear{Liu}{Liu}{2010}]%
        {liutieyan2009l2r}
\bibfield{author}{\bibinfo{person}{Tie{-}Yan Liu}.}
  \bibinfo{year}{2010}\natexlab{}.
\newblock \showarticletitle{Learning to rank for information retrieval}. In
  \bibinfo{booktitle}{\emph{Proceeding of {SIGIR} 2010, Geneva, Switzerland,
  July 19-23, 2010}}. \bibinfo{pages}{904}.
\newblock


\bibitem[\protect\citeauthoryear{Lowe, Noseworthy, Serban, Angelard{-}Gontier,
  Bengio, and Pineau}{Lowe et~al\mbox{.}}{2017}]%
        {lowe2017towards}
\bibfield{author}{\bibinfo{person}{Ryan Lowe}, \bibinfo{person}{Michael
  Noseworthy}, \bibinfo{person}{Iulian~Vlad Serban}, \bibinfo{person}{Nicolas
  Angelard{-}Gontier}, \bibinfo{person}{Yoshua Bengio}, {and}
  \bibinfo{person}{Joelle Pineau}.} \bibinfo{year}{2017}\natexlab{}.
\newblock \showarticletitle{Towards an Automatic Turing Test: Learning to
  Evaluate Dialogue Responses}. In \bibinfo{booktitle}{\emph{Proceedings of
  {ACL} 2017, Vancouver, Canada, July 30-August 4, 2017}}.
  \bibinfo{pages}{1116--1126}.
\newblock


\bibitem[\protect\citeauthoryear{Lu and Li}{Lu and Li}{2013}]%
        {lu2013deep}
\bibfield{author}{\bibinfo{person}{Zhengdong Lu} {and} \bibinfo{person}{Hang
  Li}.} \bibinfo{year}{2013}\natexlab{}.
\newblock \showarticletitle{A Deep Architecture for Matching Short Texts}. In
  \bibinfo{booktitle}{\emph{{NIPS} 2013, December 5-8, 2013, Lake Tahoe,
  Nevada, United States.}} \bibinfo{pages}{1367--1375}.
\newblock


\bibitem[\protect\citeauthoryear{Luan, Brockett, Dolan, Gao, and Galley}{Luan
  et~al\mbox{.}}{2017}]%
        {ar1}
\bibfield{author}{\bibinfo{person}{Yi Luan}, \bibinfo{person}{Chris Brockett},
  \bibinfo{person}{Bill Dolan}, \bibinfo{person}{Jianfeng Gao}, {and}
  \bibinfo{person}{Michel Galley}.} \bibinfo{year}{2017}\natexlab{}.
\newblock \showarticletitle{Multi-Task Learning for Speaker-Role Adaptation in
  Neural Conversation Models}. In \bibinfo{booktitle}{\emph{Proceedings of the
  Eighth International Joint Conference on Natural Language Processing,
  {IJCNLP} 2017, Taipei, Taiwan, November 27 - December 1, 2017 - Volume 1:
  Long Papers}}. \bibinfo{pages}{605--614}.
\newblock


\bibitem[\protect\citeauthoryear{Manning, Raghavan, and Sch{\"{u}}tze}{Manning
  et~al\mbox{.}}{2008}]%
        {manning2008introduction}
\bibfield{author}{\bibinfo{person}{Christopher~D. Manning},
  \bibinfo{person}{Prabhakar Raghavan}, {and} \bibinfo{person}{Hinrich
  Sch{\"{u}}tze}.} \bibinfo{year}{2008}\natexlab{}.
\newblock \bibinfo{booktitle}{\emph{Introduction to information retrieval}}.
\newblock \bibinfo{publisher}{Cambridge University Press}.
\newblock
\showISBNx{978-0-521-86571-5}


\bibitem[\protect\citeauthoryear{Mazar{\'{e}}, Humeau, Raison, and
  Bordes}{Mazar{\'{e}} et~al\mbox{.}}{2018}]%
        {mazare2018millionsagent}
\bibfield{author}{\bibinfo{person}{Pierre{-}Emmanuel Mazar{\'{e}}},
  \bibinfo{person}{Samuel Humeau}, \bibinfo{person}{Martin Raison}, {and}
  \bibinfo{person}{Antoine Bordes}.} \bibinfo{year}{2018}\natexlab{}.
\newblock \showarticletitle{Training Millions of Personalized Dialogue Agents}.
  In \bibinfo{booktitle}{\emph{Proceedings of {EMNLP} 2018, Brussels, Belgium,
  October 31 - November 4, 2018}}. \bibinfo{pages}{2775--2779}.
\newblock


\bibitem[\protect\citeauthoryear{Mo, Zhang, Li, Li, and Yang}{Mo
  et~al\mbox{.}}{2018}]%
        {mo2016personalizing}
\bibfield{author}{\bibinfo{person}{Kaixiang Mo}, \bibinfo{person}{Yu Zhang},
  \bibinfo{person}{Shuangyin Li}, \bibinfo{person}{Jiajun Li}, {and}
  \bibinfo{person}{Qiang Yang}.} \bibinfo{year}{2018}\natexlab{}.
\newblock \showarticletitle{Personalizing a Dialogue System With Transfer
  Reinforcement Learning}. In \bibinfo{booktitle}{\emph{Proceedings of {AAAI}
  2018, New Orleans, Louisiana, USA, February 2-7, 2018}}.
  \bibinfo{pages}{5317--5324}.
\newblock


\bibitem[\protect\citeauthoryear{Mostafazadeh, Brockett, Dolan, Galley, Gao,
  Spithourakis, and Vanderwende}{Mostafazadeh et~al\mbox{.}}{2017}]%
        {mostafazadeh2017image}
\bibfield{author}{\bibinfo{person}{Nasrin Mostafazadeh}, \bibinfo{person}{Chris
  Brockett}, \bibinfo{person}{Bill Dolan}, \bibinfo{person}{Michel Galley},
  \bibinfo{person}{Jianfeng Gao}, \bibinfo{person}{Georgios Spithourakis},
  {and} \bibinfo{person}{Lucy Vanderwende}.} \bibinfo{year}{2017}\natexlab{}.
\newblock \showarticletitle{Image-Grounded Conversations: Multimodal Context
  for Natural Question and Response Generation}. In
  \bibinfo{booktitle}{\emph{IJCNLP}}. \bibinfo{pages}{462--472}.
\newblock


\bibitem[\protect\citeauthoryear{Mrksic, S{\'{e}}aghdha, Wen, Thomson, and
  Young}{Mrksic et~al\mbox{.}}{2017}]%
        {Mrk2017ACL}
\bibfield{author}{\bibinfo{person}{Nikola Mrksic},
  \bibinfo{person}{Diarmuid~{\'{O}} S{\'{e}}aghdha},
  \bibinfo{person}{Tsung{-}Hsien Wen}, \bibinfo{person}{Blaise Thomson}, {and}
  \bibinfo{person}{Steve~J. Young}.} \bibinfo{year}{2017}\natexlab{}.
\newblock \showarticletitle{Neural Belief Tracker: Data-Driven Dialogue State
  Tracking}. In \bibinfo{booktitle}{\emph{Proceedings of {ACL} 2017, Vancouver,
  Canada, July 30-August 4, 2017}}. \bibinfo{pages}{1777--1788}.
\newblock


\bibitem[\protect\citeauthoryear{Norman}{Norman}{1963}]%
        {norman1963toward}
\bibfield{author}{\bibinfo{person}{Warren~T Norman}.}
  \bibinfo{year}{1963}\natexlab{}.
\newblock \showarticletitle{Toward an adequate taxonomy of personality
  attributes: Replicated factor structure in peer nomination personality
  ratings.}
\newblock \bibinfo{journal}{\emph{The Journal of Abnormal and Social
  Psychology}} \bibinfo{volume}{66}, \bibinfo{number}{6}
  (\bibinfo{year}{1963}), \bibinfo{pages}{574}.
\newblock


\bibitem[\protect\citeauthoryear{Nothdurft, Ultes, and Minker}{Nothdurft
  et~al\mbox{.}}{2015}]%
        {openquest15}
\bibfield{author}{\bibinfo{person}{Florian Nothdurft}, \bibinfo{person}{Stefan
  Ultes}, {and} \bibinfo{person}{Wolfgang Minker}.}
  \bibinfo{year}{2015}\natexlab{}.
\newblock \showarticletitle{Finding appropriate interaction strategies for
  proactive dialogue systems-an open quest}. In
  \bibinfo{booktitle}{\emph{Proceedings of the 2nd European and the 5th Nordic
  Symposium on Multimodal Communication, August 6-8, 2014, Tartu, Estonia}}.
  \bibinfo{pages}{73--80}.
\newblock


\bibitem[\protect\citeauthoryear{Novikova, Du{\v{s}}ek, Curry, and
  Rieser}{Novikova et~al\mbox{.}}{2017}]%
        {novikova2017we}
\bibfield{author}{\bibinfo{person}{Jekaterina Novikova},
  \bibinfo{person}{Ond{\v{r}}ej Du{\v{s}}ek}, \bibinfo{person}{Amanda~Cercas
  Curry}, {and} \bibinfo{person}{Verena Rieser}.}
  \bibinfo{year}{2017}\natexlab{}.
\newblock \showarticletitle{Why We Need New Evaluation Metrics for NLG}. In
  \bibinfo{booktitle}{\emph{EMNLP}}. \bibinfo{pages}{2241--2252}.
\newblock


\bibitem[\protect\citeauthoryear{Oraby, Reed, Tandon, S., Lukin, and
  Walker}{Oraby et~al\mbox{.}}{2018}]%
        {oraby2018controlling}
\bibfield{author}{\bibinfo{person}{Shereen Oraby}, \bibinfo{person}{Lena Reed},
  \bibinfo{person}{Shubhangi Tandon}, \bibinfo{person}{Sharath~T. S.},
  \bibinfo{person}{Stephanie~M. Lukin}, {and} \bibinfo{person}{Marilyn~A.
  Walker}.} \bibinfo{year}{2018}\natexlab{}.
\newblock \showarticletitle{Controlling Personality-Based Stylistic Variation
  with Neural Natural Language Generators}. In
  \bibinfo{booktitle}{\emph{Proceedings of {SIGDIAL} 2018, July 12-14, 2018,
  Melbourne, Australia}}. \bibinfo{pages}{180--190}.
\newblock


\bibitem[\protect\citeauthoryear{Ouchi and Tsuboi}{Ouchi and Tsuboi}{2016}]%
        {ouchi2016addressee}
\bibfield{author}{\bibinfo{person}{Hiroki Ouchi} {and} \bibinfo{person}{Yuta
  Tsuboi}.} \bibinfo{year}{2016}\natexlab{}.
\newblock \showarticletitle{Addressee and Response Selection for Multi-Party
  Conversation}. In \bibinfo{booktitle}{\emph{Proceedings of {EMNLP} 2016,
  Austin, Texas, USA, November 1-4, 2016}}. \bibinfo{pages}{2133--2143}.
\newblock


\bibitem[\protect\citeauthoryear{Pado, Cer, Galley, Jurafsky, and Manning}{Pado
  et~al\mbox{.}}{2009}]%
        {pado09measuring}
\bibfield{author}{\bibinfo{person}{Sebastian Pado}, \bibinfo{person}{Daniel
  Cer}, \bibinfo{person}{Michel Galley}, \bibinfo{person}{Dan Jurafsky}, {and}
  \bibinfo{person}{Christopher~D. Manning}.} \bibinfo{year}{2009}\natexlab{}.
\newblock \showarticletitle{Measuring Machine Translation Quality as Semantic
  Equivalence: A Metric Based on Entailment Features}.
\newblock \bibinfo{journal}{\emph{Machine Translation}} (\bibinfo{year}{2009}),
  \bibinfo{pages}{181--193}.
\newblock


\bibitem[\protect\citeauthoryear{Palangi, Deng, Shen, Gao, He, Chen, Song, and
  Ward}{Palangi et~al\mbox{.}}{2016}]%
        {palangi2015deep}
\bibfield{author}{\bibinfo{person}{Hamid Palangi}, \bibinfo{person}{Li Deng},
  \bibinfo{person}{Yelong Shen}, \bibinfo{person}{Jianfeng Gao},
  \bibinfo{person}{Xiaodong He}, \bibinfo{person}{Jianshu Chen},
  \bibinfo{person}{Xinying Song}, {and} \bibinfo{person}{Rabab~K. Ward}.}
  \bibinfo{year}{2016}\natexlab{}.
\newblock \showarticletitle{Deep Sentence Embedding Using Long Short-Term
  Memory Networks: Analysis and Application to Information Retrieval}.
\newblock \bibinfo{journal}{\emph{{IEEE/ACM} Trans. Audio, Speech {\&} Language
  Processing}} \bibinfo{volume}{24}, \bibinfo{number}{4}
  (\bibinfo{year}{2016}), \bibinfo{pages}{694--707}.
\newblock


\bibitem[\protect\citeauthoryear{Pandey, Contractor, Kumar, and Joshi}{Pandey
  et~al\mbox{.}}{2018}]%
        {Contractor:2018vl}
\bibfield{author}{\bibinfo{person}{Gaurav Pandey}, \bibinfo{person}{Danish
  Contractor}, \bibinfo{person}{Vineet Kumar}, {and} \bibinfo{person}{Sachindra
  Joshi}.} \bibinfo{year}{2018}\natexlab{}.
\newblock \showarticletitle{Exemplar Encoder-Decoder for Neural Conversation
  Generation}. In \bibinfo{booktitle}{\emph{Proceedings of {ACL} 2018,
  Melbourne, Australia, July 15-20, 2018}}. \bibinfo{pages}{1329--1338}.
\newblock


\bibitem[\protect\citeauthoryear{Pang, Lan, Guo, Xu, Wan, and Cheng}{Pang
  et~al\mbox{.}}{2016}]%
        {pang2016pyramid}
\bibfield{author}{\bibinfo{person}{Liang Pang}, \bibinfo{person}{Yanyan Lan},
  \bibinfo{person}{Jiafeng Guo}, \bibinfo{person}{Jun Xu},
  \bibinfo{person}{Shengxian Wan}, {and} \bibinfo{person}{Xueqi Cheng}.}
  \bibinfo{year}{2016}\natexlab{}.
\newblock \showarticletitle{Text Matching as Image Recognition}. In
  \bibinfo{booktitle}{\emph{Proceedings of {AAAI} 2016, February 12-17, 2016,
  Phoenix, Arizona, {USA.}}} \bibinfo{pages}{2793--2799}.
\newblock


\bibitem[\protect\citeauthoryear{Papineni, Roukos, Ward, and Zhu}{Papineni
  et~al\mbox{.}}{2002}]%
        {papineni2002bleu}
\bibfield{author}{\bibinfo{person}{Kishore Papineni}, \bibinfo{person}{Salim
  Roukos}, \bibinfo{person}{Todd Ward}, {and} \bibinfo{person}{Wei{-}Jing
  Zhu}.} \bibinfo{year}{2002}\natexlab{}.
\newblock \showarticletitle{Bleu: a Method for Automatic Evaluation of Machine
  Translation}. In \bibinfo{booktitle}{\emph{Proceedings of {ACL} 2002, July
  6-12, 2002, Philadelphia, PA, {USA.}}} \bibinfo{pages}{311--318}.
\newblock


\bibitem[\protect\citeauthoryear{Peng, Li, Li, Gao, {\c{C}}elikyilmaz, Lee, and
  Wong}{Peng et~al\mbox{.}}{2017}]%
        {Peng2017EMNLP}
\bibfield{author}{\bibinfo{person}{Baolin Peng}, \bibinfo{person}{Xiujun Li},
  \bibinfo{person}{Lihong Li}, \bibinfo{person}{Jianfeng Gao},
  \bibinfo{person}{Asli {\c{C}}elikyilmaz}, \bibinfo{person}{Sungjin Lee},
  {and} \bibinfo{person}{Kam{-}Fai Wong}.} \bibinfo{year}{2017}\natexlab{}.
\newblock \showarticletitle{Composite Task-Completion Dialogue Policy Learning
  via Hierarchical Deep Reinforcement Learning}. In
  \bibinfo{booktitle}{\emph{Proceedings of {EMNLP} 2017, Copenhagen, Denmark,
  September 9-11, 2017}}. \bibinfo{pages}{2231--2240}.
\newblock


\bibitem[\protect\citeauthoryear{Pennebaker, Francis, and Booth}{Pennebaker
  et~al\mbox{.}}{2001}]%
        {pennebaker2001linguistic}
\bibfield{author}{\bibinfo{person}{James~W Pennebaker},
  \bibinfo{person}{Martha~E Francis}, {and} \bibinfo{person}{Roger~J Booth}.}
  \bibinfo{year}{2001}\natexlab{}.
\newblock \showarticletitle{Linguistic inquiry and word count: LIWC 2001}.
\newblock \bibinfo{journal}{\emph{Mahway: Lawrence Erlbaum Associates}}
  \bibinfo{volume}{71}, \bibinfo{number}{2001} (\bibinfo{year}{2001}),
  \bibinfo{pages}{2001}.
\newblock


\bibitem[\protect\citeauthoryear{Pichl, Marek, Konr{\'{a}}d, Matul{\'{\i}}k,
  Nguyen, and Sediv{\'{y}}}{Pichl et~al\mbox{.}}{2018}]%
        {pichl2018alquist}
\bibfield{author}{\bibinfo{person}{Jan Pichl}, \bibinfo{person}{Petr Marek},
  \bibinfo{person}{Jakub Konr{\'{a}}d}, \bibinfo{person}{Martin
  Matul{\'{\i}}k}, \bibinfo{person}{Hoang~Long Nguyen}, {and}
  \bibinfo{person}{Jan Sediv{\'{y}}}.} \bibinfo{year}{2018}\natexlab{}.
\newblock \showarticletitle{Alquist: The Alexa Prize Socialbot}.
\newblock \bibinfo{journal}{\emph{CoRR}}  \bibinfo{volume}{abs/1804.06705}
  (\bibinfo{year}{2018}).
\newblock


\bibitem[\protect\citeauthoryear{Qian, Huang, Zhao, Xu, and Zhu}{Qian
  et~al\mbox{.}}{2018}]%
        {Qian2017AssigningProfile}
\bibfield{author}{\bibinfo{person}{Qiao Qian}, \bibinfo{person}{Minlie Huang},
  \bibinfo{person}{Haizhou Zhao}, \bibinfo{person}{Jingfang Xu}, {and}
  \bibinfo{person}{Xiaoyan Zhu}.} \bibinfo{year}{2018}\natexlab{}.
\newblock \showarticletitle{Assigning Personality/Profile to a Chatting Machine
  for Coherent Conversation Generation}. In
  \bibinfo{booktitle}{\emph{Proceedings of {IJCAI} 2018, Stockholm, Sweden,
  July 13-19, 2018}}. \bibinfo{pages}{4279--4285}.
\newblock


\bibitem[\protect\citeauthoryear{Radford, J{\'{o}}zefowicz, and
  Sutskever}{Radford et~al\mbox{.}}{2017}]%
        {radford2017learning}
\bibfield{author}{\bibinfo{person}{Alec Radford}, \bibinfo{person}{Rafal
  J{\'{o}}zefowicz}, {and} \bibinfo{person}{Ilya Sutskever}.}
  \bibinfo{year}{2017}\natexlab{}.
\newblock \showarticletitle{Learning to Generate Reviews and Discovering
  Sentiment}.
\newblock \bibinfo{journal}{\emph{CoRR}}  \bibinfo{volume}{abs/1704.01444}
  (\bibinfo{year}{2017}).
\newblock


\bibitem[\protect\citeauthoryear{Ram, Prasad, Khatri, Venkatesh, Gabriel, Liu,
  Nunn, Hedayatnia, Cheng, Nagar, King, Bland, Wartick, Pan, Song, Jayadevan,
  Hwang, and Pettigrue}{Ram et~al\mbox{.}}{2018}]%
        {ram2018conversational}
\bibfield{author}{\bibinfo{person}{Ashwin Ram}, \bibinfo{person}{Rohit Prasad},
  \bibinfo{person}{Chandra Khatri}, \bibinfo{person}{Anu Venkatesh},
  \bibinfo{person}{Raefer Gabriel}, \bibinfo{person}{Qing Liu},
  \bibinfo{person}{Jeff Nunn}, \bibinfo{person}{Behnam Hedayatnia},
  \bibinfo{person}{Ming Cheng}, \bibinfo{person}{Ashish Nagar},
  \bibinfo{person}{Eric King}, \bibinfo{person}{Kate Bland},
  \bibinfo{person}{Amanda Wartick}, \bibinfo{person}{Yi Pan},
  \bibinfo{person}{Han Song}, \bibinfo{person}{Sk Jayadevan},
  \bibinfo{person}{Gene Hwang}, {and} \bibinfo{person}{Art Pettigrue}.}
  \bibinfo{year}{2018}\natexlab{}.
\newblock \showarticletitle{Conversational {AI:} The Science Behind the Alexa
  Prize}.
\newblock \bibinfo{journal}{\emph{CoRR}}  \bibinfo{volume}{abs/1801.03604}
  (\bibinfo{year}{2018}).
\newblock


\bibitem[\protect\citeauthoryear{Ritter, Cherry, and Dolan}{Ritter
  et~al\mbox{.}}{2011}]%
        {a30}
\bibfield{author}{\bibinfo{person}{Alan Ritter}, \bibinfo{person}{Colin
  Cherry}, {and} \bibinfo{person}{William~B. Dolan}.}
  \bibinfo{year}{2011}\natexlab{}.
\newblock \showarticletitle{Data-Driven Response Generation in Social Media}.
  In \bibinfo{booktitle}{\emph{Proceedings of {EMNLP} 2011, John McIntyre
  Conference Centre, Edinburgh, UK, 27-31 July 2011}}.
  \bibinfo{pages}{583--593}.
\newblock


\bibitem[\protect\citeauthoryear{Rojas{-}Barahona, Gasic, Mrksic, Su, Ultes,
  Wen, Young, and Vandyke}{Rojas{-}Barahona et~al\mbox{.}}{2017}]%
        {wen2017network}
\bibfield{author}{\bibinfo{person}{Lina~Maria Rojas{-}Barahona},
  \bibinfo{person}{Milica Gasic}, \bibinfo{person}{Nikola Mrksic},
  \bibinfo{person}{Pei{-}Hao Su}, \bibinfo{person}{Stefan Ultes},
  \bibinfo{person}{Tsung{-}Hsien Wen}, \bibinfo{person}{Steve~J. Young}, {and}
  \bibinfo{person}{David Vandyke}.} \bibinfo{year}{2017}\natexlab{}.
\newblock \showarticletitle{A Network-based End-to-End Trainable Task-oriented
  Dialogue System}. In \bibinfo{booktitle}{\emph{Proceedings of the 15th
  Conference of the European Chapter of the Association for Computational
  Linguistics, {EACL} 2017, Valencia, Spain, April 3-7, 2017}}.
  \bibinfo{pages}{438--449}.
\newblock


\bibitem[\protect\citeauthoryear{Saha, Khapra, and Sankaranarayanan}{Saha
  et~al\mbox{.}}{2017}]%
        {saha2017multimodal}
\bibfield{author}{\bibinfo{person}{Amrita Saha}, \bibinfo{person}{Mitesh
  Khapra}, {and} \bibinfo{person}{Karthik Sankaranarayanan}.}
  \bibinfo{year}{2017}\natexlab{}.
\newblock \showarticletitle{Multimodal Dialogs (MMD): A large-scale dataset for
  studying multimodal domain-aware conversations}.
\newblock \bibinfo{journal}{\emph{arXiv preprint arXiv:1704.00200}}
  (\bibinfo{year}{2017}).
\newblock


\bibitem[\protect\citeauthoryear{Sai, Gupta, Khapra, and Srinivasan}{Sai
  et~al\mbox{.}}{2019}]%
        {sai2019re}
\bibfield{author}{\bibinfo{person}{Ananya Sai}, \bibinfo{person}{Mithun~Das
  Gupta}, \bibinfo{person}{Mitesh~M. Khapra}, {and} \bibinfo{person}{Mukundhan
  Srinivasan}.} \bibinfo{year}{2019}\natexlab{}.
\newblock \showarticletitle{Response Generation by Context-aware Prototype
  Editing}. In \bibinfo{booktitle}{\emph{Proceedings of {AAAI} 2019, Honolulu,
  Hawaii, USA, Jan 27-Feb 1, 2019}}.
\newblock


\bibitem[\protect\citeauthoryear{Serban, Sordoni, Bengio, Courville, and
  Pineau}{Serban et~al\mbox{.}}{2016}]%
        {hiContext}
\bibfield{author}{\bibinfo{person}{Iulian~Vlad Serban},
  \bibinfo{person}{Alessandro Sordoni}, \bibinfo{person}{Yoshua Bengio},
  \bibinfo{person}{Aaron~C. Courville}, {and} \bibinfo{person}{Joelle Pineau}.}
  \bibinfo{year}{2016}\natexlab{}.
\newblock \showarticletitle{Building End-To-End Dialogue Systems Using
  Generative Hierarchical Neural Network Models}. In
  \bibinfo{booktitle}{\emph{Proceedings of {AAAI} 2016, February 12-17, 2016,
  Phoenix, Arizona, {USA.}}} \bibinfo{pages}{3776--3784}.
\newblock


\bibitem[\protect\citeauthoryear{Serban, Sordoni, Lowe, Charlin, Pineau,
  Courville, and Bengio}{Serban et~al\mbox{.}}{2017}]%
        {serban2017hierarchical-1}
\bibfield{author}{\bibinfo{person}{Iulian~Vlad Serban},
  \bibinfo{person}{Alessandro Sordoni}, \bibinfo{person}{Ryan Lowe},
  \bibinfo{person}{Laurent Charlin}, \bibinfo{person}{Joelle Pineau},
  \bibinfo{person}{Aaron~C. Courville}, {and} \bibinfo{person}{Yoshua Bengio}.}
  \bibinfo{year}{2017}\natexlab{}.
\newblock \showarticletitle{A Hierarchical Latent Variable Encoder-Decoder
  Model for Generating Dialogues}. In \bibinfo{booktitle}{\emph{AAAI}}.
  \bibinfo{pages}{3295--3301}.
\newblock


\bibitem[\protect\citeauthoryear{Severyn and Moschitti}{Severyn and
  Moschitti}{2015}]%
        {Severyn:2015gm}
\bibfield{author}{\bibinfo{person}{Aliaksei Severyn} {and}
  \bibinfo{person}{Alessandro Moschitti}.} \bibinfo{year}{2015}\natexlab{}.
\newblock \showarticletitle{Learning to Rank Short Text Pairs with
  Convolutional Deep Neural Networks}. In \bibinfo{booktitle}{\emph{Proceedings
  of {SIGIR} 2015, Santiago, Chile, August 9-13, 2015}}.
  \bibinfo{pages}{373--382}.
\newblock


\bibitem[\protect\citeauthoryear{Shang, Lu, and Li}{Shang
  et~al\mbox{.}}{2015}]%
        {shang-2015-NRM}
\bibfield{author}{\bibinfo{person}{Lifeng Shang}, \bibinfo{person}{Zhengdong
  Lu}, {and} \bibinfo{person}{Hang Li}.} \bibinfo{year}{2015}\natexlab{}.
\newblock \showarticletitle{Neural Responding Machine for Short-Text
  Conversation}. In \bibinfo{booktitle}{\emph{Proceedings of {ACL} 2015, July
  26-31, 2015, Beijing, China}}. \bibinfo{pages}{1577--1586}.
\newblock


\bibitem[\protect\citeauthoryear{Shen, Su, Niu, and Demberg}{Shen
  et~al\mbox{.}}{2018}]%
        {shen18cvae}
\bibfield{author}{\bibinfo{person}{Xiaoyu Shen}, \bibinfo{person}{Hui Su},
  \bibinfo{person}{Shuzi Niu}, {and} \bibinfo{person}{Vera Demberg}.}
  \bibinfo{year}{2018}\natexlab{}.
\newblock \showarticletitle{Improving Variational Encoder-Decoders in Dialogue
  Generation}. In \bibinfo{booktitle}{\emph{Proceedings of {AAAI} 2018, New
  Orleans, Louisiana, USA, February 2-7, 2018}}. \bibinfo{pages}{5456--5463}.
\newblock


\bibitem[\protect\citeauthoryear{Shen, He, Gao, Deng, and Mesnil}{Shen
  et~al\mbox{.}}{2014}]%
        {Shen:2014id}
\bibfield{author}{\bibinfo{person}{Yelong Shen}, \bibinfo{person}{Xiaodong He},
  \bibinfo{person}{Jianfeng Gao}, \bibinfo{person}{Li Deng}, {and}
  \bibinfo{person}{Gr{\'{e}}goire Mesnil}.} \bibinfo{year}{2014}\natexlab{}.
\newblock \showarticletitle{A Latent Semantic Model with Convolutional-Pooling
  Structure for Information Retrieval}. In
  \bibinfo{booktitle}{\emph{Proceedings of the 23rd {ACM} International
  Conference on Conference on Information and Knowledge Management, {CIKM}
  2014, Shanghai, China, November 3-7, 2014}}. \bibinfo{pages}{101--110}.
\newblock


\bibitem[\protect\citeauthoryear{Shum, He, and Li}{Shum et~al\mbox{.}}{2018}]%
        {shum2018eliza}
\bibfield{author}{\bibinfo{person}{Heung{-}Yeung Shum},
  \bibinfo{person}{Xiaodong He}, {and} \bibinfo{person}{Di Li}.}
  \bibinfo{year}{2018}\natexlab{}.
\newblock \showarticletitle{From Eliza to XiaoIce: challenges and opportunities
  with social chatbots}.
\newblock \bibinfo{journal}{\emph{Frontiers of {IT} {\&} {EE}}}
  \bibinfo{volume}{19}, \bibinfo{number}{1} (\bibinfo{year}{2018}),
  \bibinfo{pages}{10--26}.
\newblock


\bibitem[\protect\citeauthoryear{Sohn, Lee, and Yan}{Sohn
  et~al\mbox{.}}{2015}]%
        {sohn2015learning}
\bibfield{author}{\bibinfo{person}{Kihyuk Sohn}, \bibinfo{person}{Honglak Lee},
  {and} \bibinfo{person}{Xinchen Yan}.} \bibinfo{year}{2015}\natexlab{}.
\newblock \showarticletitle{Learning Structured Output Representation using
  Deep Conditional Generative Models}. In \bibinfo{booktitle}{\emph{{NIPS}
  2015, December 7-12, 2015, Montreal, Quebec, Canada}}.
  \bibinfo{pages}{3483--3491}.
\newblock


\bibitem[\protect\citeauthoryear{Song, Li, Nie, Zhang, Zhao, and Yan}{Song
  et~al\mbox{.}}{2018}]%
        {Song:2018ve}
\bibfield{author}{\bibinfo{person}{Yiping Song}, \bibinfo{person}{Cheng{-}Te
  Li}, \bibinfo{person}{Jian{-}Yun Nie}, \bibinfo{person}{Ming Zhang},
  \bibinfo{person}{Dongyan Zhao}, {and} \bibinfo{person}{Rui Yan}.}
  \bibinfo{year}{2018}\natexlab{}.
\newblock \showarticletitle{An Ensemble of Retrieval-Based and Generation-Based
  Human-Computer Conversation Systems}. In
  \bibinfo{booktitle}{\emph{Proceedings of {IJCAI} 2018, Stockholm, Sweden,
  July 13-19, 2018}}. \bibinfo{pages}{4382--4388}.
\newblock


\bibitem[\protect\citeauthoryear{Sordoni, Galley, Auli, Brockett, Ji, Mitchell,
  Nie, Gao, and Dolan}{Sordoni et~al\mbox{.}}{2015}]%
        {sordoni2015neural}
\bibfield{author}{\bibinfo{person}{Alessandro Sordoni}, \bibinfo{person}{Michel
  Galley}, \bibinfo{person}{Michael Auli}, \bibinfo{person}{Chris Brockett},
  \bibinfo{person}{Yangfeng Ji}, \bibinfo{person}{Margaret Mitchell},
  \bibinfo{person}{Jian-Yun Nie}, \bibinfo{person}{Jianfeng Gao}, {and}
  \bibinfo{person}{Bill Dolan}.} \bibinfo{year}{2015}\natexlab{}.
\newblock \showarticletitle{A Neural Network Approach to Context-Sensitive
  Generation of Conversational Responses}. In \bibinfo{booktitle}{\emph{{NAACL}
  {HLT} 2015, Denver, Colorado, USA, May 31 - June 5, 2015}}.
  \bibinfo{pages}{196--205}.
\newblock


\bibitem[\protect\citeauthoryear{Speer, Chin, and Havasi}{Speer
  et~al\mbox{.}}{2017}]%
        {speer2013conceptnet}
\bibfield{author}{\bibinfo{person}{Robyn Speer}, \bibinfo{person}{Joshua Chin},
  {and} \bibinfo{person}{Catherine Havasi}.} \bibinfo{year}{2017}\natexlab{}.
\newblock \showarticletitle{ConceptNet 5.5: An Open Multilingual Graph of
  General Knowledge}. In \bibinfo{booktitle}{\emph{Proceedings of {AAAI} 2017,
  San Francisco, California, {USA.}, February 4-9, 2017}}.
  \bibinfo{pages}{4444--4451}.
\newblock


\bibitem[\protect\citeauthoryear{Stanojevi\'{c} and Sima'an}{Stanojevi\'{c} and
  Sima'an}{2014}]%
        {stanojevic14fitting}
\bibfield{author}{\bibinfo{person}{Milo\v{s} Stanojevi\'{c}} {and}
  \bibinfo{person}{Khalil Sima'an}.} \bibinfo{year}{2014}\natexlab{}.
\newblock \showarticletitle{Fitting Sentence Level Translation Evaluation with
  Many Dense Features}. In \bibinfo{booktitle}{\emph{Proceedings of {EMNLP}
  2014, Doha, Qatar, October 25-29, 2014}}. \bibinfo{pages}{202--206}.
\newblock


\bibitem[\protect\citeauthoryear{Su, Gasic, Mrksic, Rojas{-}Barahona, Ultes,
  Vandyke, Wen, and Young}{Su et~al\mbox{.}}{2016}]%
        {Su2016ACL}
\bibfield{author}{\bibinfo{person}{Pei{-}Hao Su}, \bibinfo{person}{Milica
  Gasic}, \bibinfo{person}{Nikola Mrksic}, \bibinfo{person}{Lina~Maria
  Rojas{-}Barahona}, \bibinfo{person}{Stefan Ultes}, \bibinfo{person}{David
  Vandyke}, \bibinfo{person}{Tsung{-}Hsien Wen}, {and}
  \bibinfo{person}{Steve~J. Young}.} \bibinfo{year}{2016}\natexlab{}.
\newblock \showarticletitle{On-line Active Reward Learning for Policy
  Optimisation in Spoken Dialogue Systems}. In
  \bibinfo{booktitle}{\emph{Proceedings of {ACL} 2016, Berlin, Germany, August
  7-12, 2016}}.
\newblock


\bibitem[\protect\citeauthoryear{Sutskever, Vinyals, and Le}{Sutskever
  et~al\mbox{.}}{2014}]%
        {2014sequence}
\bibfield{author}{\bibinfo{person}{Ilya Sutskever}, \bibinfo{person}{Oriol
  Vinyals}, {and} \bibinfo{person}{Quoc~V. Le}.}
  \bibinfo{year}{2014}\natexlab{}.
\newblock \showarticletitle{Sequence to Sequence Learning with Neural
  Networks}. In \bibinfo{booktitle}{\emph{{NIPS} 2014, Montreal, Quebec,
  Canada, December 8-13, 2014}}. \bibinfo{pages}{3104--3112}.
\newblock


\bibitem[\protect\citeauthoryear{Szegedy, Zaremba, Sutskever, Bruna, Erhan,
  Goodfellow, and Fergus}{Szegedy et~al\mbox{.}}{2013}]%
        {szegedy13intriging}
\bibfield{author}{\bibinfo{person}{Christian Szegedy},
  \bibinfo{person}{Wojciech Zaremba}, \bibinfo{person}{Ilya Sutskever},
  \bibinfo{person}{Joan Bruna}, \bibinfo{person}{Dumitru Erhan},
  \bibinfo{person}{Ian~J. Goodfellow}, {and} \bibinfo{person}{Rob Fergus}.}
  \bibinfo{year}{2013}\natexlab{}.
\newblock \showarticletitle{Intriguing properties of neural networks}.
\newblock \bibinfo{journal}{\emph{CoRR}}  \bibinfo{volume}{abs/1312.6199}
  (\bibinfo{year}{2013}).
\newblock


\bibitem[\protect\citeauthoryear{Tao, Mou, Zhao, and Yan}{Tao
  et~al\mbox{.}}{2018}]%
        {tao2018ruber}
\bibfield{author}{\bibinfo{person}{Chongyang Tao}, \bibinfo{person}{Lili Mou},
  \bibinfo{person}{Dongyan Zhao}, {and} \bibinfo{person}{Rui Yan}.}
  \bibinfo{year}{2018}\natexlab{}.
\newblock \showarticletitle{{RUBER:} An Unsupervised Method for Automatic
  Evaluation of Open-Domain Dialog Systems}. In
  \bibinfo{booktitle}{\emph{Proceedings of {AAAI} 2018, New Orleans, Louisiana,
  USA, February 2-7, 2018}}. \bibinfo{pages}{722--729}.
\newblock


\bibitem[\protect\citeauthoryear{Turing}{Turing}{1950}]%
        {alan1950}
\bibfield{author}{\bibinfo{person}{Alan~M Turing}.}
  \bibinfo{year}{1950}\natexlab{}.
\newblock \showarticletitle{Computing machinery and intelligence}.
\newblock \bibinfo{journal}{\emph{Mind}} \bibinfo{volume}{59},
  \bibinfo{number}{236} (\bibinfo{year}{1950}), \bibinfo{pages}{433--460}.
\newblock


\bibitem[\protect\citeauthoryear{Vijayakumar, Cogswell, Selvaraju, Sun, Lee,
  Crandall, and Batra}{Vijayakumar et~al\mbox{.}}{2018}]%
        {Ashwin2018DBS}
\bibfield{author}{\bibinfo{person}{Ashwin~K. Vijayakumar},
  \bibinfo{person}{Michael Cogswell}, \bibinfo{person}{Ramprasaath~R.
  Selvaraju}, \bibinfo{person}{Qing Sun}, \bibinfo{person}{Stefan Lee},
  \bibinfo{person}{David~J. Crandall}, {and} \bibinfo{person}{Dhruv Batra}.}
  \bibinfo{year}{2018}\natexlab{}.
\newblock \showarticletitle{Diverse Beam Search for Improved Description of
  Complex Scenes}. In \bibinfo{booktitle}{\emph{Proceedings of {AAAI} 2018, New
  Orleans, Louisiana, USA, February 2-7, 2018}}. \bibinfo{pages}{7371--7379}.
\newblock


\bibitem[\protect\citeauthoryear{Vinyals and Le}{Vinyals and Le}{2015}]%
        {vinyals2015neural}
\bibfield{author}{\bibinfo{person}{Oriol Vinyals} {and}
  \bibinfo{person}{Quoc~V. Le}.} \bibinfo{year}{2015}\natexlab{}.
\newblock \showarticletitle{A Neural Conversational Model}.
\newblock \bibinfo{journal}{\emph{CoRR}}  \bibinfo{volume}{abs/1506.05869}
  (\bibinfo{year}{2015}).
\newblock


\bibitem[\protect\citeauthoryear{Wallace}{Wallace}{2009}]%
        {wallace2009anatomy}
\bibfield{author}{\bibinfo{person}{Richard~S Wallace}.}
  \bibinfo{year}{2009}\natexlab{}.
\newblock \showarticletitle{The anatomy of ALICE}.
\newblock In \bibinfo{booktitle}{\emph{Parsing the Turing Test}}.
  \bibinfo{publisher}{Springer}, \bibinfo{pages}{181--210}.
\newblock


\bibitem[\protect\citeauthoryear{Wang, Jojic, Brockett, and Nyberg}{Wang
  et~al\mbox{.}}{2017b}]%
        {wang2017steering}
\bibfield{author}{\bibinfo{person}{Di Wang}, \bibinfo{person}{Nebojsa Jojic},
  \bibinfo{person}{Chris Brockett}, {and} \bibinfo{person}{Eric Nyberg}.}
  \bibinfo{year}{2017}\natexlab{b}.
\newblock \showarticletitle{Steering Output Style and Topic in Neural Response
  Generation}. In \bibinfo{booktitle}{\emph{Proceedings of {EMNLP} 2017,
  Copenhagen, Denmark, September 9-11, 2017}}. \bibinfo{pages}{2140--2150}.
\newblock


\bibitem[\protect\citeauthoryear{Wang, Wang, Li, Xu, Wang, and Wang}{Wang
  et~al\mbox{.}}{2017c}]%
        {wang2017groupbias}
\bibfield{author}{\bibinfo{person}{Jianan Wang}, \bibinfo{person}{Xin Wang},
  \bibinfo{person}{Fang Li}, \bibinfo{person}{Zhen Xu},
  \bibinfo{person}{Zhuoran Wang}, {and} \bibinfo{person}{Baoxun Wang}.}
  \bibinfo{year}{2017}\natexlab{c}.
\newblock \showarticletitle{Group Linguistic Bias Aware Neural Response
  Generation}. In \bibinfo{booktitle}{\emph{Proceedings of the 9th {SIGHAN}
  Workshop on Chinese Language Processing, SIGHAN@IJCNLP 2017, Taipei, Taiwan,
  December 1, 2017}}. \bibinfo{pages}{1--10}.
\newblock


\bibitem[\protect\citeauthoryear{Wang and Wan}{Wang and Wan}{2018}]%
        {wang2018sentigan}
\bibfield{author}{\bibinfo{person}{Ke Wang} {and} \bibinfo{person}{Xiaojun
  Wan}.} \bibinfo{year}{2018}\natexlab{}.
\newblock \showarticletitle{SentiGAN: Generating Sentimental Texts via Mixture
  Adversarial Networks}. In \bibinfo{booktitle}{\emph{Proceedings of {IJCAI}
  2018, Stockholm, Sweden, July 13-19, 2018}}. \bibinfo{pages}{4446--4452}.
\newblock


\bibitem[\protect\citeauthoryear{Wang, Huang, Xu, Shen, and Nie}{Wang
  et~al\mbox{.}}{2018a}]%
        {wang2018chat}
\bibfield{author}{\bibinfo{person}{Wenjie Wang}, \bibinfo{person}{Minlie
  Huang}, \bibinfo{person}{Xin{-}Shun Xu}, \bibinfo{person}{Fumin Shen}, {and}
  \bibinfo{person}{Liqiang Nie}.} \bibinfo{year}{2018}\natexlab{a}.
\newblock \showarticletitle{Chat More: Deepening and Widening the Chatting
  Topic via {A} Deep Model}. In \bibinfo{booktitle}{\emph{{SIGIR} 2018, Ann
  Arbor, MI, USA, July 08-12, 2018}}. \bibinfo{pages}{255--264}.
\newblock


\bibitem[\protect\citeauthoryear{Wang, Liu, Huang, and Nie}{Wang
  et~al\mbox{.}}{2018b}]%
        {askquestion18}
\bibfield{author}{\bibinfo{person}{Yansen Wang}, \bibinfo{person}{Chenyi Liu},
  \bibinfo{person}{Minlie Huang}, {and} \bibinfo{person}{Liqiang Nie}.}
  \bibinfo{year}{2018}\natexlab{b}.
\newblock \showarticletitle{Learning to Ask Questions in Open-domain
  Conversational Systems with Typed Decoders}. In
  \bibinfo{booktitle}{\emph{Proceedings of {ACL} 2018, Melbourne, Australia,
  July 15-20, 2018}}. \bibinfo{pages}{2193--2203}.
\newblock


\bibitem[\protect\citeauthoryear{Wang, Hamza, and Florian}{Wang
  et~al\mbox{.}}{2017a}]%
        {Wang2017bimpm}
\bibfield{author}{\bibinfo{person}{Zhiguo Wang}, \bibinfo{person}{Wael Hamza},
  {and} \bibinfo{person}{Radu Florian}.} \bibinfo{year}{2017}\natexlab{a}.
\newblock \showarticletitle{Bilateral Multi-Perspective Matching for Natural
  Language Sentences}. In \bibinfo{booktitle}{\emph{Proceedings of {IJCAI}
  2017, Melbourne, Australia, August 19-25, 2017}}.
  \bibinfo{pages}{4144--4150}.
\newblock


\bibitem[\protect\citeauthoryear{Warriner, Kuperman, and Brysbaert}{Warriner
  et~al\mbox{.}}{2013}]%
        {warriner2013norms}
\bibfield{author}{\bibinfo{person}{Amy~Beth Warriner}, \bibinfo{person}{Victor
  Kuperman}, {and} \bibinfo{person}{Marc Brysbaert}.}
  \bibinfo{year}{2013}\natexlab{}.
\newblock \showarticletitle{Norms of valence, arousal, and dominance for 13,915
  English lemmas}.
\newblock \bibinfo{journal}{\emph{Behavior research methods}}
  \bibinfo{volume}{45}, \bibinfo{number}{4} (\bibinfo{year}{2013}),
  \bibinfo{pages}{1191--1207}.
\newblock


\bibitem[\protect\citeauthoryear{Weizenbaum}{Weizenbaum}{1966}]%
        {weizenbaum1966eliza}
\bibfield{author}{\bibinfo{person}{Joseph Weizenbaum}.}
  \bibinfo{year}{1966}\natexlab{}.
\newblock \showarticletitle{{ELIZA} - a computer program for the study of
  natural language communication between man and machine}.
\newblock \bibinfo{journal}{\emph{Commun. {ACM}}} \bibinfo{volume}{9},
  \bibinfo{number}{1} (\bibinfo{year}{1966}), \bibinfo{pages}{36--45}.
\newblock


\bibitem[\protect\citeauthoryear{Winata, Kampman, Yang, Dey, and Fung}{Winata
  et~al\mbox{.}}{2017}]%
        {winata2017nora}
\bibfield{author}{\bibinfo{person}{Genta~Indra Winata}, \bibinfo{person}{Onno
  Kampman}, \bibinfo{person}{Yang Yang}, \bibinfo{person}{Anik Dey}, {and}
  \bibinfo{person}{Pascale Fung}.} \bibinfo{year}{2017}\natexlab{}.
\newblock \showarticletitle{Nora the empathetic psychologist}. In
  \bibinfo{booktitle}{\emph{Proc. Interspeech}}. \bibinfo{pages}{3437--3438}.
\newblock


\bibitem[\protect\citeauthoryear{Wu, Wei, Huang, Li, and Zhou}{Wu
  et~al\mbox{.}}{2019}]%
        {wu2019prototype}
\bibfield{author}{\bibinfo{person}{Yu Wu}, \bibinfo{person}{Furu Wei},
  \bibinfo{person}{Shaohan Huang}, \bibinfo{person}{Zhoujun Li}, {and}
  \bibinfo{person}{Ming Zhou}.} \bibinfo{year}{2019}\natexlab{}.
\newblock \showarticletitle{Response Generation by Context-aware Prototype
  Editing}. In \bibinfo{booktitle}{\emph{Proceedings of {AAAI} 2019, Honolulu,
  Hawaii, USA, January 27-February 1, 2019}}.
\newblock


\bibitem[\protect\citeauthoryear{Wu, Wu, Xing, Zhou, and Li}{Wu
  et~al\mbox{.}}{2017}]%
        {wu2017sequential}
\bibfield{author}{\bibinfo{person}{Yu Wu}, \bibinfo{person}{Wei Wu},
  \bibinfo{person}{Chen Xing}, \bibinfo{person}{Ming Zhou}, {and}
  \bibinfo{person}{Zhoujun Li}.} \bibinfo{year}{2017}\natexlab{}.
\newblock \showarticletitle{Sequential Matching Network: {A} New Architecture
  for Multi-turn Response Selection in Retrieval-Based Chatbots}. In
  \bibinfo{booktitle}{\emph{Proceedings of {ACL} 2017, Vancouver, Canada, July
  30-August 4, 2017}}. \bibinfo{pages}{496--505}.
\newblock


\bibitem[\protect\citeauthoryear{Xing, Wu, Wu, Liu, Huang, Zhou, and Ma}{Xing
  et~al\mbox{.}}{2017}]%
        {xing2017topic}
\bibfield{author}{\bibinfo{person}{Chen Xing}, \bibinfo{person}{Wei Wu},
  \bibinfo{person}{Yu Wu}, \bibinfo{person}{Jie Liu}, \bibinfo{person}{Yalou
  Huang}, \bibinfo{person}{Ming Zhou}, {and} \bibinfo{person}{Wei{-}Ying Ma}.}
  \bibinfo{year}{2017}\natexlab{}.
\newblock \showarticletitle{Topic Aware Neural Response Generation}. In
  \bibinfo{booktitle}{\emph{Proceedings of {AAAI} 2017, San Francisco,
  California, {USA.}, February 4-9, 2017}}. \bibinfo{pages}{3351--3357}.
\newblock


\bibitem[\protect\citeauthoryear{Xu, Ren, Lin, and Sun}{Xu
  et~al\mbox{.}}{2018b}]%
        {xu2018dpgan}
\bibfield{author}{\bibinfo{person}{Jingjing Xu}, \bibinfo{person}{Xuancheng
  Ren}, \bibinfo{person}{Junyang Lin}, {and} \bibinfo{person}{Xu Sun}.}
  \bibinfo{year}{2018}\natexlab{b}.
\newblock \showarticletitle{Diversity-Promoting {GAN:} {A} Cross-Entropy Based
  Generative Adversarial Network for Diversified Text Generation}. In
  \bibinfo{booktitle}{\emph{Proceedings of {EMNLP} 2018, Brussels, Belgium,
  October 31 - November 4, 2018}}. \bibinfo{pages}{3940--3949}.
\newblock


\bibitem[\protect\citeauthoryear{Xu, Madotto, Wu, Park, and Fung}{Xu
  et~al\mbox{.}}{2018a}]%
        {xu2018emo2vec}
\bibfield{author}{\bibinfo{person}{Peng Xu}, \bibinfo{person}{Andrea Madotto},
  \bibinfo{person}{Chien-Sheng Wu}, \bibinfo{person}{Ji~Ho Park}, {and}
  \bibinfo{person}{Pascale Fung}.} \bibinfo{year}{2018}\natexlab{a}.
\newblock \showarticletitle{Emo2Vec: Learning Generalized Emotion
  Representation by Multi-task Training}.
\newblock \bibinfo{journal}{\emph{arXiv preprint arXiv:1809.04505}}
  (\bibinfo{year}{2018}).
\newblock


\bibitem[\protect\citeauthoryear{Yan, Song, and Wu}{Yan et~al\mbox{.}}{2016}]%
        {DL2R}
\bibfield{author}{\bibinfo{person}{Rui Yan}, \bibinfo{person}{Yiping Song},
  {and} \bibinfo{person}{Hua Wu}.} \bibinfo{year}{2016}\natexlab{}.
\newblock \showarticletitle{Learning to Respond with Deep Neural Networks for
  Retrieval-Based Human-Computer Conversation System}. In
  \bibinfo{booktitle}{\emph{Proceedings of {SIGIR} 2016, Pisa, Italy, July
  17-21, 2016}}. \bibinfo{pages}{55--64}.
\newblock


\bibitem[\protect\citeauthoryear{Yan and Zhao}{Yan and Zhao}{2018}]%
        {prosuggestion18}
\bibfield{author}{\bibinfo{person}{Rui Yan} {and} \bibinfo{person}{Dongyan
  Zhao}.} \bibinfo{year}{2018}\natexlab{}.
\newblock \showarticletitle{Smarter Response with Proactive Suggestion: {A} New
  Generative Neural Conversation Paradigm}. In
  \bibinfo{booktitle}{\emph{Proceedings of {IJCAI} 2018, Stockholm, Sweden,
  July 13-19, 2018}}. \bibinfo{pages}{4525--4531}.
\newblock


\bibitem[\protect\citeauthoryear{Yang, Hu, Qiu, Qu, Gao, Croft, Liu, Shen, and
  Liu}{Yang et~al\mbox{.}}{2019}]%
        {yang2019hybrid}
\bibfield{author}{\bibinfo{person}{Liu Yang}, \bibinfo{person}{Junjie Hu},
  \bibinfo{person}{Minghui Qiu}, \bibinfo{person}{Chen Qu},
  \bibinfo{person}{Jianfeng Gao}, \bibinfo{person}{W~Bruce Croft},
  \bibinfo{person}{Xiaodong Liu}, \bibinfo{person}{Yelong Shen}, {and}
  \bibinfo{person}{Jingjing Liu}.} \bibinfo{year}{2019}\natexlab{}.
\newblock \showarticletitle{A Hybrid Retrieval-Generation Neural Conversation
  Model}.
\newblock \bibinfo{journal}{\emph{arXiv preprint arXiv:1904.09068}}
  (\bibinfo{year}{2019}).
\newblock


\bibitem[\protect\citeauthoryear{Yang, Zhao, Zhao, Chen, Zhu, Zhou, and
  Cao}{Yang et~al\mbox{.}}{2017}]%
        {npm}
\bibfield{author}{\bibinfo{person}{Min Yang}, \bibinfo{person}{Zhou Zhao},
  \bibinfo{person}{Wei Zhao}, \bibinfo{person}{Xiaojun Chen},
  \bibinfo{person}{Jia Zhu}, \bibinfo{person}{Lianqiang Zhou}, {and}
  \bibinfo{person}{Zigang Cao}.} \bibinfo{year}{2017}\natexlab{}.
\newblock \showarticletitle{Personalized Response Generation via Domain
  adaptation}. In \bibinfo{booktitle}{\emph{Proceedings of {SIGIR} 2017, Tokyo,
  Japan, August 7-11, 2017}}. \bibinfo{pages}{1021--1024}.
\newblock


\bibitem[\protect\citeauthoryear{Young, Cambria, Chaturvedi, Zhou, Biswas, and
  Huang}{Young et~al\mbox{.}}{2018}]%
        {young2018augmenting}
\bibfield{author}{\bibinfo{person}{Tom Young}, \bibinfo{person}{Erik Cambria},
  \bibinfo{person}{Iti Chaturvedi}, \bibinfo{person}{Hao Zhou},
  \bibinfo{person}{Subham Biswas}, {and} \bibinfo{person}{Minlie Huang}.}
  \bibinfo{year}{2018}\natexlab{}.
\newblock \showarticletitle{Augmenting End-to-End Dialogue Systems With
  Commonsense Knowledge}. In \bibinfo{booktitle}{\emph{Proceedings of {AAAI}
  2018, New Orleans, Louisiana, USA, February 2-7, 2018}}.
  \bibinfo{pages}{4970--4977}.
\newblock


\bibitem[\protect\citeauthoryear{Yu, Xu, Black, and Rudnicky}{Yu
  et~al\mbox{.}}{2016}]%
        {Yu2016strategy}
\bibfield{author}{\bibinfo{person}{Zhou Yu}, \bibinfo{person}{Ziyu Xu},
  \bibinfo{person}{Alan~W. Black}, {and} \bibinfo{person}{Alexander~I.
  Rudnicky}.} \bibinfo{year}{2016}\natexlab{}.
\newblock \showarticletitle{Strategy and Policy Learning for Non-Task-Oriented
  Conversational Systems}. In \bibinfo{booktitle}{\emph{Proceedings of
  {SIGDIAL} 2016, Los Angeles, CA, {USA}, September 13-15, 2016}}.
  \bibinfo{pages}{404--412}.
\newblock


\bibitem[\protect\citeauthoryear{Zhang, Chang, Danescu{-}Niculescu{-}Mizil,
  Dixon, Hua, Taraborelli, and Thain}{Zhang et~al\mbox{.}}{2018a}]%
        {earlysign18}
\bibfield{author}{\bibinfo{person}{Justine Zhang}, \bibinfo{person}{Jonathan~P.
  Chang}, \bibinfo{person}{Cristian Danescu{-}Niculescu{-}Mizil},
  \bibinfo{person}{Lucas Dixon}, \bibinfo{person}{Yiqing Hua},
  \bibinfo{person}{Dario Taraborelli}, {and} \bibinfo{person}{Nithum Thain}.}
  \bibinfo{year}{2018}\natexlab{a}.
\newblock \showarticletitle{Conversations Gone Awry: Detecting Early Signs of
  Conversational Failure}. In \bibinfo{booktitle}{\emph{Proceedings of {ACL}
  2018, Melbourne, Australia, July 15-20, 2018}}. \bibinfo{pages}{1350--1361}.
\newblock


\bibitem[\protect\citeauthoryear{Zhang, Guo, Fan, Lan, Xu, and Cheng}{Zhang
  et~al\mbox{.}}{2018c}]%
        {zhang2018learning}
\bibfield{author}{\bibinfo{person}{Ruqing Zhang}, \bibinfo{person}{Jiafeng
  Guo}, \bibinfo{person}{Yixing Fan}, \bibinfo{person}{Yanyan Lan},
  \bibinfo{person}{Jun Xu}, {and} \bibinfo{person}{Xueqi Cheng}.}
  \bibinfo{year}{2018}\natexlab{c}.
\newblock \showarticletitle{Learning to Control the Specificity in Neural
  Response Generation}. In \bibinfo{booktitle}{\emph{Proceedings of {ACL} 2018,
  Melbourne, Australia, July 15-20, 2018}}. \bibinfo{pages}{1108--1117}.
\newblock


\bibitem[\protect\citeauthoryear{Zhang, Lee, Polymenakos, and Radev}{Zhang
  et~al\mbox{.}}{2018d}]%
        {zhang2018addressee}
\bibfield{author}{\bibinfo{person}{Rui Zhang}, \bibinfo{person}{Honglak Lee},
  \bibinfo{person}{Lazaros Polymenakos}, {and} \bibinfo{person}{Dragomir~R.
  Radev}.} \bibinfo{year}{2018}\natexlab{d}.
\newblock \showarticletitle{Addressee and Response Selection in Multi-Party
  Conversations With Speaker Interaction RNNs}. In
  \bibinfo{booktitle}{\emph{Proceedings of {AAAI} 2018, New Orleans, Louisiana,
  USA, February 2-7, 2018}}. \bibinfo{pages}{5690--5697}.
\newblock


\bibitem[\protect\citeauthoryear{Zhang, Dinan, Urbanek, Szlam, Kiela, and
  Weston}{Zhang et~al\mbox{.}}{2018b}]%
        {Zhang2018Personalizing-dogpet}
\bibfield{author}{\bibinfo{person}{Saizheng Zhang}, \bibinfo{person}{Emily
  Dinan}, \bibinfo{person}{Jack Urbanek}, \bibinfo{person}{Arthur Szlam},
  \bibinfo{person}{Douwe Kiela}, {and} \bibinfo{person}{Jason Weston}.}
  \bibinfo{year}{2018}\natexlab{b}.
\newblock \showarticletitle{Personalizing Dialogue Agents: {I} have a dog, do
  you have pets too?}. In \bibinfo{booktitle}{\emph{Proceedings of {ACL} 2018,
  Melbourne, Australia, July 15-20, 2018}}. \bibinfo{pages}{2204--2213}.
\newblock


\bibitem[\protect\citeauthoryear{Zhang, Zhu, Wang, Zhao, and Liu}{Zhang
  et~al\mbox{.}}{2017}]%
        {zhang2017neural}
\bibfield{author}{\bibinfo{person}{Wei-Nan Zhang}, \bibinfo{person}{Qingfu
  Zhu}, \bibinfo{person}{Yifa Wang}, \bibinfo{person}{Yanyan Zhao}, {and}
  \bibinfo{person}{Ting Liu}.} \bibinfo{year}{2017}\natexlab{}.
\newblock \showarticletitle{Neural personalized response generation as domain
  adaptation}.
\newblock \bibinfo{journal}{\emph{World Wide Web}} (\bibinfo{year}{2017}),
  \bibinfo{pages}{1--20}.
\newblock


\bibitem[\protect\citeauthoryear{Zhang, Huang, Zhao, Ji, Chen, and Zhu}{Zhang
  et~al\mbox{.}}{2019a}]%
        {zhang2019tois}
\bibfield{author}{\bibinfo{person}{Zheng Zhang}, \bibinfo{person}{Minlie
  Huang}, \bibinfo{person}{Zhongzhou Zhao}, \bibinfo{person}{Feng Ji},
  \bibinfo{person}{Haiqing Chen}, {and} \bibinfo{person}{Xiaoyan Zhu}.}
  \bibinfo{year}{2019}\natexlab{a}.
\newblock \showarticletitle{Memory-augmented Dialogue Management for
  Task-oriented Dialogue Systems}.
\newblock \bibinfo{journal}{\emph{ACM Transactions on Information Systems}}
  \bibinfo{volume}{1} (\bibinfo{year}{2019}).
\newblock


\bibitem[\protect\citeauthoryear{Zhang, Liao, Huang, Zhu, and Chua}{Zhang
  et~al\mbox{.}}{2019b}]%
        {zhang2019mnbt}
\bibfield{author}{\bibinfo{person}{Zheng Zhang}, \bibinfo{person}{Lizi Liao},
  \bibinfo{person}{Minlie Huang}, \bibinfo{person}{Xiaoyan Zhu}, {and}
  \bibinfo{person}{Tat-Seng Chua}.} \bibinfo{year}{2019}\natexlab{b}.
\newblock \showarticletitle{Neural Multimodal Belief Tracker with Adaptive
  Attention for Dialogue Systems}. In \bibinfo{booktitle}{\emph{The World Wide
  Web Conference}}. ACM, \bibinfo{pages}{2401--2412}.
\newblock


\bibitem[\protect\citeauthoryear{Zhao and Esk{\'{e}}nazi}{Zhao and
  Esk{\'{e}}nazi}{2016}]%
        {zhao2016towards}
\bibfield{author}{\bibinfo{person}{Tiancheng Zhao} {and}
  \bibinfo{person}{Maxine Esk{\'{e}}nazi}.} \bibinfo{year}{2016}\natexlab{}.
\newblock \showarticletitle{Towards End-to-End Learning for Dialog State
  Tracking and Management using Deep Reinforcement Learning}. In
  \bibinfo{booktitle}{\emph{Proceedings of {SIGDIAL} 2016, Los Angeles, CA,
  {USA}, September 13-15, 2016}}. \bibinfo{pages}{1--10}.
\newblock


\bibitem[\protect\citeauthoryear{Zhao, Lee, and Esk{\'{e}}nazi}{Zhao
  et~al\mbox{.}}{2018}]%
        {zhao18discrete}
\bibfield{author}{\bibinfo{person}{Tiancheng Zhao}, \bibinfo{person}{Kyusong
  Lee}, {and} \bibinfo{person}{Maxine Esk{\'{e}}nazi}.}
  \bibinfo{year}{2018}\natexlab{}.
\newblock \showarticletitle{Unsupervised Discrete Sentence Representation
  Learning for Interpretable Neural Dialog Generation}. In
  \bibinfo{booktitle}{\emph{Proceedings of {ACL} 2018, Melbourne, Australia,
  July 15-20, 2018}}. \bibinfo{pages}{1098--1107}.
\newblock


\bibitem[\protect\citeauthoryear{Zhao, Zhao, and Esk{\'{e}}nazi}{Zhao
  et~al\mbox{.}}{2017}]%
        {zhao2017cvae}
\bibfield{author}{\bibinfo{person}{Tiancheng Zhao}, \bibinfo{person}{Ran Zhao},
  {and} \bibinfo{person}{Maxine Esk{\'{e}}nazi}.}
  \bibinfo{year}{2017}\natexlab{}.
\newblock \showarticletitle{Learning Discourse-level Diversity for Neural
  Dialog Models using Conditional Variational Autoencoders}. In
  \bibinfo{booktitle}{\emph{Proceedings of {ACL} 2017, Vancouver, Canada, July
  30-August 4, 2017}}. \bibinfo{pages}{654--664}.
\newblock


\bibitem[\protect\citeauthoryear{Zheng, Chen, Huang, Liu, and Zhu}{Zheng
  et~al\mbox{.}}{2019}]%
        {zheng2019personalized}
\bibfield{author}{\bibinfo{person}{Yinhe Zheng}, \bibinfo{person}{Guanyi Chen},
  \bibinfo{person}{Minlie Huang}, \bibinfo{person}{Song Liu}, {and}
  \bibinfo{person}{Xuan Zhu}.} \bibinfo{year}{2019}\natexlab{}.
\newblock \showarticletitle{Personalized Dialogue Generation with Diversified
  Traits}.
\newblock \bibinfo{journal}{\emph{CoRR}}  \bibinfo{volume}{abs/1901.09672}
  (\bibinfo{year}{2019}).
\newblock


\bibitem[\protect\citeauthoryear{Zhou, Luo, Cao, Lin, Chen, and He}{Zhou
  et~al\mbox{.}}{2017}]%
        {zhou2017mechanism}
\bibfield{author}{\bibinfo{person}{Ganbin Zhou}, \bibinfo{person}{Ping Luo},
  \bibinfo{person}{Rongyu Cao}, \bibinfo{person}{Fen Lin}, \bibinfo{person}{Bo
  Chen}, {and} \bibinfo{person}{Qing He}.} \bibinfo{year}{2017}\natexlab{}.
\newblock \showarticletitle{Mechanism-Aware Neural Machine for Dialogue
  Response Generation}. In \bibinfo{booktitle}{\emph{Proceedings of {AAAI}
  2017, San Francisco, California, {USA.}, February 4-9, 2017}}.
  \bibinfo{pages}{3400--3407}.
\newblock


\bibitem[\protect\citeauthoryear{Zhou, Luo, Xiao, Lin, Chen, and He}{Zhou
  et~al\mbox{.}}{2018c}]%
        {zhou2018elastic}
\bibfield{author}{\bibinfo{person}{Ganbin Zhou}, \bibinfo{person}{Ping Luo},
  \bibinfo{person}{Yijun Xiao}, \bibinfo{person}{Fen Lin}, \bibinfo{person}{Bo
  Chen}, {and} \bibinfo{person}{Qing He}.} \bibinfo{year}{2018}\natexlab{c}.
\newblock \showarticletitle{Elastic Responding Machine for Dialog Generation
  with Dynamically Mechanism Selecting}. In
  \bibinfo{booktitle}{\emph{Proceedings of {AAAI} 2018, New Orleans, Louisiana,
  USA, February 2-7, 2018}}. \bibinfo{pages}{5730--5737}.
\newblock


\bibitem[\protect\citeauthoryear{Zhou, Huang, Zhang, Zhu, and Liu}{Zhou
  et~al\mbox{.}}{2018b}]%
        {Zhou2018EmotionalCM}
\bibfield{author}{\bibinfo{person}{Hao Zhou}, \bibinfo{person}{Minlie Huang},
  \bibinfo{person}{Tianyang Zhang}, \bibinfo{person}{Xiaoyan Zhu}, {and}
  \bibinfo{person}{Bing Liu}.} \bibinfo{year}{2018}\natexlab{b}.
\newblock \showarticletitle{Emotional Chatting Machine: Emotional Conversation
  Generation with Internal and External Memory}. In
  \bibinfo{booktitle}{\emph{Proceedings of {AAAI} 2018, New Orleans, Louisiana,
  USA, February 2-7, 2018}}. \bibinfo{pages}{730--739}.
\newblock


\bibitem[\protect\citeauthoryear{Zhou, Young, Huang, Zhao, Xu, and Zhu}{Zhou
  et~al\mbox{.}}{2018e}]%
        {Zhou2018Commonsense}
\bibfield{author}{\bibinfo{person}{Hao Zhou}, \bibinfo{person}{Tom Young},
  \bibinfo{person}{Minlie Huang}, \bibinfo{person}{Haizhou Zhao},
  \bibinfo{person}{Jingfang Xu}, {and} \bibinfo{person}{Xiaoyan Zhu}.}
  \bibinfo{year}{2018}\natexlab{e}.
\newblock \showarticletitle{Commonsense Knowledge Aware Conversation Generation
  with Graph Attention}. In \bibinfo{booktitle}{\emph{Proceedings of {IJCAI}
  2018, Stockholm, Sweden, July 13-19, 2018}}. \bibinfo{pages}{4623--4629}.
\newblock


\bibitem[\protect\citeauthoryear{Zhou, Prabhumoye, and Black}{Zhou
  et~al\mbox{.}}{2018d}]%
        {zhou2018DoG}
\bibfield{author}{\bibinfo{person}{Kangyan Zhou}, \bibinfo{person}{Shrimai
  Prabhumoye}, {and} \bibinfo{person}{Alan~W. Black}.}
  \bibinfo{year}{2018}\natexlab{d}.
\newblock \showarticletitle{A Dataset for Document Grounded Conversations}. In
  \bibinfo{booktitle}{\emph{Proceedings of {EMNLP} 2018, Brussels, Belgium,
  October 31 - November 4, 2018}}. \bibinfo{pages}{708--713}.
\newblock


\bibitem[\protect\citeauthoryear{Zhou, Gao, Li, and Shum}{Zhou
  et~al\mbox{.}}{2018a}]%
        {zhou2018design}
\bibfield{author}{\bibinfo{person}{Li Zhou}, \bibinfo{person}{Jianfeng Gao},
  \bibinfo{person}{Di Li}, {and} \bibinfo{person}{Heung{-}Yeung Shum}.}
  \bibinfo{year}{2018}\natexlab{a}.
\newblock \showarticletitle{The Design and Implementation of XiaoIce, an
  Empathetic Social Chatbot}.
\newblock \bibinfo{journal}{\emph{CoRR}}  \bibinfo{volume}{abs/1812.08989}
  (\bibinfo{year}{2018}).
\newblock


\bibitem[\protect\citeauthoryear{Zhou, Dong, Wu, Zhao, Yu, Tian, Liu, and
  Yan}{Zhou et~al\mbox{.}}{2016}]%
        {zhou2016multi}
\bibfield{author}{\bibinfo{person}{Xiangyang Zhou}, \bibinfo{person}{Daxiang
  Dong}, \bibinfo{person}{Hua Wu}, \bibinfo{person}{Shiqi Zhao},
  \bibinfo{person}{Dianhai Yu}, \bibinfo{person}{Hao Tian},
  \bibinfo{person}{Xuan Liu}, {and} \bibinfo{person}{Rui Yan}.}
  \bibinfo{year}{2016}\natexlab{}.
\newblock \showarticletitle{Multi-view Response Selection for Human-Computer
  Conversation}. In \bibinfo{booktitle}{\emph{Proceedings of {EMNLP} 2016,
  Austin, Texas, USA, November 1-4, 2016}}. \bibinfo{pages}{372--381}.
\newblock


\bibitem[\protect\citeauthoryear{Zhou and Wang}{Zhou and Wang}{2018}]%
        {zhou2017mojitalk}
\bibfield{author}{\bibinfo{person}{Xianda Zhou} {and}
  \bibinfo{person}{William~Yang Wang}.} \bibinfo{year}{2018}\natexlab{}.
\newblock \showarticletitle{MojiTalk: Generating Emotional Responses at Scale}.
  In \bibinfo{booktitle}{\emph{Proceedings of {ACL} 2018, Melbourne, Australia,
  July 15-20, 2018}}. \bibinfo{pages}{1128--1137}.
\newblock


\bibitem[\protect\citeauthoryear{Zhu, Mo, Zhang, Zhu, Peng, and Yang}{Zhu
  et~al\mbox{.}}{2017}]%
        {zhuwenya2017}
\bibfield{author}{\bibinfo{person}{Wenya Zhu}, \bibinfo{person}{Kaixiang Mo},
  \bibinfo{person}{Yu Zhang}, \bibinfo{person}{Zhangbin Zhu},
  \bibinfo{person}{Xuezheng Peng}, {and} \bibinfo{person}{Qiang Yang}.}
  \bibinfo{year}{2017}\natexlab{}.
\newblock \showarticletitle{Flexible End-to-End Dialogue System for Knowledge
  Grounded Conversation}.
\newblock \bibinfo{journal}{\emph{CoRR}}  \bibinfo{volume}{abs/1709.04264}
  (\bibinfo{year}{2017}).
\newblock


\end{thebibliography}
	
	\begin{table}[t]
	
	\centering
	\begin{threeparttable}
		
		\setlength{\tabcolsep}{1.0mm}
		\setlength{\leftskip}{-10pt}
		\scalebox{0.8}{
			\begin{tabular}{llllll}
				\toprule
				
				Name & Topic & Source & Language & Corpus Statistics & Corpus Features \\
				\midrule
				
				%%%%%opensubtitle 和 cornell movie暂时不包括了       
				%       \hline
				%       \multirow{3}{*}{OpenSubtitles\cite{tiedemann2012parallel}} & \multirow{3}{*}{\shortstack{Administration, \\ legislation, news, etc. \\ several domains}} & \multirow{3}{*}{Movie subtitles} & \multirow{3}{*}{90 languages} & $36 million^*$ dialogs & \multirow{3}{*}{Parallel corpus} \\
				%        &&&& 3.9 turns per dialog & \\ 
				%        &&&& 7.1 words per turn & \\
				
				%        \hline 
				%        \multirow{3}{*}{Cornell Movie-Dialog \cite{danescu2011chameleons}} & \multirow{3}{*}{} & \multirow{3}{*}{Movie subtitles} & \multirow{3}{*}{English} & 220,579 dialogs & \multirow{3}{*}{\shortstack{Gender/position \\ on movie credits}} \\
				%        &&&& 1.4 turns per dialog & \\ 
				%        &&&& $29.5^*$ words per turn & \\
				
				\multirow{2}{*}{STC\cite{shang-2015-NRM}} & \multirow{2}{*}{Open topics} & \multirow{2}{*}{\shortstack{Social media\\(Weibo)}} & \multirow{2}{*}{Chinese} & 219,905 posts & \multirow{2}{*}{\shortstack{One post\\multiple responses}} \\
				&&&& 4,308,211 responses & \\ 
				
				\hline 
				
				\multirow{2}{*}{Twitter Triple\cite{sordoni2015neural}} & \multirow{2}{*}{Open topics} & \multirow{2}{*}{\shortstack{Social media\\(Twitter)}} & \multirow{2}{*}{English} & 29M (c,m,r) triples\tnote{1} & \multirow{2}{*}{\shortstack{Context\\information}} \\
				&&&& 4,232 test/val triples & \\ 
				
				\hline 
				\multirow{3}{*}{Ubuntu Dialog\cite{lowe2015ubuntu}} & Ubuntu & \multirow{3}{*}{\shortstack{Online\\chat log}} & \multirow{3}{*}{English} & 930,000 dialogs & \multirow{3}{*}{\shortstack{Task-specific\\dialog}} \\
				& technical &&& 7.71 turns per dialog & \\ 
				& issues &&& 10.34 words per turn & \\
				
				%%       \hline
				%%       Reddit Corpus\tnote{2} & Open topics & Social media (Reddit) & English & 3.33 billion comments & \\

				\hline 
				\multirow{3}{*}{PersonalDialog\cite{zheng2019personalized}} & \multirow{3}{*}{Open topics} & \multirow{3}{*}{\shortstack{Social media\\(Weibo)}} & \multirow{3}{*}{Chinese} & 20.83 million dialogs & \multirow{3}{*}{\shortstack{Personalization,\\rich user profiles}} \\
				&&&& 56.26M utterances & \\ 
				&&&& 8.47M user profiles & \\
				
				\hline 
				\multirow{3}{*}{Persona-Chat\cite{Zhang2018Personalizing-dogpet}} & \multirow{3}{*}{Daily life} & \multirow{3}{*}{Crowd source} & \multirow{3}{*}{English} & 10,981 dialogs & \multirow{3}{*}{\shortstack{Personalization}} \\
				&&&& 164,356 utterances & \\ 
				%%%&&&&   & \\
				
				\hline        
				
				\multirow{3}{*}{DailyDialog\cite{li2017dailydialog}} & \multirow{3}{*}{\shortstack{Daily life}} & \multirow{3}{*}{Web} & \multirow{3}{*}{English} & 13,118 dialogs & \multirow{3}{*}{\shortstack{Emotion and intent \\annotation}}\\
				&&&& 7.9 turns per dialog & \\ 
				&&&& 14.6 words per turn \\

				\hline
				\multirow{3}{*}{CMU DOG\cite{zhou2018DoG}} & \multirow{3}{*}{\shortstack{30 movies'\\wikipedia page}} & \multirow{3}{*}{Crowd source} & \multirow{3}{*}{English} & 4,112 dialogs & \multirow{3}{*}{Knowledge-grounded} \\
				&&&& 31.6 turns per dialog & \\ 
				&&&& 10.8 words per turn & \\
				
				\hline
				\multirow{3}{*}{Holl-E\cite{moghe2018towards}} & \multirow{3}{*}{921 movies} & \multirow{3}{*}{Crowd source} & \multirow{3}{*}{English} & 9,071 dialogs & 
				\multirow{3}{*}{Knowledge-grounded} \\
				&  &&& 10.0 turns per dialog & \\
				&  &&& 15.3 words per turn & \\
				
				\hline
				Wizard of  & \multirow{2}{*}{\shortstack{1,365 Wikipedia\\ articles}} & \multirow{2}{*}{Crowd source} & \multirow{2}{*}{English} & 22,311 dialogs & \multirow{2}{*}{Knowledge-grounded} \\
				Wikipedia\cite{dinan2018WOW} &&&& 9.0 turns per dialog & \\
				
				\hline
				\multirow{3}{*}{\shortstack{Grounded Response \\ Generation DSTC7} \cite{qin2019conversing}}
				& \multirow{3}{*}{Web articles} 
				&\multirow{3}{*}{ Reddit }
				& \multirow{3}{*}{English }
				& 32.7K dialog-document pairs
				& \multirow{3}{*}{Knowledge-grounded} \\
				&&&& 2.8M utterances & \\
				&&&& 17M document sentences & \\

				\hline
				\multirow{3}{*}{Topical-Chat\cite{gopalakrishnan2019topical}} & \multirow{3}{*}{\shortstack{8 domains, e.g.\\politics, fashion}}& \multirow{3}{*}{Crowd source} & \multirow{3}{*}{English} & 11,319 dialogs & \multirow{3}{*}{Knowledge-grounded}\\
				&&&& 22 turns per dialog & \\
				&&&& 19.8 words per turn \\

				\hline
				\multirow{3}{*}{OpenDialKG\cite{moon2019opendialkg}} & \multirow{2}{*}{\shortstack{Movie, book, \\ sports, music}} & \multirow{2}{*}{Crowd source} & \multirow{2}{*}{English} & 15,673 dialogs & \multirow{2}{*}{Knowlege-grounded}\\
				&&&& 91,209 turns & \\
				
				\hline
				\multirow{3}{*}{DuConv\cite{wu2019proactive}} & \multirow{3}{*}{\shortstack{Films and\\film stars}} & \multirow{3}{*}{Crowd source} & \multirow{3}{*}{Chinese} & 29,858 dialogs & \multirow{3}{*}{\shortstack{Knowledge-grounded/\\Proactivity modeling}}\\
				&&&& 9.1 turns per dialog & \\ 
				&&&& 10.6 words per turn & \\
				
				\hline
				
				\multirow{3}{*}{DyKgChat\cite{chendykgchat}} & \multirow{3}{*}{2 TV series} & \multirow{3}{*}{TV series} & \multirow{3}{*}{\shortstack{Chinese \\English}} & 1,247/3,092 dialogs\tnote{2} & \multirow{3}{*}{{Knowledge-grounded}}\\
				&&&& 13.8/18.7 turns per dialog\tnote{2} & \\ 
				&&&&  27.0/16.5 words per turn\tnote{2} & \\
				
				\hline
				& \multirow{3}{*}{Dialy life} & \multirow{3}{*}{Crowd source} & \multirow{3}{*}{English} & 24,850 dialogs & \multirow{3}*{\shortstack{Emotional/empathetic\\ dialog modeling}}\\
				Empathetic &&&& 4.31 turns per dialog & \\ 
				Dialogues\cite{rashkin2019empa} &&&& 15.2 words per turn & \\  
				
				\hline
				& \multirow{3}{*}{Daily life} & \multirow{3}{*}{Crowd source} & \multirow{3}{*}{English} & 8,939 dialogs & \multirow{3}{*}{\shortstack{Proactivity,\\behavior and strategy}}\\
				Target-Guided   & & & & 101,935 utterances & \\
				Conversation \cite{tang2019target} & & & & 2,678 keywords & \\
				
				\hline
				\multirow{3}{*}{\shortstack{PERSUASION-\\FOR-GOOD}\cite{wang2019persuasion}} & \multirow{3}{*}{Charity donation} & \multirow{3}{*}{Crowd source} & \multirow{3}{*}{English} & 1,017 dialogs & \multirow{3}{*}{\shortstack{Personalization,\\behavior and strategy}}\\
				&&&& 10.43 turns per dialog & \\ 
				&&&& 19.36 words per utterance & \\
				
				\bottomrule
			\end{tabular}
		}
		\caption{Open-domain Dialog Corpora. We only list the datasets that are frequently used or recently proposed\tnote{3}. %%%A complete corpus survey was published in 2015 \cite{serban2015survey}. %%%%
		} 
		\label{tab:corpora}
		\begin{tablenotes}
			\footnotesize
			\item[1] (c,m,r) means a triple of (context, message, response).
			%%%%%       \item[2] {\url{https://www.reddit.com/r/datasets/comments/6mvrb5/reddit_june_2017_comments_are_now_available/}}.
			\item[2] The first number is for the Chinese TV series and the second for the English one.
			\item[3] A complete survey on older datasets was published in 2015 \cite{serban2015survey} so that we do not include those corpora.
		\end{tablenotes}
	\end{threeparttable}

	\end{table}
	
	%%%%可能要加表格注释
	
\end{document}